\documentclass[journal]{vgtc}                     

\onlineid{1782}

\vgtccategory{Research}

\vgtcpapertype{application}

\title{CD-TVD: Contrastive Diffusion for 3D Super-Resolution with Scarce High-Resolution Time-Varying Data}

\author{%
  \authororcid{Chongke Bi}{0000-0002-4324-8028},
  \authororcid{Xin Gao}{0000-0003-2715-7633},
  \authororcid{Jiakang Deng}{0009-0001-0257-2390},
  \authororcid{Guan Li \textsuperscript{*}}{0000-0001-6436-3650},
  and
  \authororcid{Jun Han}{0000-0002-7286-062X}
}

\authorfooter{
  \item
    Chongke Bi, Xin Gao, and Jiakang Deng are with College of Intelligence and Computing, Tianjin University.
    E-mails: bichongke@tju.edu.cn, gao\_xin\_private@163.com, closernh@163.com.
  \item
    Guan Li (corresponding author) is with the Computer Network Information Center, Chinese Academy of Sciences. Guan Li is also with the University of Chinese Academy of Sciences.
    E-mail: liguan@cnic.cn.
  \item
    Jun Han is with the Division of Emerging Interdisciplinary Areas and Center for Ocean Research in Hong Kong and Macau (CORE), The Hong Kong University of Science and Technology. E-mail: hanjun@ust.hk.
}

\abstract{%
	Large-scale scientific simulations require significant resources to generate high-resolution time-varying data (TVD). While super-resolution is an efficient post-processing strategy to reduce costs, existing methods rely on a large amount of HR training data, limiting their applicability to diverse simulation scenarios. To address this constraint, we proposed CD-TVD, a novel framework that combines contrastive learning and an improved diffusion-based super-resolution model to achieve accurate 3D super-resolution from limited time-step high-resolution data. During pre-training on historical simulation data, the contrastive encoder and diffusion super-resolution modules learn degradation patterns and detailed features of high-resolution and low-resolution samples. In the training phase, the improved diffusion model with a local attention mechanism is fine-tuned using only one newly generated high-resolution timestep, leveraging the degradation knowledge learned by the encoder. This design minimizes the reliance on large-scale high-resolution datasets while maintaining the capability to recover fine-grained details. Experimental results on fluid and atmospheric simulation datasets confirm that CD-TVD delivers accurate and resource-efficient 3D super-resolution, marking a significant advancement in data augmentation for large-scale scientific simulations. The code is available at \url{https://github.com/Xin-Gao-private/CD-TVD}.
	
}


\teaser{
  \centering
  \includegraphics[width=0.85\linewidth, alt={Overview of the CD-TVD framework for 3D super-resolution.}]{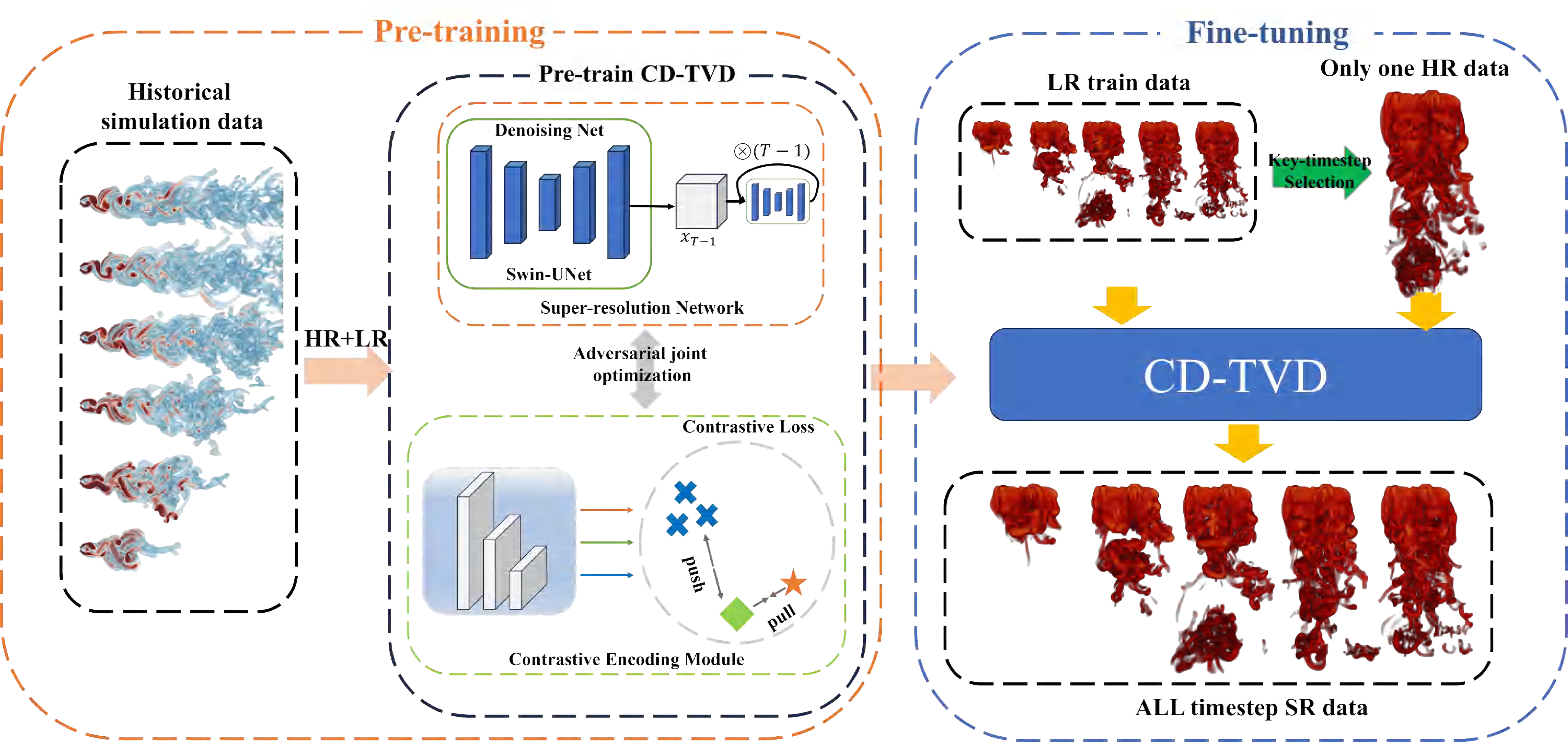}
  \caption{%
  	Overview of the CD-TVD framework for 3D super-resolution. The framework consists of two stages: pre-training with historical simulation data and fine-tuning for new scenarios. In the pre-training phase (left), the contrastive encoding module learns degradation patterns between high-resolution, low-resolution, and super-resolution data, while the diffusion super-resolution module captures fine-grained details using adversarial training with a local attention mechanism to reduce computational costs. In the fine-tuning phase (right), the contrastive module is frozen, and minimal high-resolution samples are used to adapt the model, ensuring accurate reconstruction across all low-resolution time steps in new datasets with minimal reliance on high-resolution data.%
  }
  \label{fig:teaser}
}

\graphicspath{{figs/}{figures/}{pictures/}{images/}{./}} 

\usepackage{tabularx}
\usepackage{booktabs}                  
\usepackage{lipsum}                    
\usepackage{mwe}                       
\usepackage{amsmath}
\usepackage{amssymb}
\usepackage{stfloats}
\usepackage{multirow}
\usepackage{mathptmx}
\usepackage{makecell}
\usepackage{fontawesome5} 
\usepackage{marvosym}

\begin{document}

\firstsection{Introduction}
\maketitle

In scientific numerical simulation, improving simulation accuracy allows results to approximate the essential characteristics of complex physical phenomena more closely. However, constrained by the conflict between computational resources and spatio-temporal resolution, directly conducting high-resolution (HR) simulations often faces bottlenecks, such as exponentially increasing computational costs and soaring data storage demands~\cite{wang2022dl4scivis}. Although traditional low-precision calculations enhance efficiency, they often fail to capture evolutionary details of microscopic structures or abrupt transition features in critical states, leading to significant degradation in prediction accuracy. Super-resolution (SR) technology addresses this challenge by establishing a mapping relationship from low-precision data to HR spaces, thereby enabling the effective recovery of fine-grained structures in key physical fields with limited computational resources~\cite{shen2024generative,yang2024adaptive}.

Deep learning models provide a powerful methodology for super-resolution tasks in scientific visualization. These approaches construct deep neural networks capable of automatically extracting multiscale spatio-temporal features from massive low-resolution (LR) simulation data, thereby establishing end-to-end mapping relationships from low-dimensional features to HR physical fields~\cite{shen2024pcsagan}. The superior performance of deep learning models is highly dependent on the support of large-scale training data. For super-resolution tasks, the key challenge lies in constructing high-quality low-resolution-to-high-resolution datasets. The model is then trained under a supervised learning paradigm by minimizing the discrepancy between predicted results and authentic HR physical fields, progressively mastering the intricate nonlinear mapping patterns between them~\cite{liu2024uginr}. Well-trained models can effectively perform super-resolution tasks in specific scenarios, with empirical studies demonstrating that the reconstruction quality exhibits a strong positive correlation with both the quantity and quality of training data. Substantial training samples with precise alignment have become a critical point in achieving optimal super-resolution reconstruction outcomes~\cite{deng2019super}.

However, the scarcity of high-precision data renders existing super-resolution methods challenging to be effectively applied in real-world applications. Training deep learning models requires substantial training data support, but getting high-fidelity scientific data remains prohibitively expensive~\cite{karniadakis2021physics}. High-precision simulations demand enormous computational resources; for instance, a single HR case in CFD simulations often requires weeks of GPU cluster computation, leading to exponentially increasing costs for obtaining high-low resolution data pairs. Furthermore, scientific data exhibit unique characteristics, such as coupled multi-physics fields, nonlinear spatiotemporal evolution, and strict conservation law constraints. Simple data augmentation techniques (e.g., rotation, cropping) risk disrupting the continuity of physical fields, while cross-domain transfer learning approaches (e.g., natural image pre-trained models~\cite{liu2023pre}) may introduce artifacts that violate physical principles. 

To address this problem, we propose CD-TVD, a novel model for three-dimensional super-resolution tasks with scarce HR temporal data. By explicitly modeling HR-LR degradation relationships through contrastive pairs (HR as positives, LR as negatives), CD-TVD learns discriminative features of structural and high-frequency losses, enhancing generalization to unseen data distributions. Furthermore, we design a two-stage super-resolution framework that pre-trains an embedding network (ED) and super-resolution network (FSR) via iterative adversarial training. Once ED encodes degradation-aware features, a frozen diffusion-based FSR jointly optimizes pixel-level reconstruction and contrastive loss, enabling comprehensive degradation modeling from historical data. For new scenarios, only a single HR timestep is required to fine-tune the pre-trained model for accurate reconstruction across all LR timesteps. To balance fidelity and efficiency, we also introduce a local attention-enhanced diffusion architecture that shares parameters with the contrastive module, preserving detail recovery while reducing computational overhead. This synergy allows stable 3D reconstruction from LR inputs by leveraging pre-learned degradation patterns. Experiments demonstrate our framework's dual capability: capturing degradation patterns from historical data while adapting to new scenarios with minimal HR samples, proving its scalability for large-scale simulations. Our main contributions can be summarized as follows: 
\begin{itemize}
	\item We explicitly treat the degradation process between HR and LR as a contrastive learning task, thereby extracting strongly discriminative degradation features from historical data and achieving 3D super-resolution generalization across various scenarios.
	\item A local attention mechanism is integrated into the diffusion model for super-resolution tasks, substantially alleviating conventional diffusion approaches' computational and memory burdens while enabling fine-grained recovery of HR structures.
	\item Leveraging the universal degradation patterns learned during pre-training, our model can reconstruct all subsequent LR timesteps with only minimal HR timesteps in a new dataset, significantly reducing the need for additional HR data and further enhancing SR's practicality in large-scale scientific simulations.
\end{itemize}

\section{Related Work}

We adopt the approach based on conditional diffusion models for the super-resolution task of time-varying data\cite{rombach2022high}. In this section, we provide a comprehensive overview of the related work on super-resolution techniques specifically tailored for scientific data, along with a focused discussion on the rapidly emerging field of diffusion models.

\subsection{Super-Resolution in Scientific Visualization}


Rapid developments in deep learning have significantly advanced super-resolution techniques in scientific visualization \cite{9920542, shen2023psrflow, jiao2024ffeinr}, particularly for scientific data \cite{wang2020deep, yang2019deep, wang2024semi}.

For scalar data, convolutional neural networks (CNNs) were first utilized by Zhou \textit{et al.} \cite{zhou2017volume}, converting LR volumetric data into HR to enhance exploration efficiency. Generative Adversarial Networks (GANs) have been applied for both temporal super-resolution (TSR) and spatial super-resolution (SSR) in time-varying datasets~\cite{wang2024pmim}, leading to methodologies like TSR-TVD \cite{han2019tsr} and SSR-TVD \cite{han2020ssr}. Han \textit{et al.} \cite{han2021stnet} proposed STNet, a generative framework using unsupervised pre-training and cycle-consistent loss on octree boundaries to reconstruct HR spatiotemporal volumes directly from LR data.

For vector data, Guo \textit{et al.} \cite{guo2020ssr} introduced SSR-VFD, the first framework leveraging separate neural networks for each vector component, effectively preserving streamline rendering details. Han and Wang \cite{han2022tsr} incorporated recurrent generative networks into vector data super-resolution, synthesizing intermediate volume sequences via bidirectional predictions. An \textit{et al.} \cite{an2021stsrnet} proposed STSRNet, a deep joint spatiotemporal super-resolution method well-suited for large-scale simulations constrained by storage limits \cite{song2024forecasting,jiao2023esrgan}.

Despite these advancements, current techniques heavily depend on training data characteristics, limiting their generalizability to significantly different datasets and their ability to reconstruct complex textures and subtle features accurately \cite{zuo2023high,lepcha2023image}.

The methods mentioned above have brought about significant improvements in scientific visualization. However, they have certain limitations as they rely heavily on specific patterns and features in the training data\cite{zuo2023high}. Consequently, their performance may not be optimal when applied to data that diverges significantly from that in the training set. Additionally, these methods rely on the limited information extracted from LR data, potentially limiting their effectiveness in accurately reconstructing complex textures and subtle features\cite{lepcha2023image}.

\subsection{Diffusion Models for Super-Resolution}

Diffusion models \cite{Yang2022DiffusionMA} represent advanced probabilistic generative deep learning frameworks with remarkable performance in image and audio synthesis tasks. These models rely on a data diffusion process, gradually introducing noise and subsequently learning the reverse process to restore original data \cite{Nichol2021ImprovedDD,cao2024survey}. Unlike GANs \cite{xie2018tempogan}, diffusion models avoid training instability issues \cite{NEURIPS2021_49ad23d1,xiao2021tackling}.

In recent research, Saharia \textit{et al.} \cite{saharia2022image} employed diffusion models to generate high-quality images from LR inputs, learning a reverse process to achieve detailed outputs. Daniels \textit{et al.} \cite{daniels2021score} introduced a score-based super-resolution method utilizing Sinkhorn couplings and optimal transport theory, while Metzger \textit{et al.} \cite{Metzger_2023_CVPR} improved guided depth super-resolution through anisotropic diffusion guided by deep convolutional networks. Yue \textit{et al.} \cite{NEURIPS2023_2ac2eac5} implemented a Markov chain approach to significantly reduce diffusion steps by manipulating residuals between HR and LR images. Gao \textit{et al.} \cite{Gao_2023_CVPR} developed an end-to-end framework combining implicit neural representations and denoising diffusion, introducing scale-controllable conditioning. Li \textit{et al.} \cite{LI202247} accelerated convergence in Single-Image Super-Resolution diffusion models through residual prediction.

Diffusion models have also improved precision in medical \cite{CHEN201944,VIS2021118673,ning2016joint} and remote sensing imaging \cite{wang2022comprehensive,karatsiolis2022exploiting,gao2024bayesian}. Notably, Chung \textit{et al.} \cite{9941138} proposed score-based reverse diffusion for denoising complex noise distributions, while Croitoru \textit{et al.} \cite{croitoru2023diffusion} utilized diffusion for resolution enhancement. Liu \textit{et al.} \cite{rs14194834} applied diffusion models for detailed supplementation in remote sensing super-resolution. Schranz \textit{et al.} \cite{schanz2023stochastic} employed denoising diffusion with filter-boosted training for cosmic structure super-resolution, and Wu \textit{et al.} \cite{WU2023104901} introduced self-attention mechanisms for efficient and precise MRI image super-resolution.

Although diffusion models have significantly advanced image \cite{song2023sparse,rombach2022high} and video generation \cite{ho2022cascaded,ho2022video}, their integration with scientific visualization remains unexplored. Incorporating diffusion methods into scientific visualization could potentially overcome current limitations in generalization, complex textures, and detail handling in super-resolution tasks.

\begin{figure}[t] 
	\centering 
	\includegraphics[width=0.5\textwidth,height=0.5\textheight,keepaspectratio]{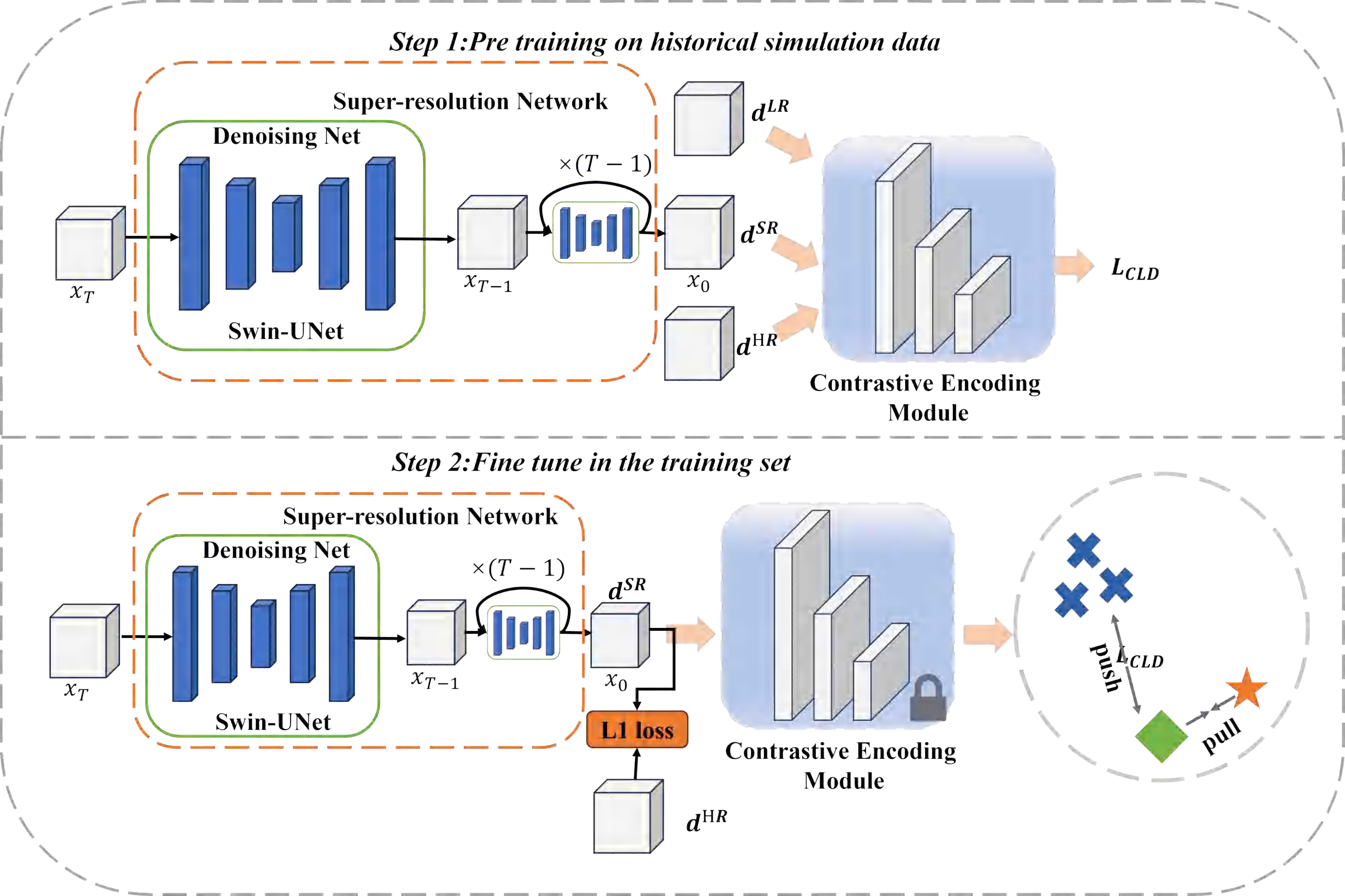} 
	\caption{Overview of CD-TVD. The model is trained in two stages: in the first stage, the super-resolution module is frozen while training the contrastive encoding module; in the second stage, the encoding module is frozen while training the super-resolution module. The training is done through adversarial learning, optimizing both modules simultaneously.} 
	\label{fig:Overview} 
\end{figure}

\section{Method}

In scientific super-resolution tasks, a key challenge is the difficulty in acquiring HR data, as well as the impact of scarce HR data on the effectiveness of super-resolution methods. To address this issue and reduce reliance on HR data, we propose the CD-TVD model, illustrated in Fig.~\ref{fig:teaser}, which leverages contrastive learning on historical simulation data to capture the degradation patterns between HR and LR data. At the same time, the diffusion super-resolution module learns fine-grained and detailed features, enabling precise reconstruction of the missing high-frequency components. For new scenarios, only a single HR timestep is required to fine-tune the pre-trained model, enabling accurate reconstruction across all LR timesteps.

The method follows a two-stage pipeline: pre-training with historical simulation data and fine-tuning for new scenarios. In the pre-training phase, both the contrastive encoding module and the diffusion super-resolution module are jointly trained on a large set of historical simulation data, as shown in Fig.~\ref{fig:Overview}. The contrastive encoding module learns degradation patterns by contrasting HR, LR, and SR data, while the diffusion super-resolution module focuses on the super-resolution task, incorporating a local attention mechanism to reduce computational costs while ensuring fine-grained reconstruction. Both modules are jointly optimized through adversarial training, allowing the model to capture general degradation features and improve robustness to various degradation scenarios. In the fine-tuning stage, the contrastive encoding module is frozen to preserve the learned prior knowledge, and only a small number of HR samples are used to fine-tune the diffusion super-resolution module. This fine-tuning process compensates for the missing high-frequency details in the new dataset, enabling precise super-resolution with minimal HR data.

%
%
\begin{figure}[tb]
	\centering 
	\includegraphics[width=0.5\textwidth,height=0.5\textheight,keepaspectratio]{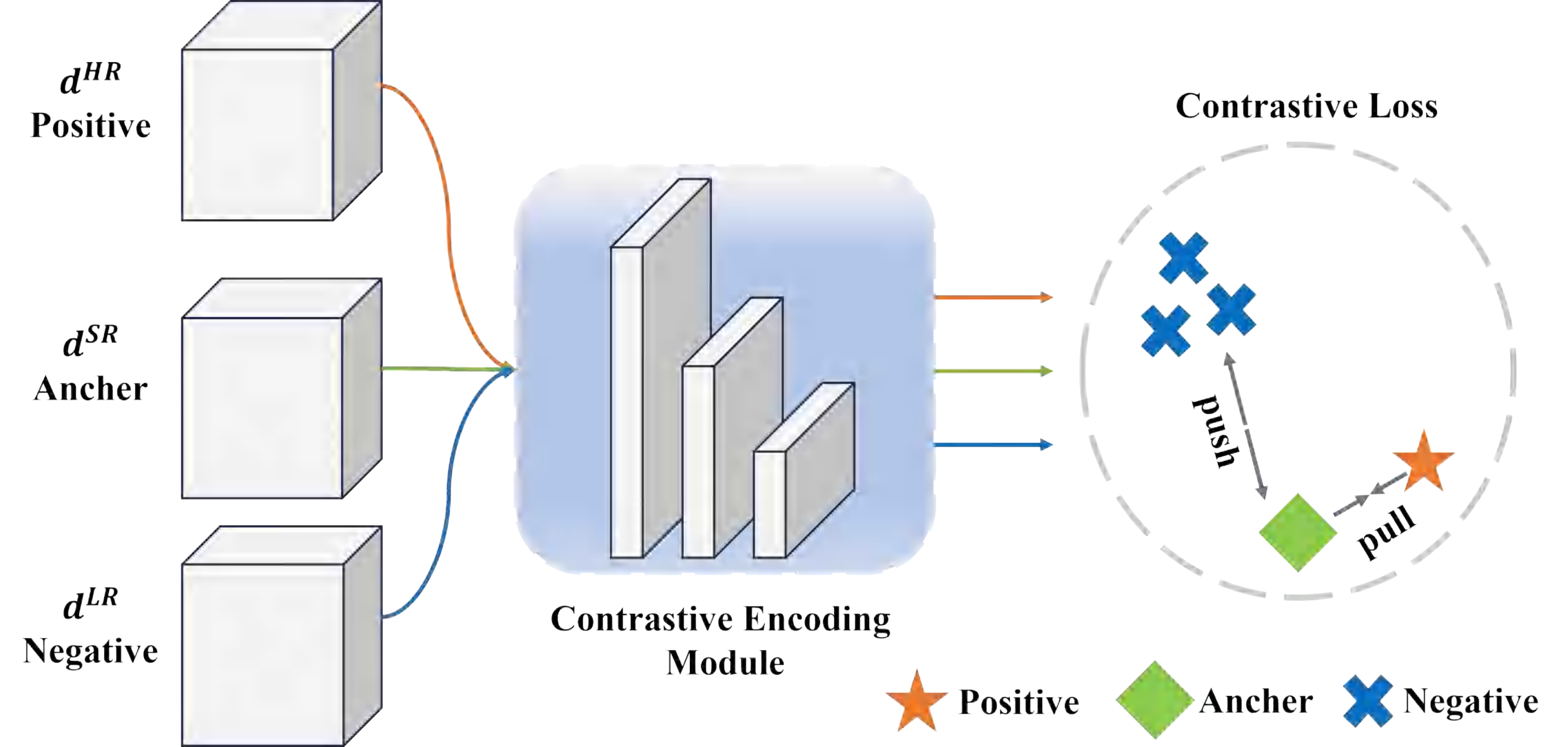} 
	\caption{
		Illustration of the Contrastive Encoding Module. $d^{HR}$, $d^{SR}$, and $d^{LR}$ denote original HR, SR, and LR data, respectively. A convolutional encoder extracts features, trained with contrastive regularization in latent space to emphasize degradation-aware distinctions, guiding SR data closer to HR data.
	} 
	\label{fig:Contrastive_Encoding} 
\end{figure}

\subsection{Contrastive Encoding Module}
The contrastive encoding module is built upon contrastive learning, where the model learns meaningful representations by comparing similar and dissimilar examples. Specifically, we define positive and negative sample pairs to reflect the degradation-aware characteristics of the task. The model then learns embeddings that distinguish between HR (positive) and LR (negative) representations using a contrastive learning loss. Finally, a degradation-aware embedding network is introduced to extract features that are sensitive to high-frequency differences, especially in scientific 3D data. This structure enables the model to effectively recover fine details in super-resolution outputs. Based on this, we built the Contrastive Encoding Module, as shown in Fig. \ref{fig:Contrastive_Encoding}.

\subsubsection{Positive and Negative Sample Strategy}

In our method, the goal is to learn the degradation patterns between HR and LR 3D scientific data. Specifically, in our Contrastive Encoding Module, we generate positive and negative pairs using clear 3D data volumes (\( J \)) and their deblurred counterparts (\( \hat{J} \)) produced by the encoder trained through the adversarial learning, as well as pairs involving \( \hat{J} \) and the blurry 3D data volume (\( I \)). For simplicity, we refer to the restored volume, clear volume, and blurry volume as the anchor, positive anchor, and negative anchor, respectively.

Unlike autoencoder-based methods, we train the encoder through adversarial learning to generate high-quality restored 3D volumes. The adversarial training approach allows the encoder to learn more realistic and high-frequency details, making it more robust to complex degradation patterns typically observed between HR and LR 3D data. The objective function can be reformulated as follows:

\begin{equation}\label{equ-4-6}
	\min \| J - \phi(I, w) \| + \beta \cdot \rho(G(I), G(J), G(\phi(I, w))),
\end{equation}
where the first term represents the reconstruction loss aligning the restored 3D volume $\phi(I, w)$ with its ground truth volume $J$ in the data manifold. We use the L1 loss, as it has been shown to perform better than L2 loss in super-resolution tasks \cite{zuo2023high}. The second term $\rho(G(I), G(J), G(\phi(I, w)))$ represents the contrastive regularization calculated within the latent feature space generated by the \textbf{Contrastive Encoding Module}, denoted by $G(\cdot)$. Specifically, this module maps input volumes into a latent feature space to capture discriminative, degradation-aware features. This regularization acts as a contrasting force: it pulls the restored volume $\phi(I, w)$ closer to the clear volume $J$, while pushing it away from the blurry volume $I$. The hyperparameter $\beta$ balances the reconstruction loss and the contrastive regularization term, and is selected through cross-validation.

To enhance the contrastive capability, we extract hidden features from different layers of a fixed pre-trained model. This approach enables the model to focus on fine-grained details at multiple levels of abstraction, facilitating better feature alignment and more effective recovery of high-frequency details in 3D data volumes.

\begin{figure}[t] 
	\centering 
	\includegraphics[width=0.5\textwidth,height=0.5\textheight,keepaspectratio]{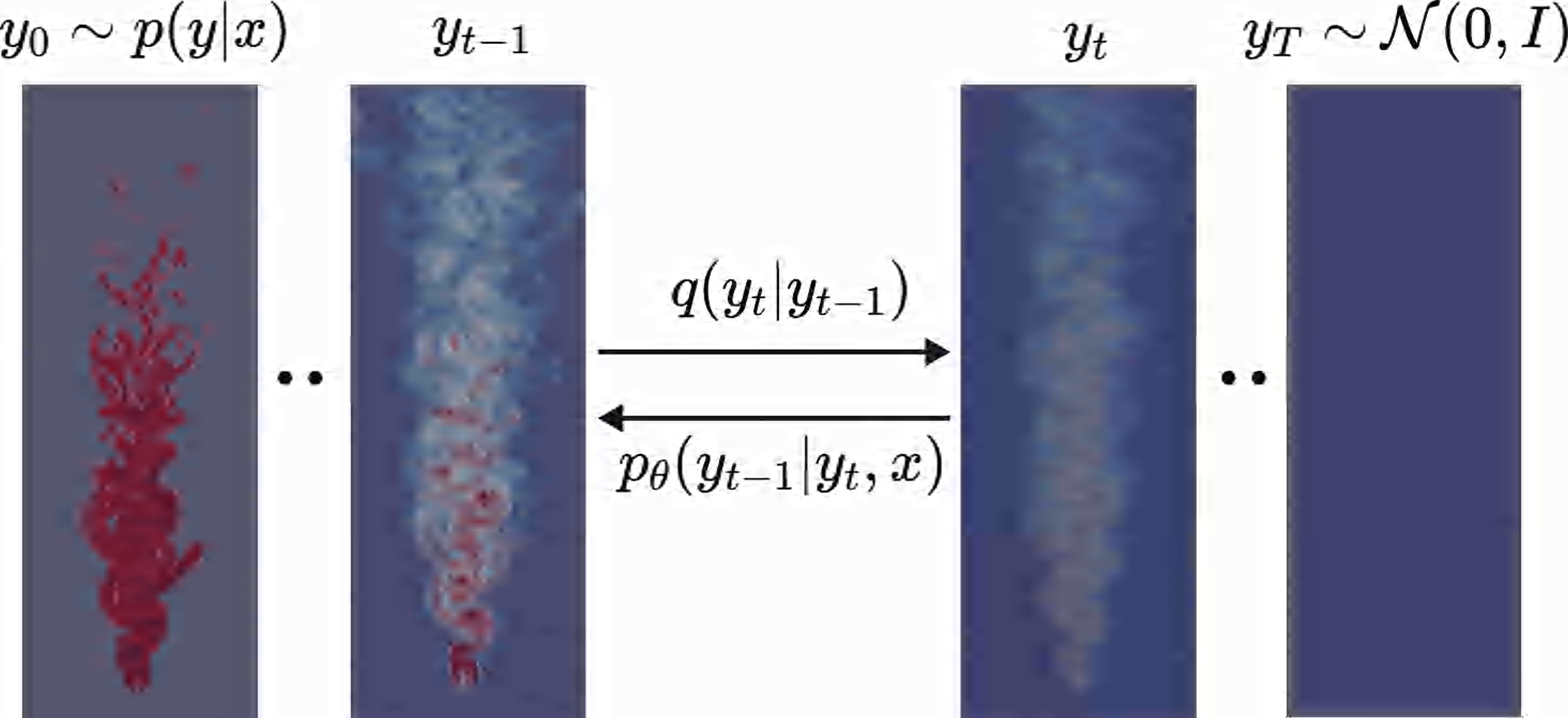} 
	\caption{Forward and reverse processes of CD-TVD, with forward process $q$ generating a noisy data sequence (left to right) by gradually adding Gaussian noise, and reverse process $p$ iteratively refining HR data (right to left).} 
	\label{fig:diffusion_prosses} 
\end{figure}

\subsubsection{Network Architecture}

In the original data space, the data typically has high dimensionality and may contain a significant amount of noise or redundant information. By mapping the data to a latent space, the model can extract more compact and meaningful representations. Traditional contrastive learning methods often rely on pre-trained models like VGG to learn latent space representations. However, these pre-trained models are not specifically tailored to the task or data at hand and may not be optimal for domain-specific tasks, such as super-resolution of scientific data.

For the task of super-resolution in scientific data, we believe that identifying degradation differences is the most crucial goal for contrastive learning. In traditional tasks, latent space representations are learned based on high-level semantic features, but in the case of scientific data, the focus should be on learning the fine-grained differences between HR data and LR data, especially the high-frequency details that are lost during the degradation process.

To achieve this, we employ a convolutional encoder trained using a contrastive loss, enabling it to extract degradation-aware features from the input data. This encoder consists of multiple convolutional layers with three downsampling operations, progressively capturing critical degradation patterns at different scales. By focusing explicitly on the subtle differences between HR and LR data, the encoder effectively filters out irrelevant information and extracts meaningful low- and high-frequency features crucial for super-resolution.

Unlike traditional binary classification approaches, we incorporate a contrastive regularization term in the same latent feature space to train the encoder~\cite{Wu2023Practical}.
This training approach provides discriminator-like supervision, effectively enabling the encoder to capture subtle degradation differences between HR and LR data. Specifically, this contrastive mechanism guides the encoder to distinguish clearly between high-frequency and structural details crucial for accurate super-resolution reconstructions.

The contrastive learning loss is formulated as follows:

\begin{multline}\label{equ-4-4}
	\mathcal{L}_{\mathrm{CLD}} = \mathbb{E}_{I^{\text{HR}}} \biggl[ 
	-\log \left( \frac{
		\exp\bigl(E_{\text{D}}(I_{\text{HR}})\bigr)
	}{
		\sum_{I_{\text{HR}}} \exp\bigl(E_{\text{D}}(I_{\text{HR}})\bigr) +
		\sum_{I_{\text{LR}}} \exp\bigl(E_{\text{D}}(I_{\text{SR}})\bigr)
	} \right)
	\biggr] \\[8pt]
	+ \mathbb{E}_{I^{\text{LR}}} \biggl[
	-\log \left( \frac{
		\exp\bigl(-E_{\text{D}}(I_{\text{SR}})\bigr)
	}{
		\sum_{I_{\text{HR}}} \exp\bigl(-E_{\text{D}}(I_{\text{HR}})\bigr) +
		\sum_{I_{\text{LR}}} \exp\bigl(-E_{\text{D}}(I_{\text{SR}})\bigr)
	} \right)
	\biggr].
\end{multline}

\begin{figure}[tb] 
	\centering 
	\includegraphics[width=0.5\textwidth,height=0.5\textheight,keepaspectratio]{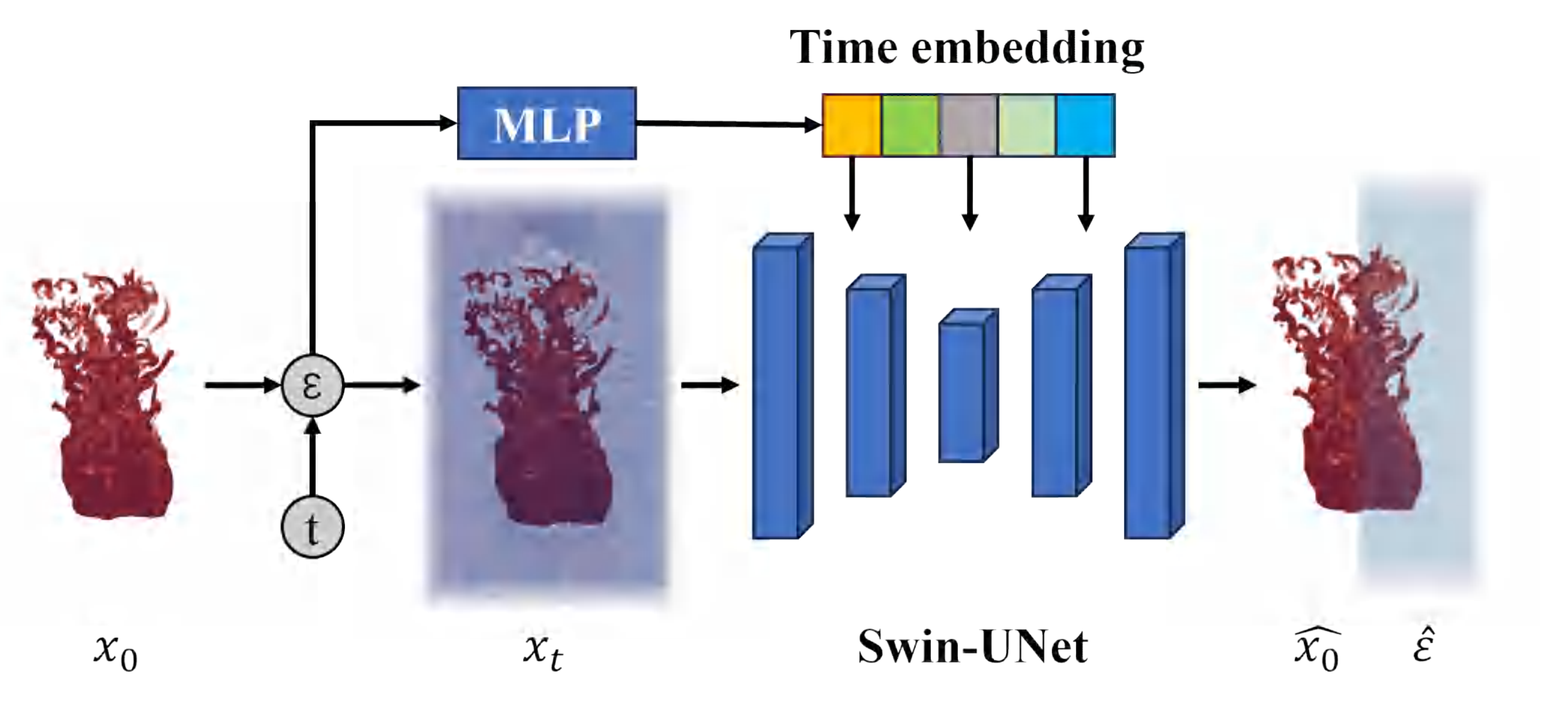} 
	\caption{The denoising network in the diffusion model. The Swin-Conv block combines the advantages of convolution and attention mechanisms, enabling the integration of both global and local information for improved resolution of 3D data.} 
	\label{fig:diffusion_unet} 
\end{figure}

\begin{figure*}[t] 
	\centering 
	\includegraphics[width=1\textwidth,height=0.25\textheight]{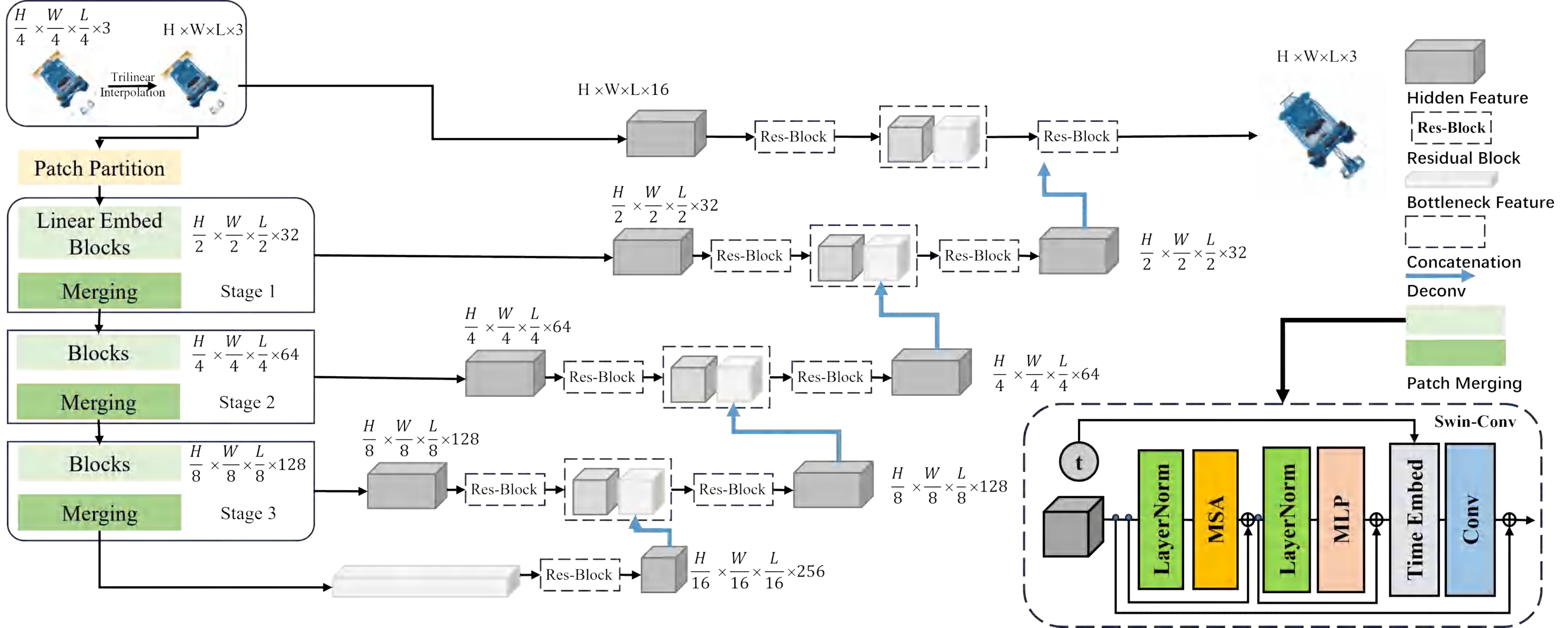} 
	\caption{
		The overall architecture of the proposed three-stage SwinUNet integrated with the Local Attention Block. The encoder partitions input 3D data into non-overlapping patches, embedding them into feature tokens processed through MSA layers. Residual connections and convolution layers within each Local Attention Block enhance the extraction of high-frequency structural details, while skip connections facilitate precise feature integration during decoding, enabling accurate reconstruction of fine-grained textures and edges.
	} 
	\label{fig:swinunet} 
\end{figure*}


where \( I_{\text{SR}} = F_{\text{SR}}(I_{\text{LR}}) \). By using a one-against-the-batch classification, the discriminator can identify subtle degradation differences between HR and SR data, which helps the encoder focus on recovering high-frequency details. 

This architecture ensures that the model not only preserves global consistency in 3D super-resolution tasks but also recovers fine structures essential for scientific analysis.

\subsection{Super-resolution Network}

Super-resolution techniques require the capture of fine-grained details to achieve high-quality reconstruction, especially critical in scientific data analysis. Diffusion-based methods have recently demonstrated superior performance and generalizability in image super-resolution tasks due to their robust capability to model intricate patterns and subtle textures. The diffusion process, as shown in Fig.~\ref{fig:diffusion_prosses}, includes two steps: a forward process, where Gaussian noise is gradually added to the data, and a reverse process, where the noisy data is iteratively refined to recover HR information. This denoising process enables the model to achieve high-quality reconstructions. However, existing diffusion models are primarily employed for image-based super-resolution or high-level vision tasks, such as classification in 3D scenarios, mainly due to their high computational and memory demands.

Specifically, during the training phase, we adopt the cosine noise scheduling strategy, employing a total of $1000$ diffusion steps to ensure sufficient noise refinement and effective capture of high-frequency details. For inference, we significantly reduce computational overhead by limiting the diffusion process to $20$ steps, which we empirically found to effectively balance the quality of reconstruction and computational efficiency. Additionally, the diffusion model is explicitly conditioned on LR data by concatenating LR feature maps directly to the input of the denoising diffusion network. This conditioning mechanism guides the denoising process, enabling a precise and efficient recovery of fine-scale details from the LR input.

\subsubsection{Network Architecture}

Directly extending diffusion-based approaches to HR 3D scientific data poses significant computational challenges, severely limiting their applicability under resource-constrained conditions. To overcome these limitations, we propose a diffusion-based architecture integrated with a local attention mechanism. Fig.~\ref{fig:diffusion_unet} illustrates the proposed denoising process. This design effectively manages computational complexity by allowing the network to selectively focus on informative regions. Given the regularity and spatial-temporal correlations inherent in 3D scientific data, local attention is particularly effective for capturing relevant features in localized areas, enabling efficient computation without sacrificing performance.

The super-resolution network adopts a diffusion-based architecture augmented with a local attention mechanism. The diffusion layers iteratively enhance the resolution of the input data, while the local attention mechanism enables the model to selectively prioritize regions needing finer detail. Such a design is well-suited for three-dimensional scientific data, where both spatial and temporal dependencies must be leveraged for accurate reconstruction.

\subsubsection{Local Attention Block}

The Local Attention Block is a key module in our three-stage SwinUNet architecture, designed to enhance the extraction of fine-grained features within a hierarchical encoder-decoder framework. Our block applies localized attention within non-overlapping 3D windows at each resolution level of the encoder.

Specifically, the input 3D volume is initially partitioned into patches through a patch partitioning layer. These patches are then linearly embedded into feature tokens and processed by Swin Transformer Blocks. Each block consists of Window-based Multi-head Self-Attention layers, interleaved with Layer Normalization and Multi-Layer Perceptron (MLP) units. This architectural choice enables the model to capture both localized and context-aware features while maintaining computational efficiency.

Our SwinUNet employs three hierarchical encoder stages, illustrated in Fig.~\ref{fig:swinunet}. At each stage, the Local Attention Block refines the feature representations through residual connections and convolutional layers, which are particularly effective in preserving high-frequency components such as edges and textures. These refined features are then progressively passed through the decoder, where each level integrates information from the corresponding encoder stage via skip connections and is upsampled through deconvolution operations. This structure ensures the accurate reconstruction of spatial details during the decoding process.

\subsection{Entropy-based Key-timestep Selection}\label{sec:entropy_selection}

Unlike conventional super-resolution methods that require multiple HR snapshots for model adaptation, our framework fine-tunes using only \textbf{a single HR timestep} in the new scenario. This capability is distinctive and crucial, but it raises the essential question: \textbf{how to select the most representative timestep} for fine-tuning to ensure effective generalization across the entire sequence.

We think that the timestep exhibiting the highest entropy in its LR counterpart is the most informative, as it typically corresponds to the richest and most complex system structures. Hence, we propose an \textbf{entropy-guided keyframe selection} strategy.

To select the optimal timestep, we compute entropy for each LR timestep by evaluating the distribution of pixel intensities or feature values, defined by the formula:

\begin{equation}
	H(X_t) = - \sum_{i=1}^{n} p(x_i) \log(p(x_i)),
\end{equation}
where \( X_t \) represents the feature set or pixel intensities in the \( t \)-th LR timestep, and \( p(x_i) \) is the probability distribution of these values. The entropy reflects the uncertainty or the richness of information within a given LR timestep—higher entropy indicates more variation or complexity in the data.

Once the entropy values \( H(X_t) \) for all timesteps are calculated, we select the timestep \( t_{\text{max}} \) corresponding to the highest entropy value, i.e.,

\begin{equation}
	t_{\text{max}} = \arg\max_{t} H(X_t).
\end{equation}

This ensures that the timestep \( t_{\text{max}} \) represents the most informative timestep, which will be used for fine-tuning, allowing for better generalization to unseen timesteps.

\section{Results And Discussion}
\subsection{Datasets and Network Training}


\begin{figure*}[t]
	\centering
	\begin{tabular}{cccccc}
		\includegraphics[width=0.14\textwidth,height=0.15\textheight,keepaspectratio]{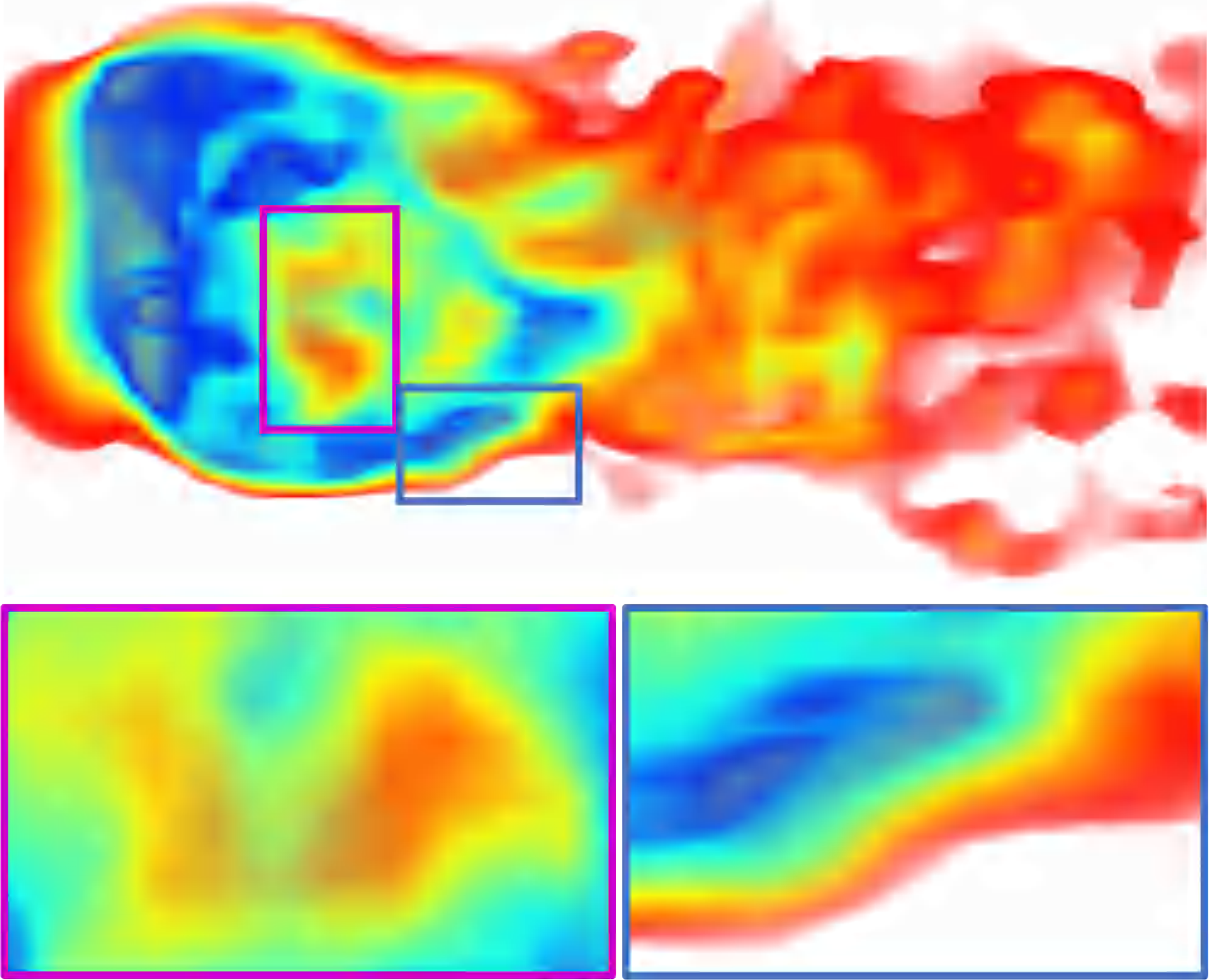} &
		\includegraphics[width=0.14\textwidth,height=0.15\textheight,keepaspectratio]{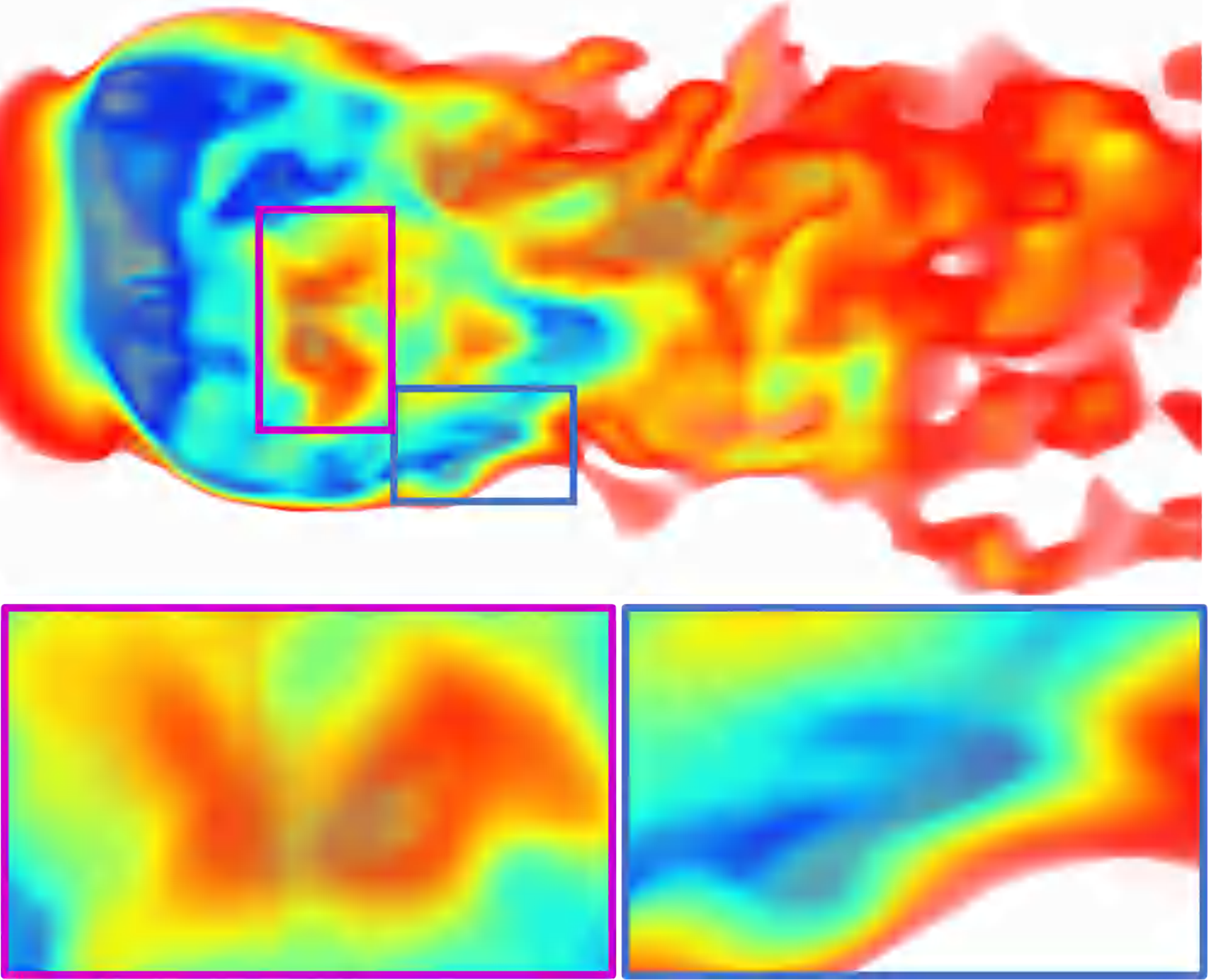} &
		\includegraphics[width=0.14\textwidth,height=0.15\textheight,keepaspectratio]{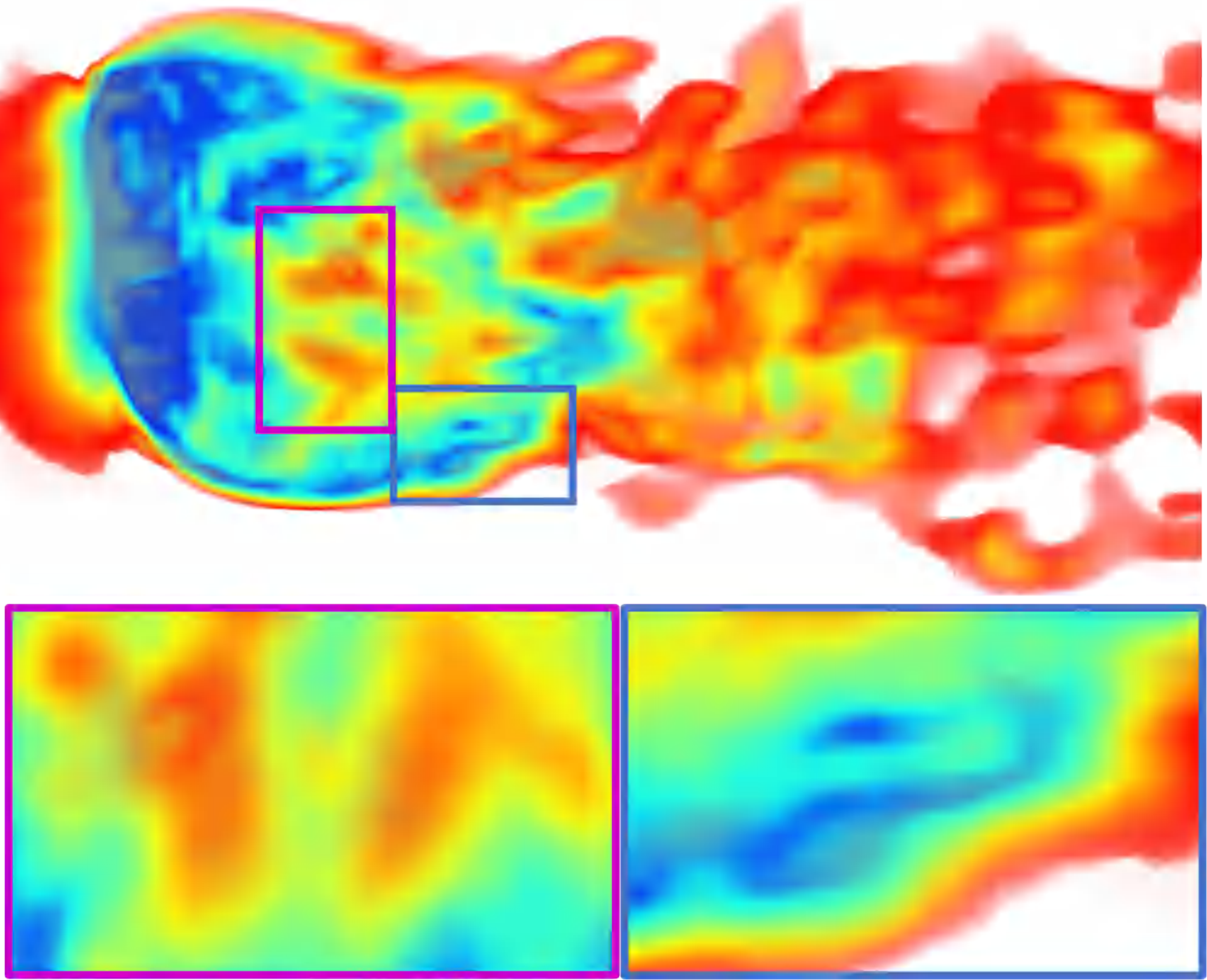} &
		\includegraphics[width=0.14\textwidth,height=0.15\textheight,keepaspectratio]{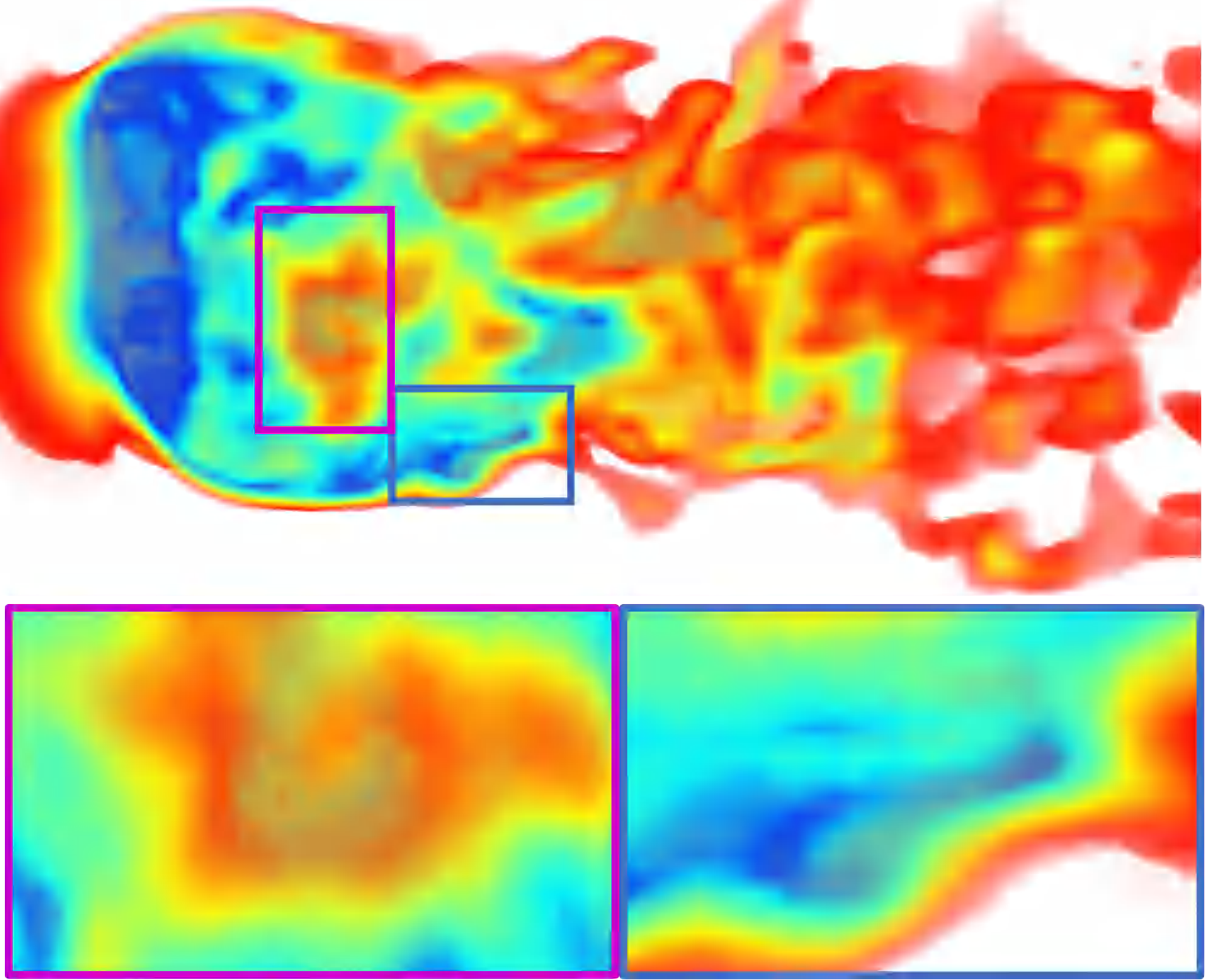} & 
		\includegraphics[width=0.14\textwidth,height=0.15\textheight,keepaspectratio]{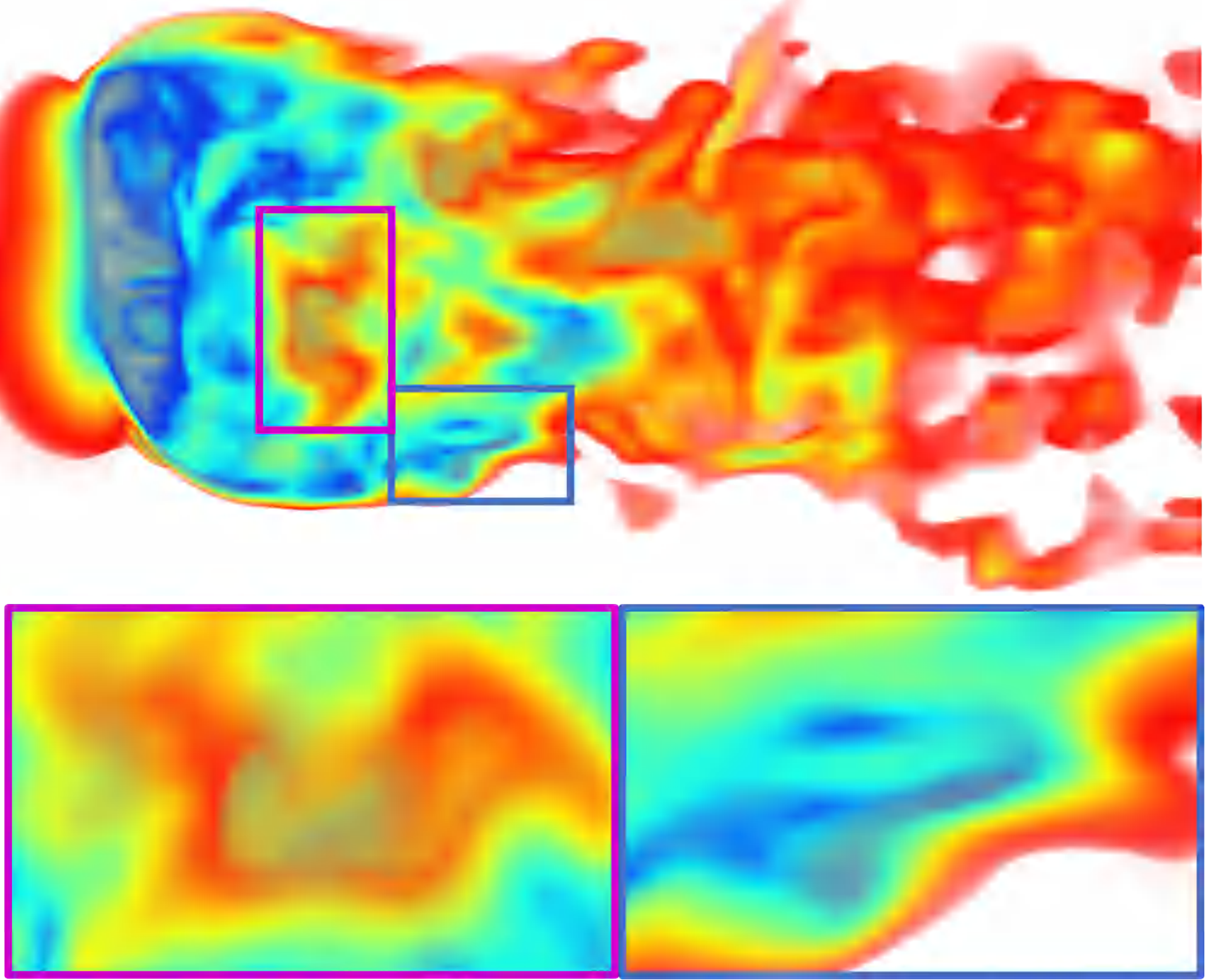} &
		\includegraphics[width=0.14\textwidth,height=0.15\textheight,keepaspectratio]{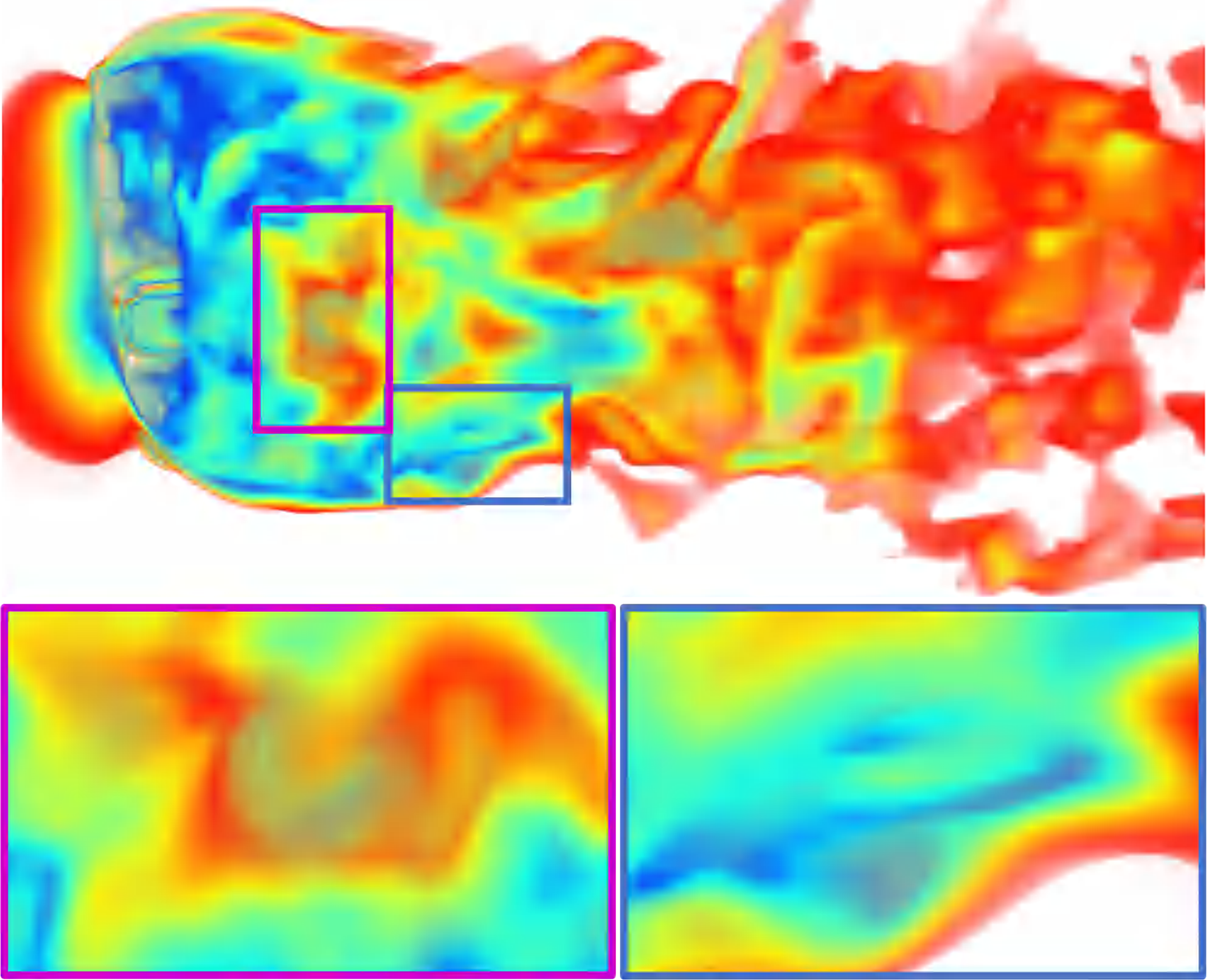} \\
		\includegraphics[width=0.14\textwidth,height=0.15\textheight,keepaspectratio]{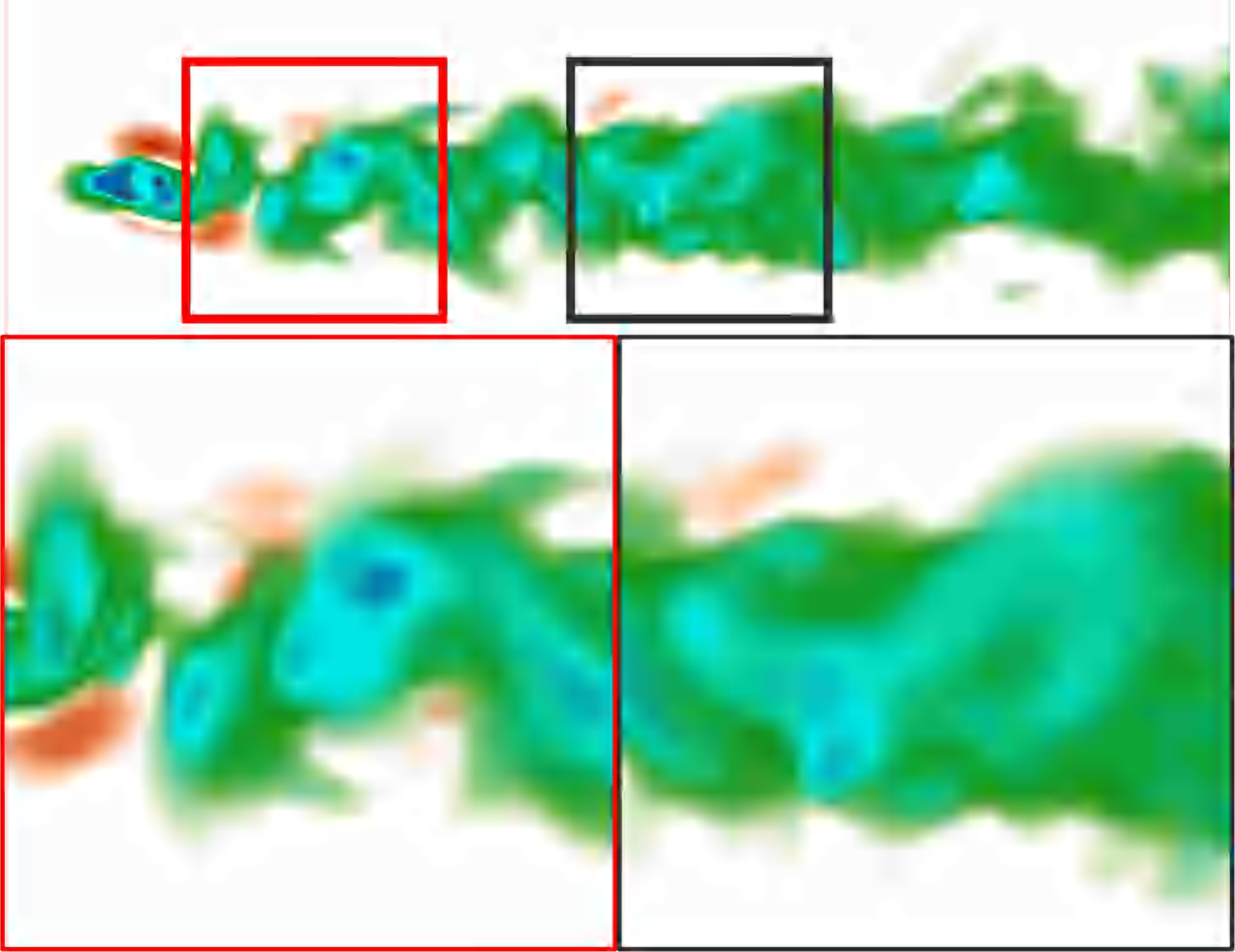} &
		\includegraphics[width=0.14\textwidth,height=0.15\textheight,keepaspectratio]{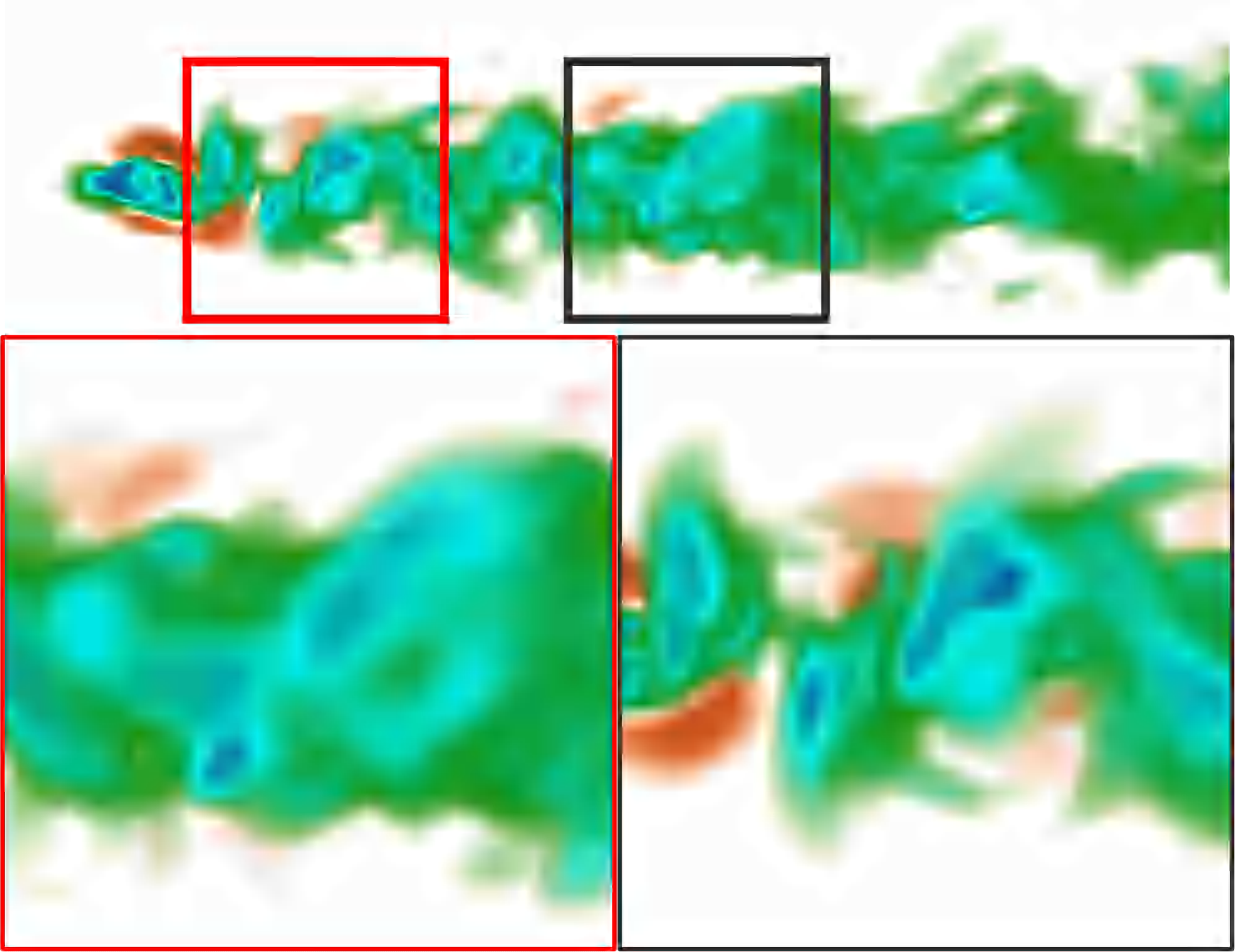} &
		\includegraphics[width=0.14\textwidth,height=0.15\textheight,keepaspectratio]{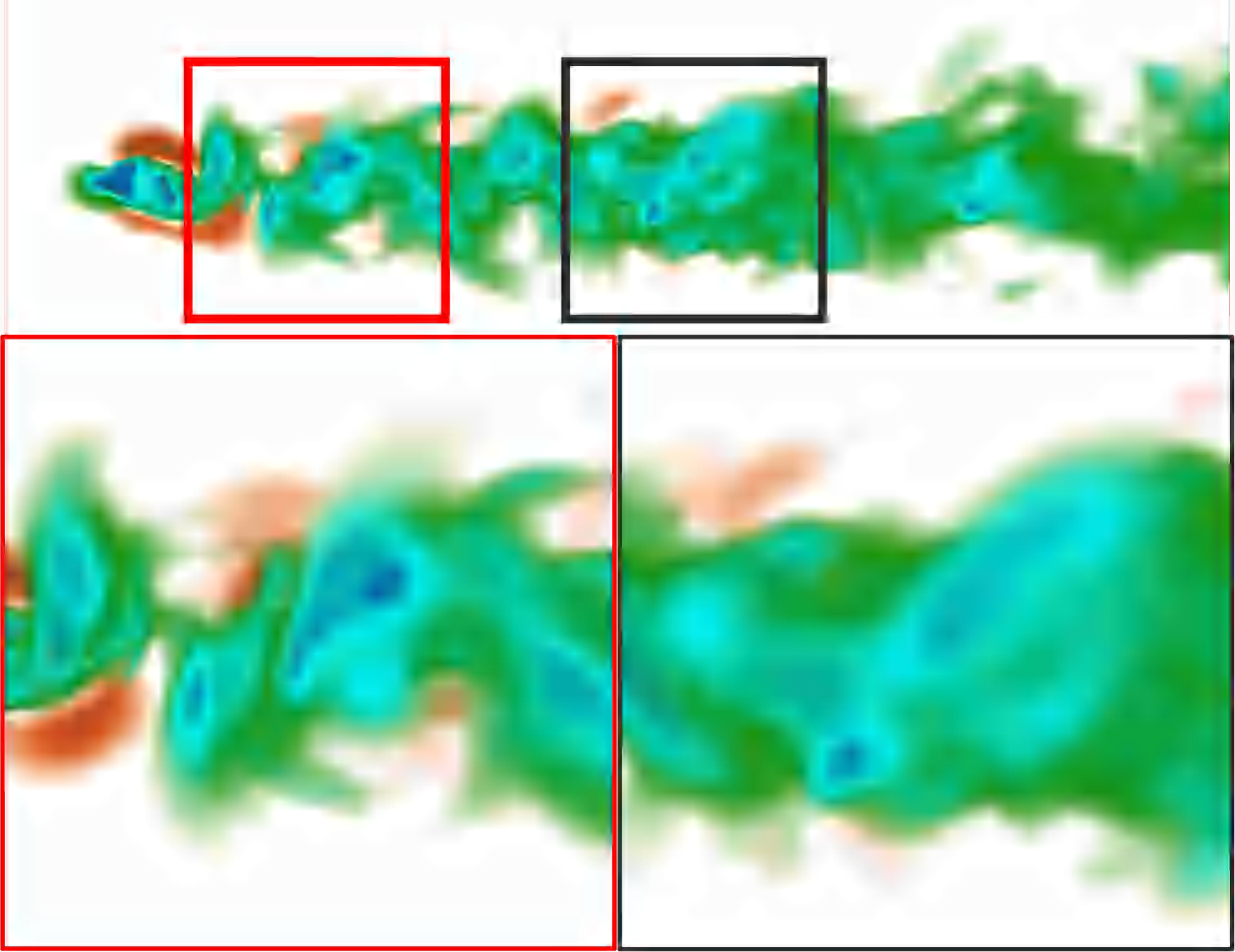} &
		\includegraphics[width=0.14\textwidth,height=0.15\textheight,keepaspectratio]{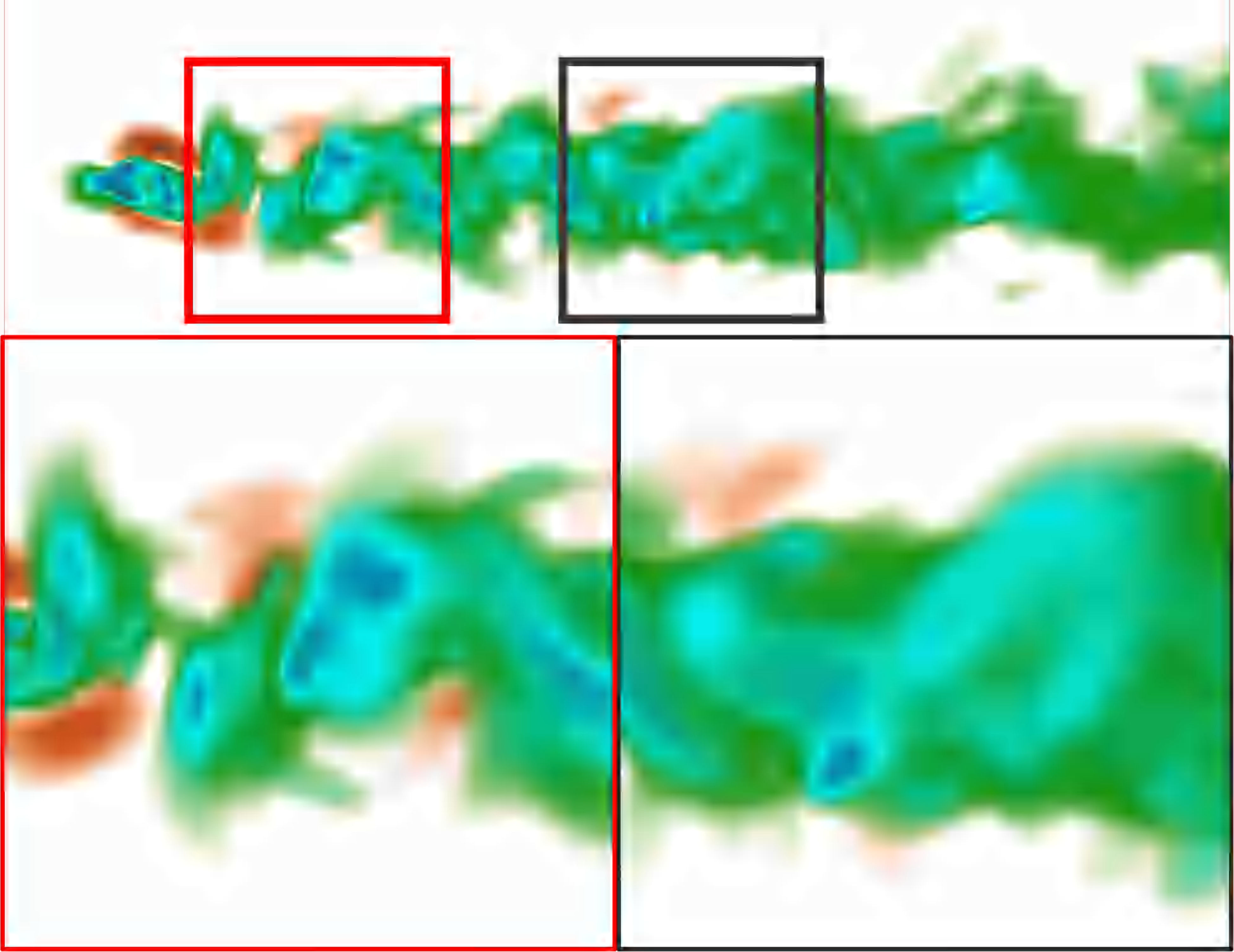} &
		\includegraphics[width=0.14\textwidth,height=0.15\textheight,keepaspectratio]{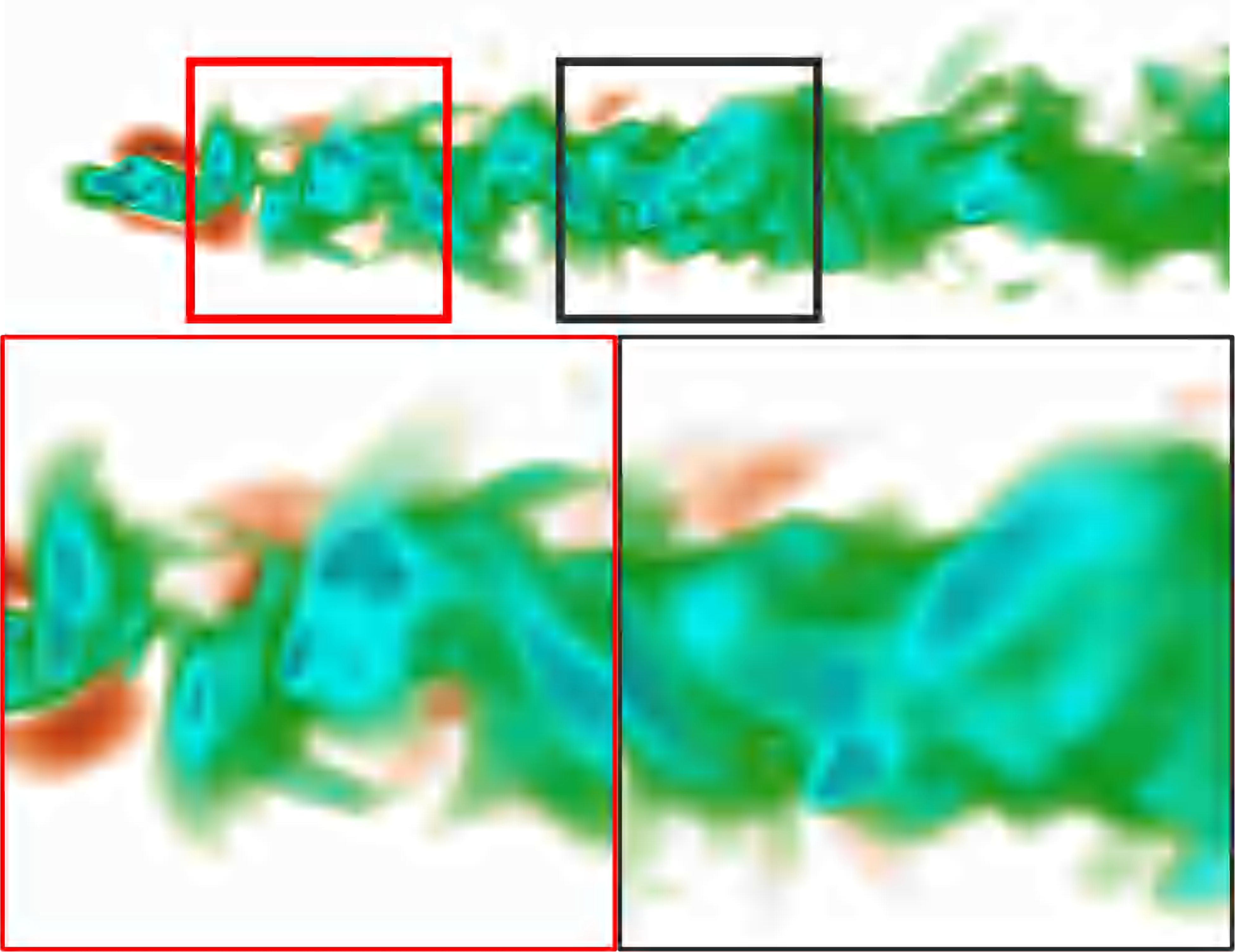} &
		\includegraphics[width=0.14\textwidth,height=0.15\textheight,keepaspectratio]{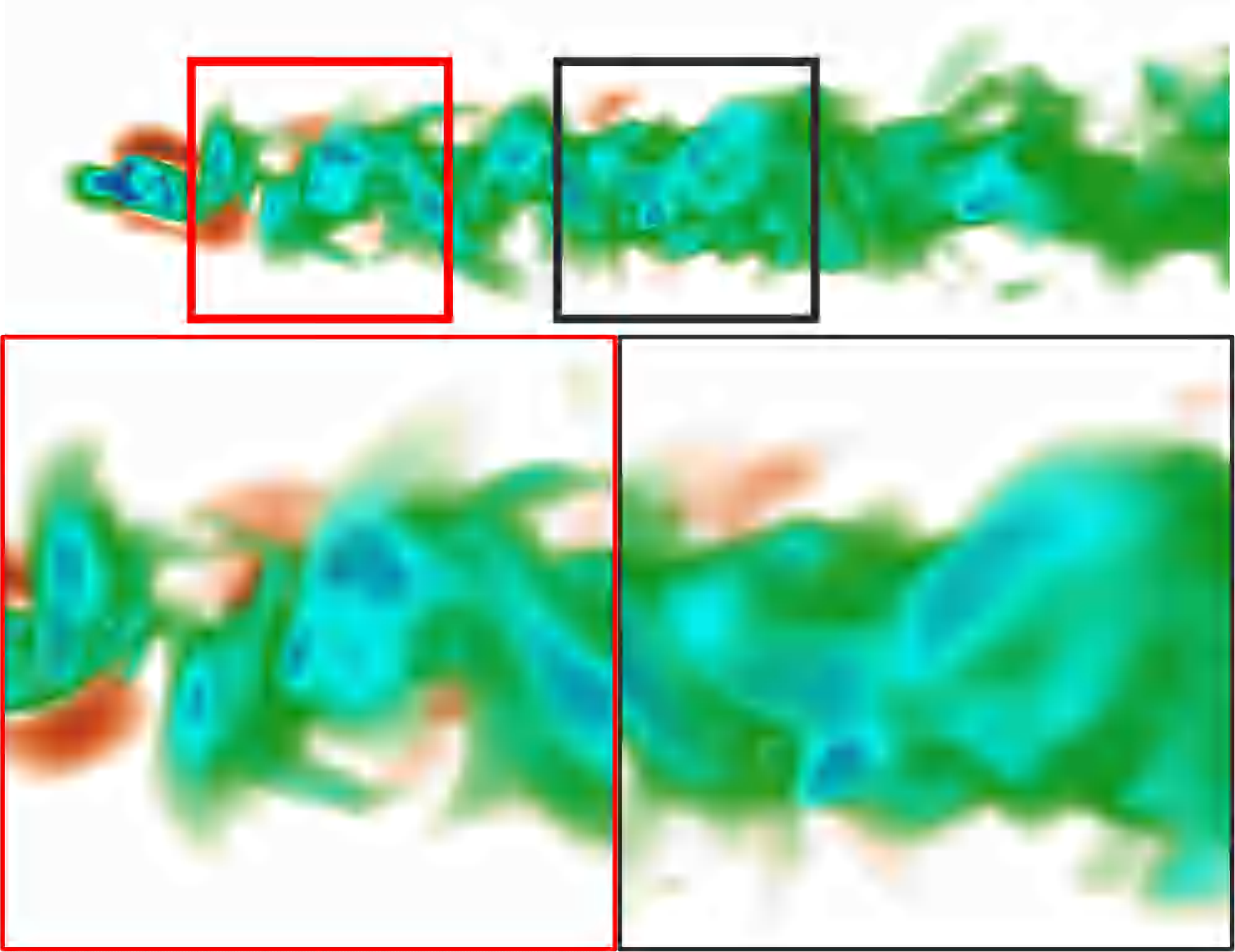} \\
		\includegraphics[width=0.14\textwidth,height=0.15\textheight,keepaspectratio]{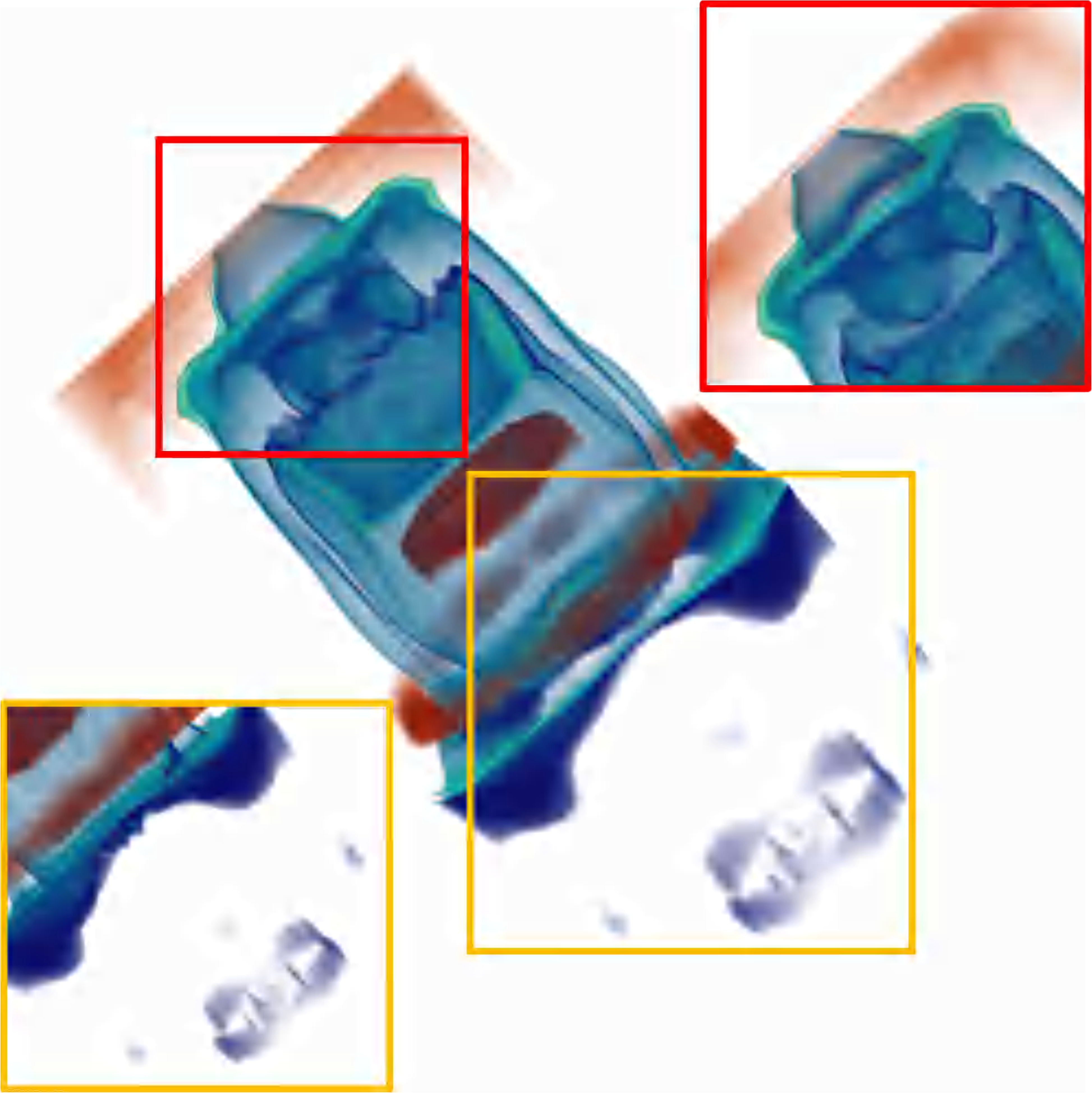} &
		\includegraphics[width=0.14\textwidth,height=0.15\textheight,keepaspectratio]{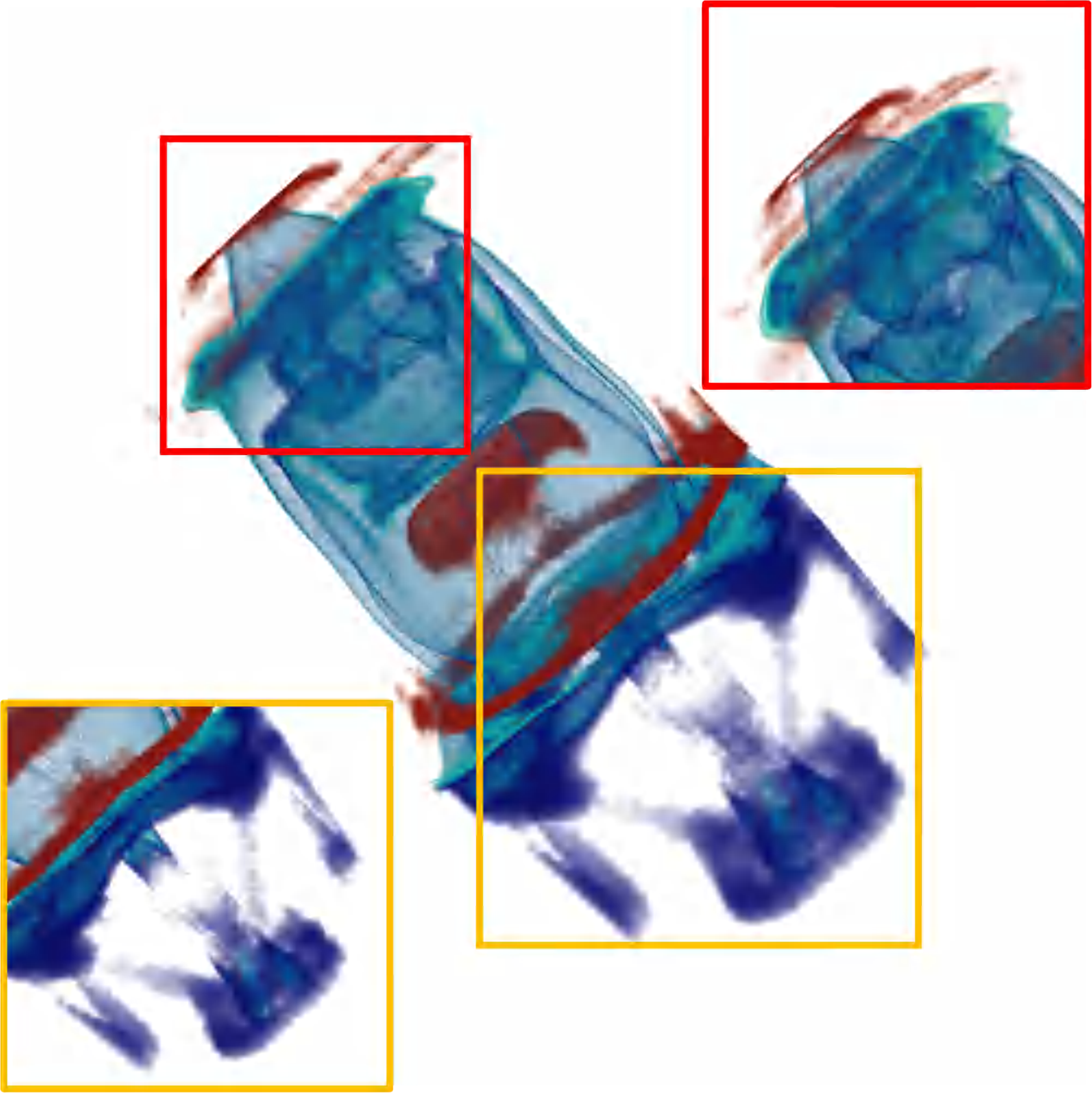} &
		\includegraphics[width=0.14\textwidth,height=0.15\textheight,keepaspectratio]{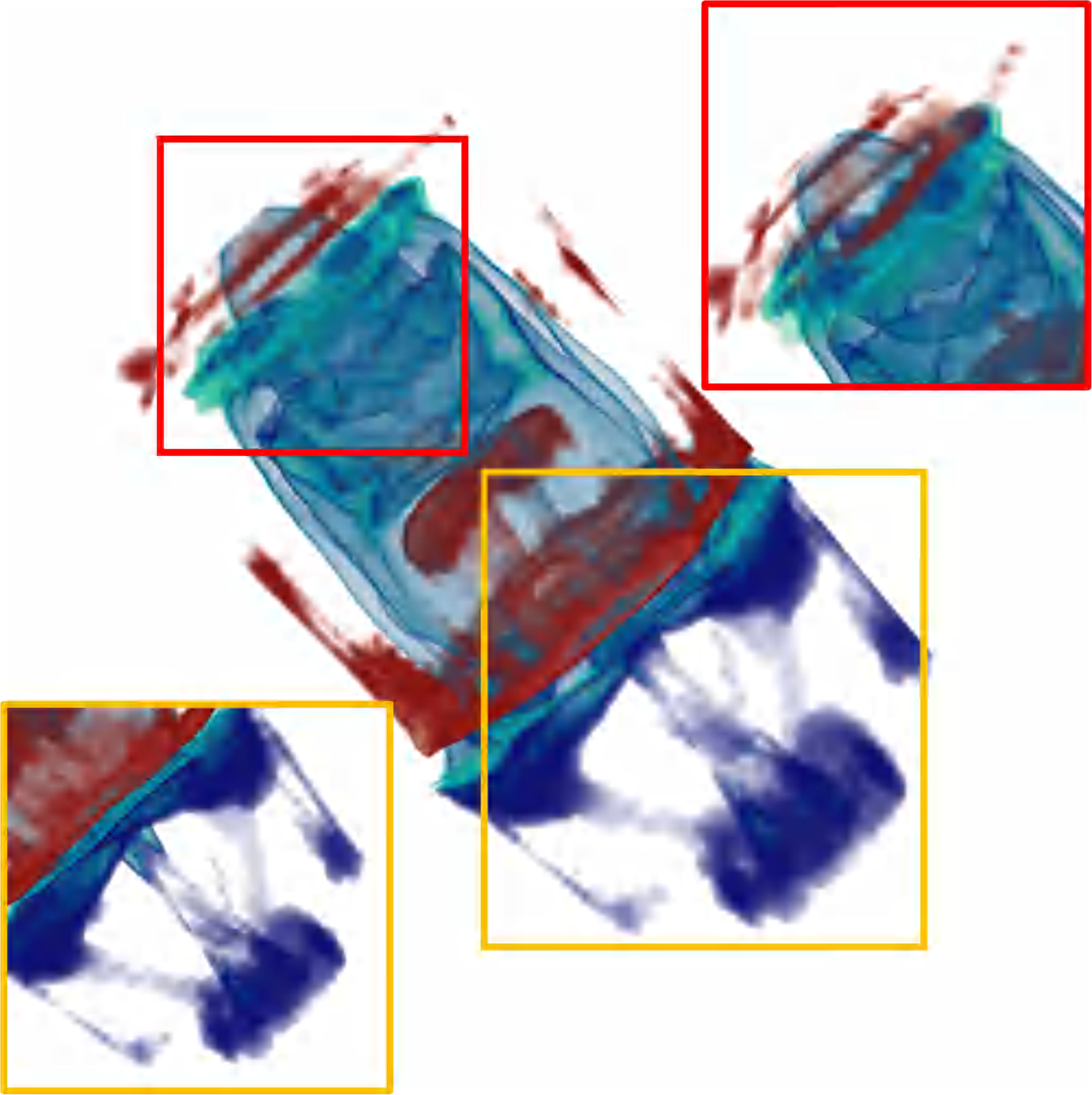} &
		\includegraphics[width=0.14\textwidth,height=0.15\textheight,keepaspectratio]{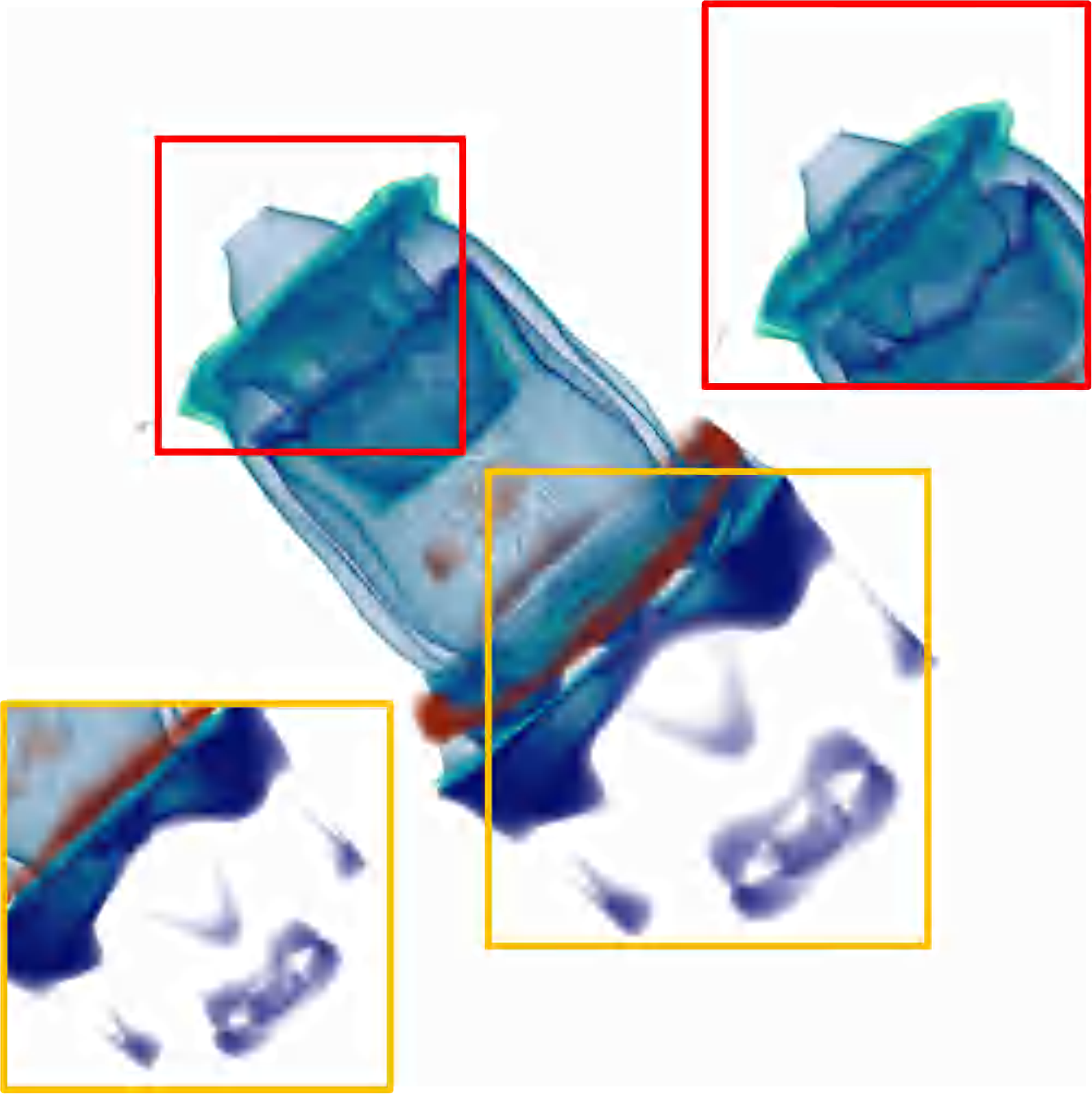} &
		\includegraphics[width=0.14\textwidth,height=0.15\textheight,keepaspectratio]{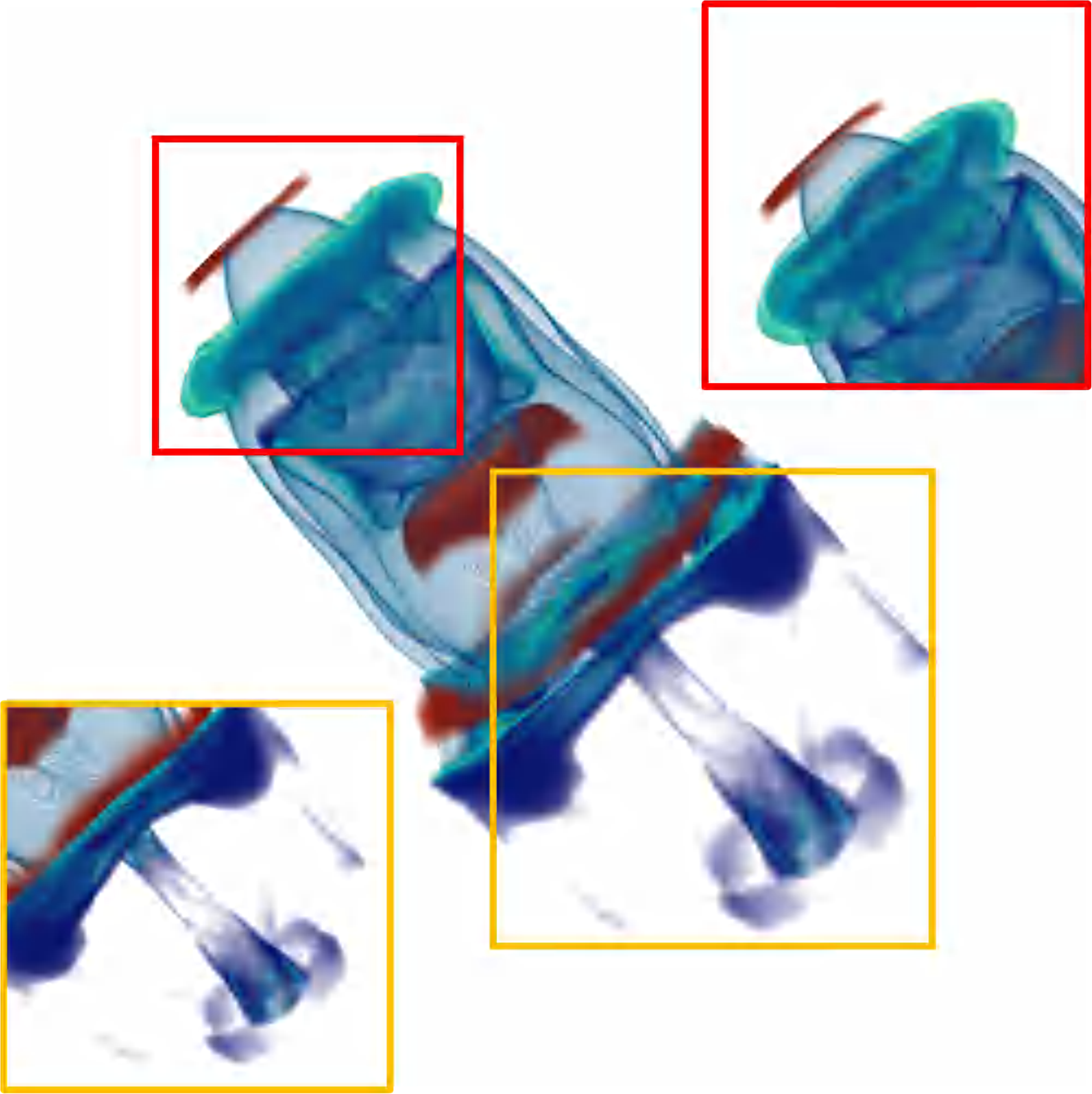} &
		\includegraphics[width=0.14\textwidth,height=0.15\textheight,keepaspectratio]{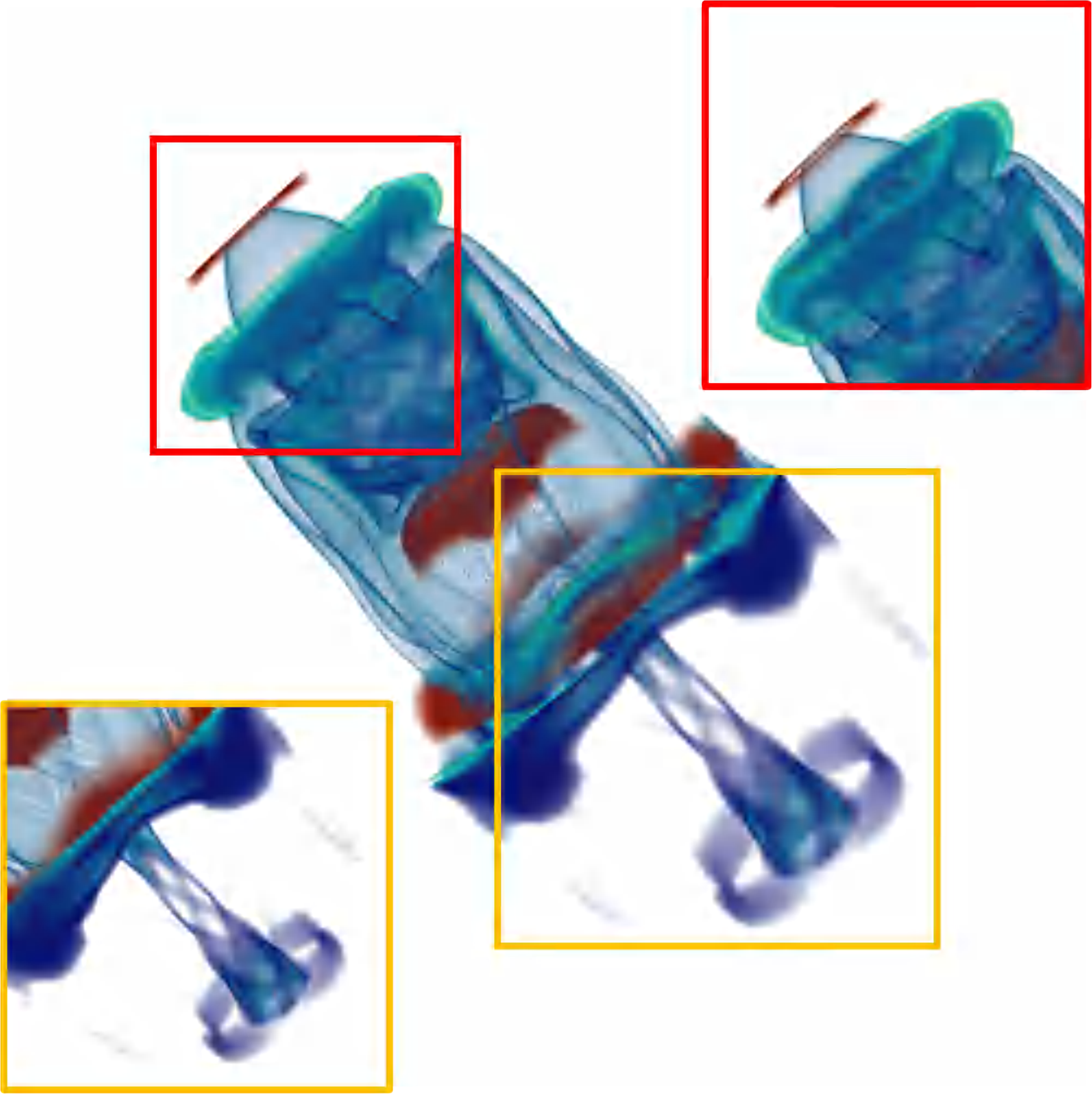} \\
		\includegraphics[width=0.14\textwidth,height=0.15\textheight,keepaspectratio]{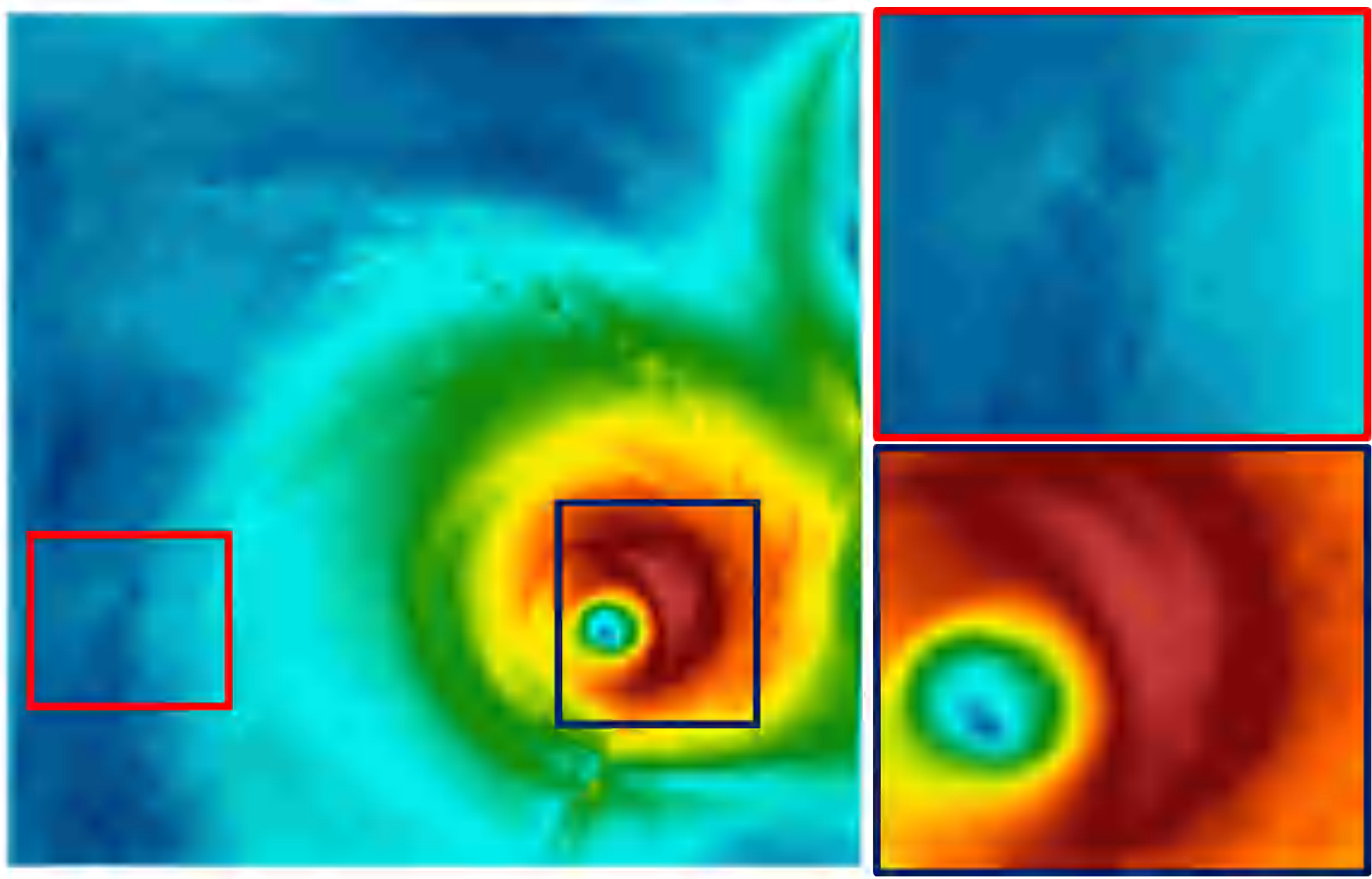} &
		\includegraphics[width=0.14\textwidth,height=0.15\textheight,keepaspectratio]{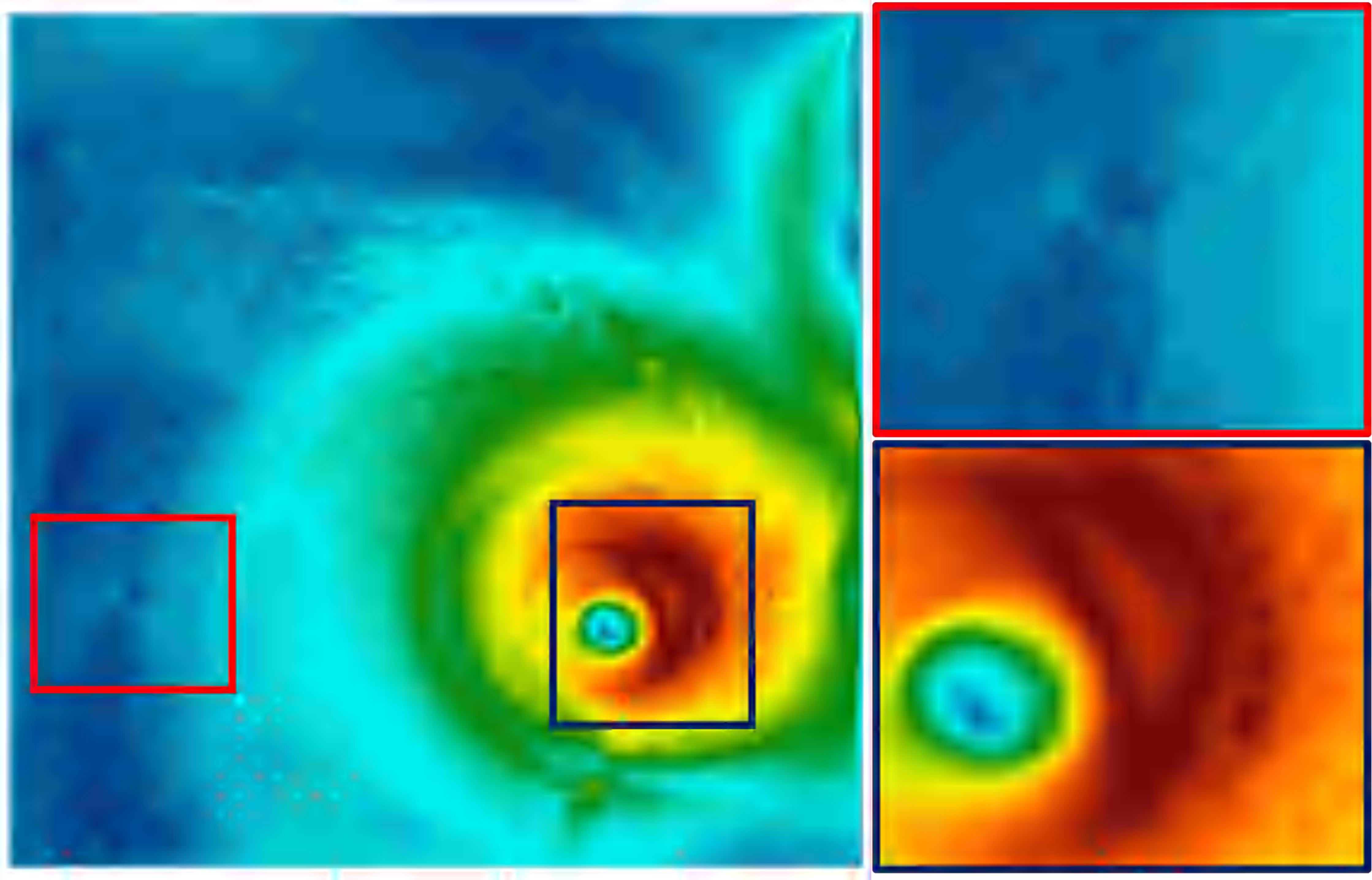} &
		\includegraphics[width=0.14\textwidth,height=0.15\textheight,keepaspectratio]{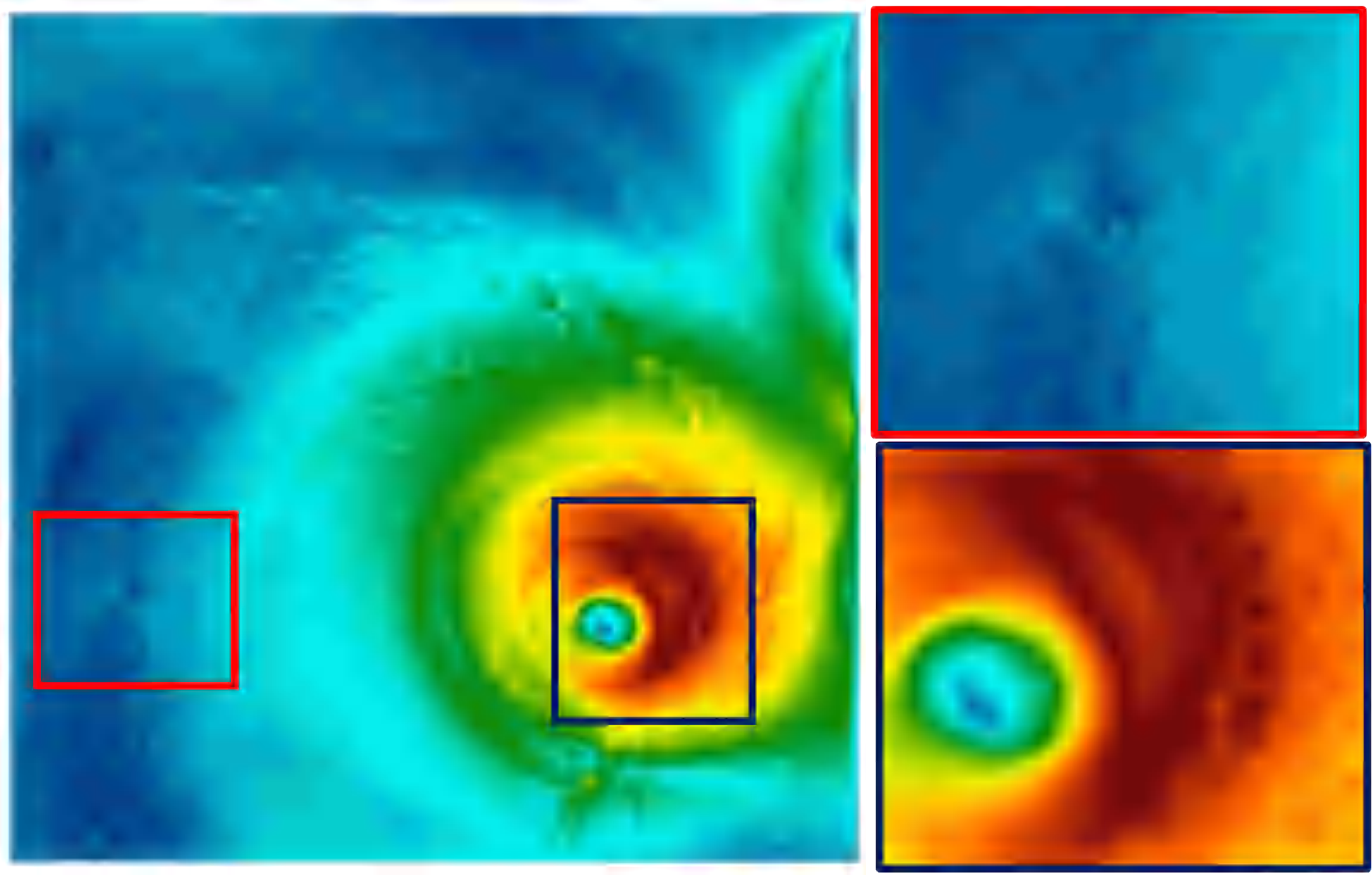} &
		\includegraphics[width=0.14\textwidth,height=0.15\textheight,keepaspectratio]{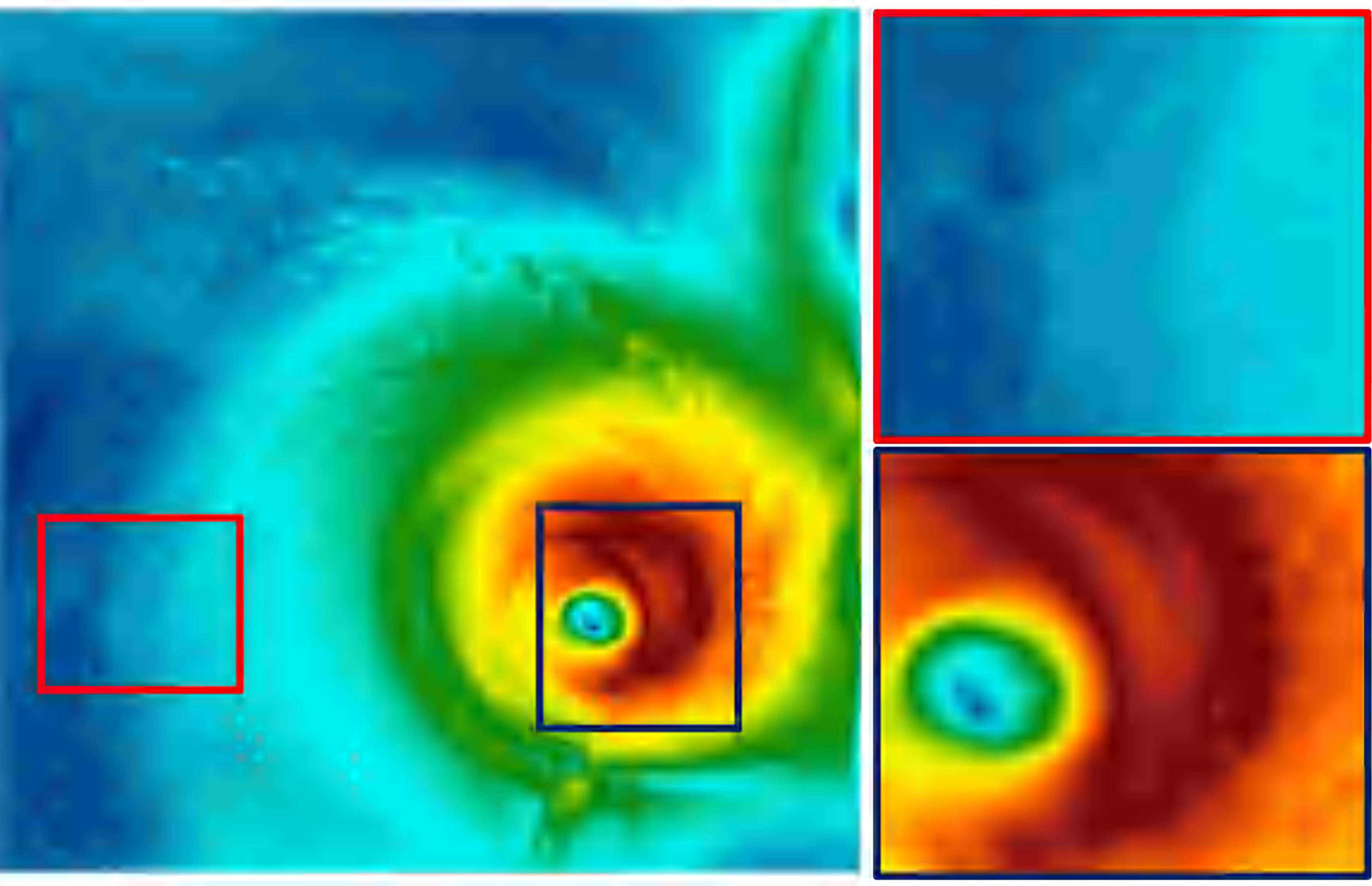} &
		\includegraphics[width=0.14\textwidth,height=0.15\textheight,keepaspectratio]{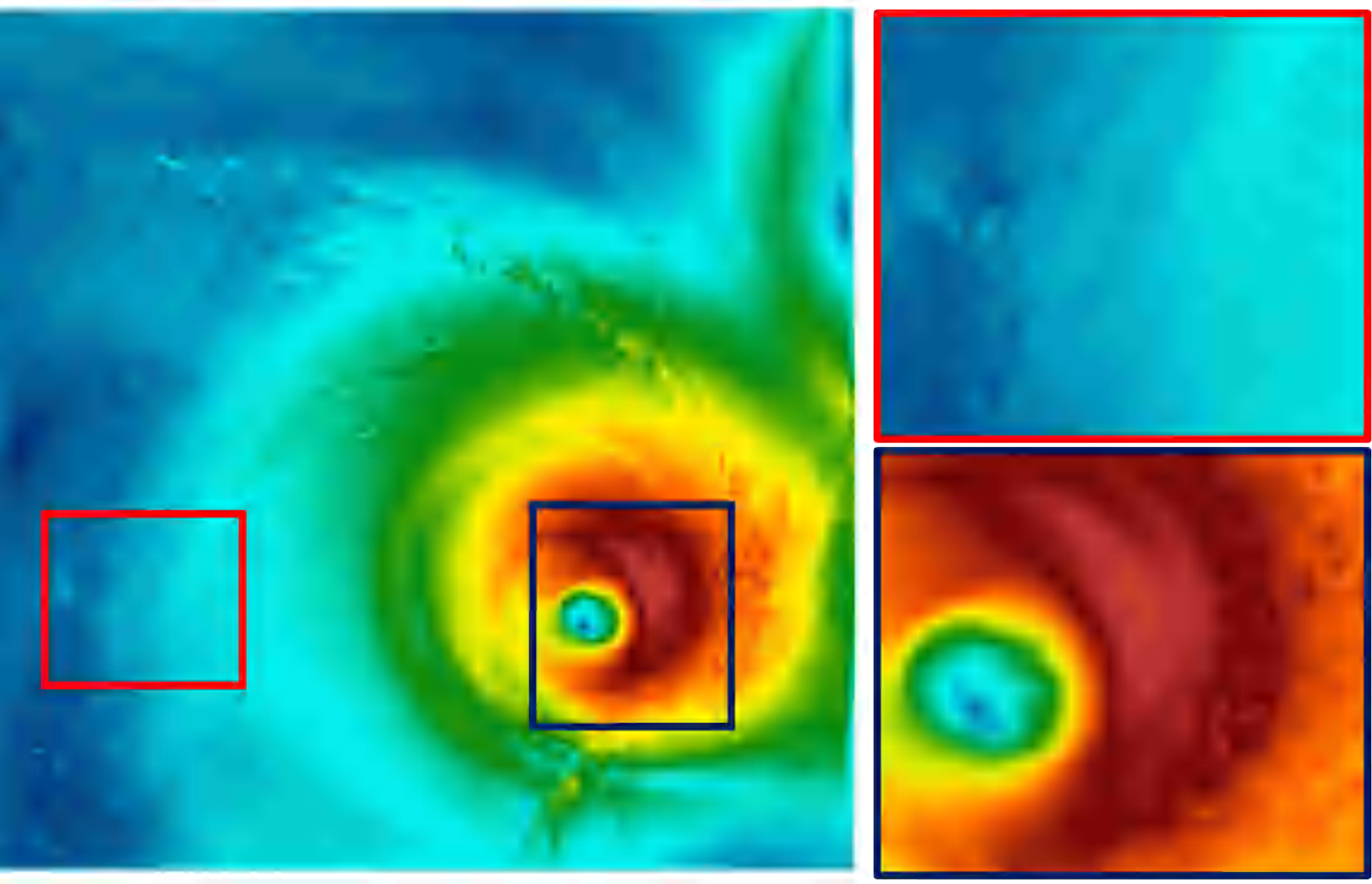} &
		\includegraphics[width=0.14\textwidth,height=0.15\textheight,keepaspectratio]{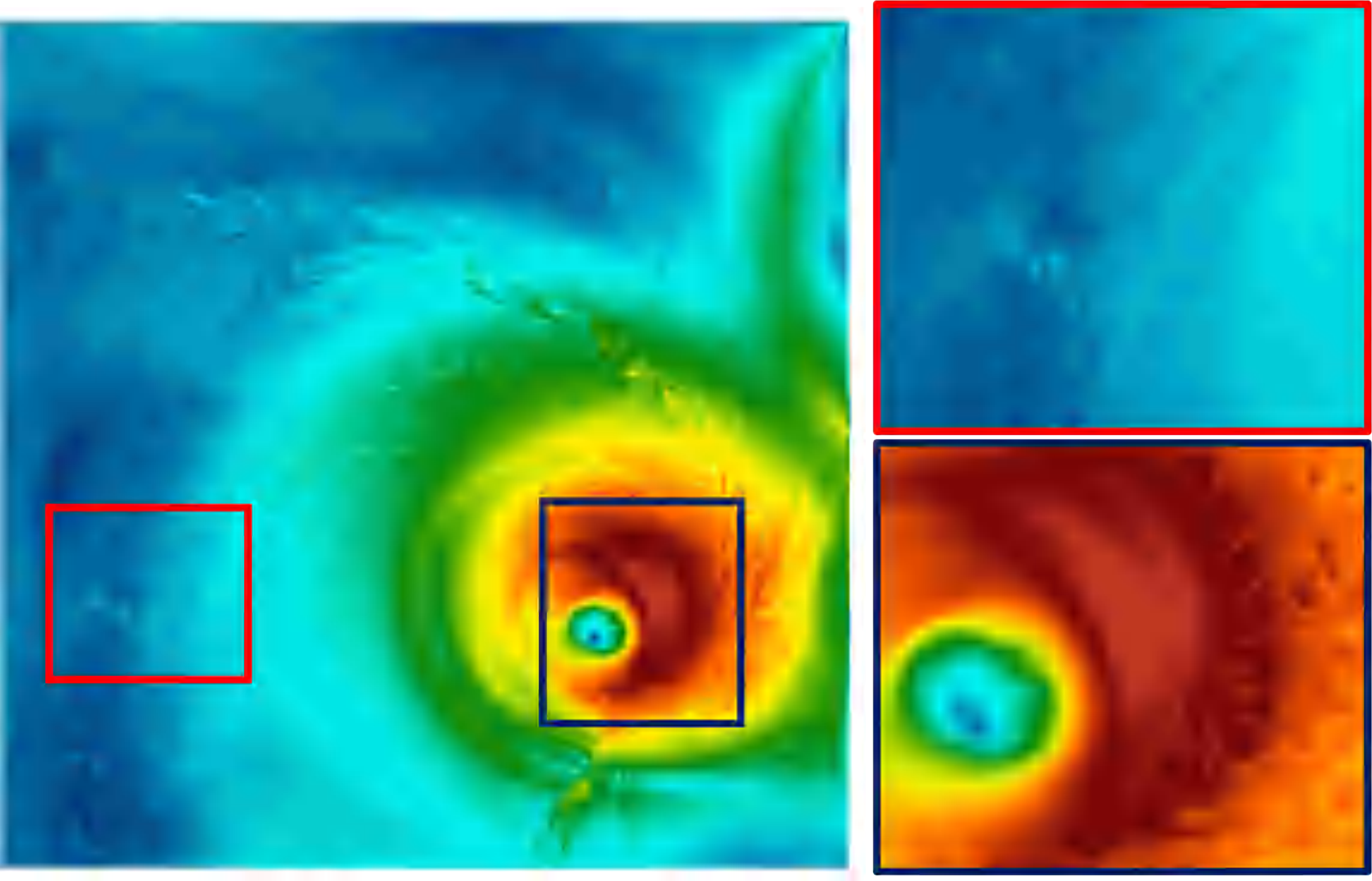} \\
		(a) TI & (b) SRGAN & (c) SSR-VFD & (d) PSRFlow & (e) CD-TVD & (f) GT \\
		
	\end{tabular}
	\vspace{-.15in} 
	\caption{Comparison of volume rendering results. Top to bottom: Research Vessel Tangaroa, Half Cylinder Ensemble, Shock Interaction Vortex, Hurricane.}
	\label{fig:volume-rendering}
\end{figure*}

In this study, we evaluated the effectiveness of our proposed method using four distinct datasets and compared it with existing baseline methods. Here, we provided a detailed description of these datasets in our study.

\textbf{Research Vessel Tangaroa}: A simulation of incompressible three-dimensional flow around the “Tangaroa” research vessel. The data resolution is $300\times 180\times 120$ with 201 timesteps. We selected this dataset to test our method on complex and large-scale flow structures over time.

\textbf{Half Cylinder Ensemble}: A three-dimensional flow simulation of a half cylinder using Gerris. We focus on the case with a Reynolds number of 6400, and the data is resampled onto a regular grid. This dataset captures flows with varying turbulence levels, providing a rigorous test of our approach’s ability to handle turbulent features.

\textbf{Shock Interaction Vortex}: A numerical simulation capturing the interaction between shock waves and longitudinal vortices. It has a grid resolution of $160\times 80\times 80$. The resulting multi-spiral vortex structures and turbulent tail region highlight our method’s capability to handle complex shock-vortex dynamics.

\textbf{Hurricane}: A large-scale atmospheric simulation from the National Center for Atmospheric Research. The resolution of this data set is $500\times 500\times 100$ and encompasses multiple time-varying scalar and vector variables. Considering its broad dynamic range and data volume, we performed normalization preprocessing, making it an ideal benchmark for testing scalability and robustness.

The training process was conducted on a single NVIDIA A40 GPU. We derived LR vector fields from HR fields by trilinear down-sampling, and used Adam optimizer \cite{kingma2014adam} with a learning rate of \(1 \times 10^{-4}\) to update the model parameters. To maintain consistency, all rendered results within the same dataset were produced under identical settings. For each dataset, we traced 200 streamlines when visualizing the reconstructed results, ensuring a comprehensive depiction of the flow field.

Additionally, the CD-TVD model was pre-trained on the Half Cylinder Ensemble dataset, which includes simulations at three distinct Reynolds numbers of 160, 320, and 640. Each Reynolds number simulation consists of approximately $150$ timesteps, resulting in a total of approximately $450$ timesteps combined across the dataset. During pre-training, the data were randomly split into training and testing sets with an 80\% and 20\% ratio, respectively, ensuring a robust evaluation and model generalization capability.

\begin{table}[b]
	\centering
	\caption{
		Effect of hyper-parameter $\beta$ on PSNR performance.
	}
	\label{table_beta}
	\begin{tabular}{c|cccccc}
		\hline
		$\beta$ & 0.01 & 0.1 & 0.25 & 0.5 & 1 & 5 \\[0.5ex]
		\hline
		PSNR & 44.09 & \textbf{45.15} & 45.11 & 45.04 & 44.79 & 44.11 \\[0.5ex]
		\hline
	\end{tabular}
\end{table}

To determine the optimal value of the hyper-parameter $\beta$ in our loss function (Equation \ref{equ-4-6}), we performed a grid search on the Tangaroa dataset. We systematically explored a range of values and evaluated the reconstruction performance based on Peak Signal-to-Noise Ratio (PSNR). The tested values and corresponding PSNR results are summarized in Table \ref{table_beta}. Based on these results, we selected $\beta = 0.1$ as the default setting, since it achieved the highest PSNR, thus effectively balancing reconstruction accuracy and contrastive regularization.

\subsection{Baselines and Evaluation Metrics.} 
\subsubsection{Baselines}
We compared CD-TVD with four baseline methods:

\begin{itemize}
	\item \textbf{Trilinear Interpolation (TI):} Trilinear interpolation is a simple and frequently used method for scaling up data resolution.
	\item \textbf{SRGAN~\cite{8099502}:} SRGAN is a deep learning-based super-resolution method originally designed for image super-resolution. In the context of 3D vector fields, SRGAN requires more GPU memory and does not provide a classifier suitable for 3D data. Therefore, we used five residual blocks (RB) and applied perceptual losses instead of the perceptual loss used in the original implementation.
	\item \textbf{SSR-VFD~\cite{guo2020ssr}:} SSR-VFD is a deep learning-based super-resolution method designed for scientific data. Some research has found that SSR-VFD performs better without the discriminator, so we used this version for comparison. The model uses magnitude and angle losses as specified in the original work.
	\item \textbf{PSRFlow~\cite{shen2023psrflow}:} PSRFlow is a probabilistic super-resolution method that utilizes normalizing flows to model HR data from LR inputs. For this baseline, we performed two consecutive $2\times$ upscaling operations, resulting in a $4\times$ upscaling effect. This follows the structure used in the original PSRFlow paper.
\end{itemize}

Considering that our method CD-TVD specifically addresses scenarios with extremely limited HR data (only a single HR timestep for fine-tuning), we conducted the primary comparative experiments under strictly identical conditions. Specifically, each baseline method (SRGAN, SSR-VFD, PSRFlow) was trained using the same single HR timestep, selected via our entropy-based selection method (detailed in Section~\ref{sec:entropy_selection}). This ensured consistency across methods, allowing a direct and fair comparison of their capabilities under scarce HR conditions.

\begin{figure*}[tb]
	\centering
	\begin{tabular}{cccccc}
		
		\includegraphics[width=0.14\textwidth,height=0.18\textheight,keepaspectratio]{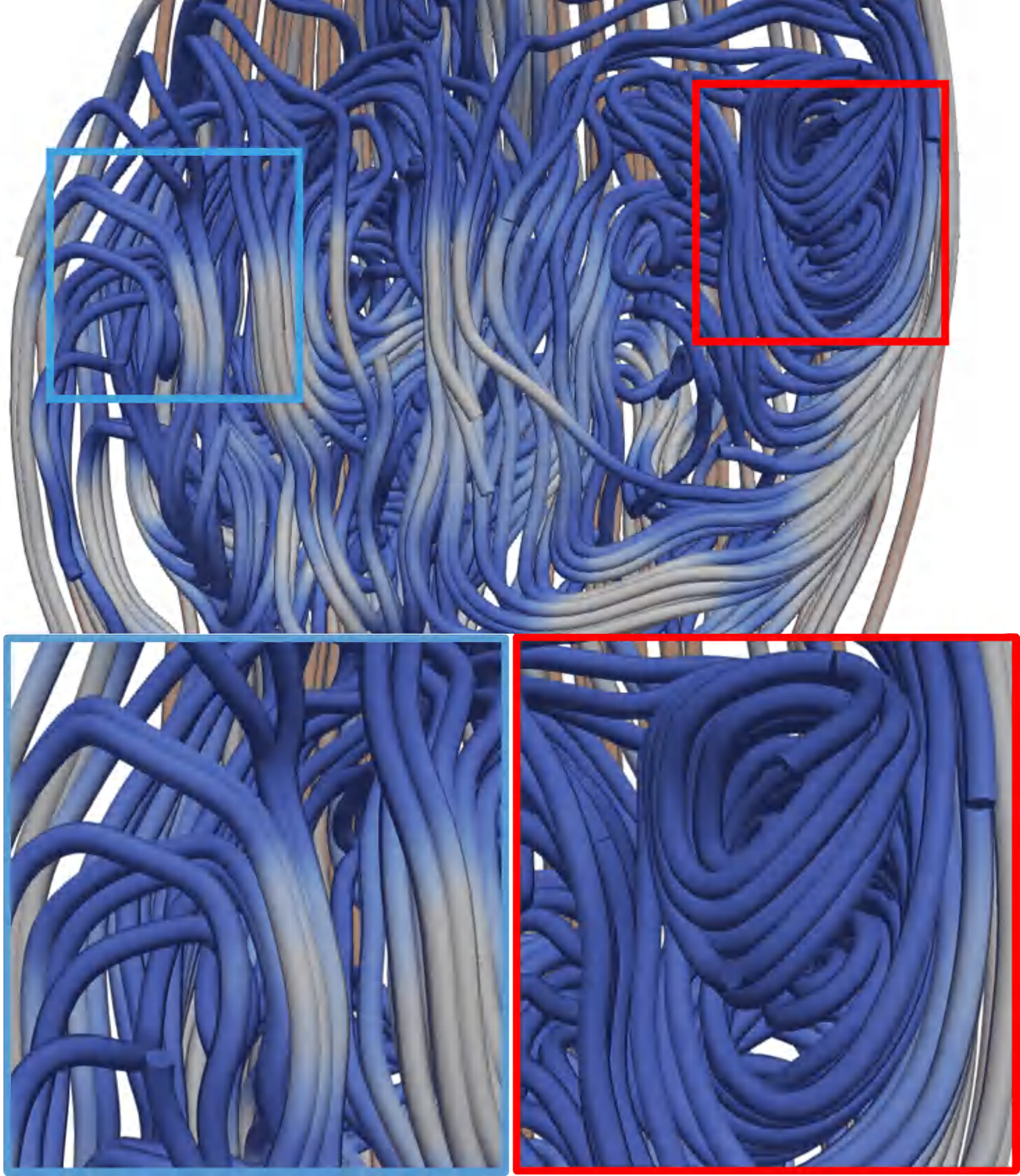} &
		\includegraphics[width=0.14\textwidth,height=0.18\textheight,keepaspectratio]{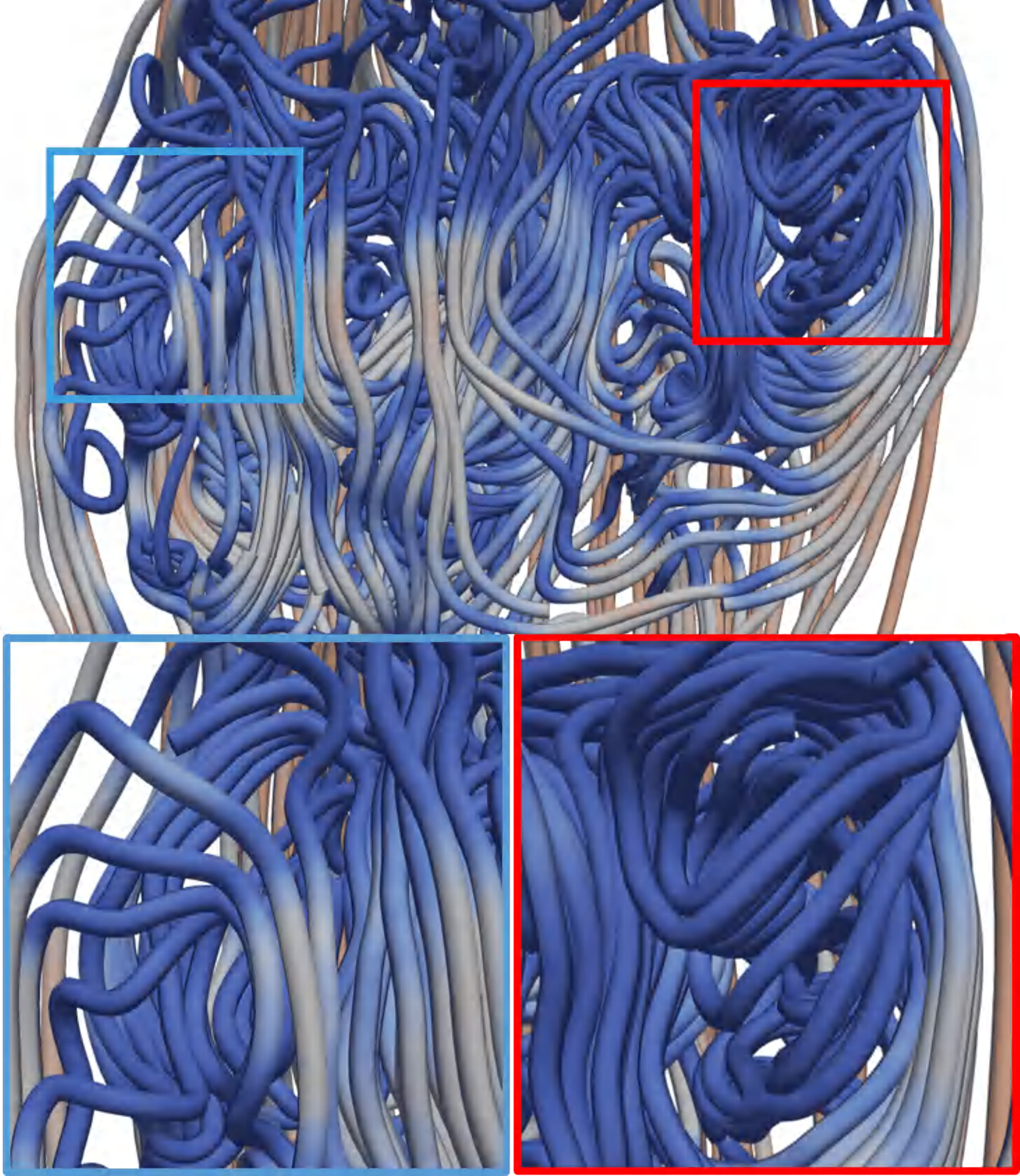} &
		\includegraphics[width=0.14\textwidth,height=0.18\textheight,keepaspectratio]{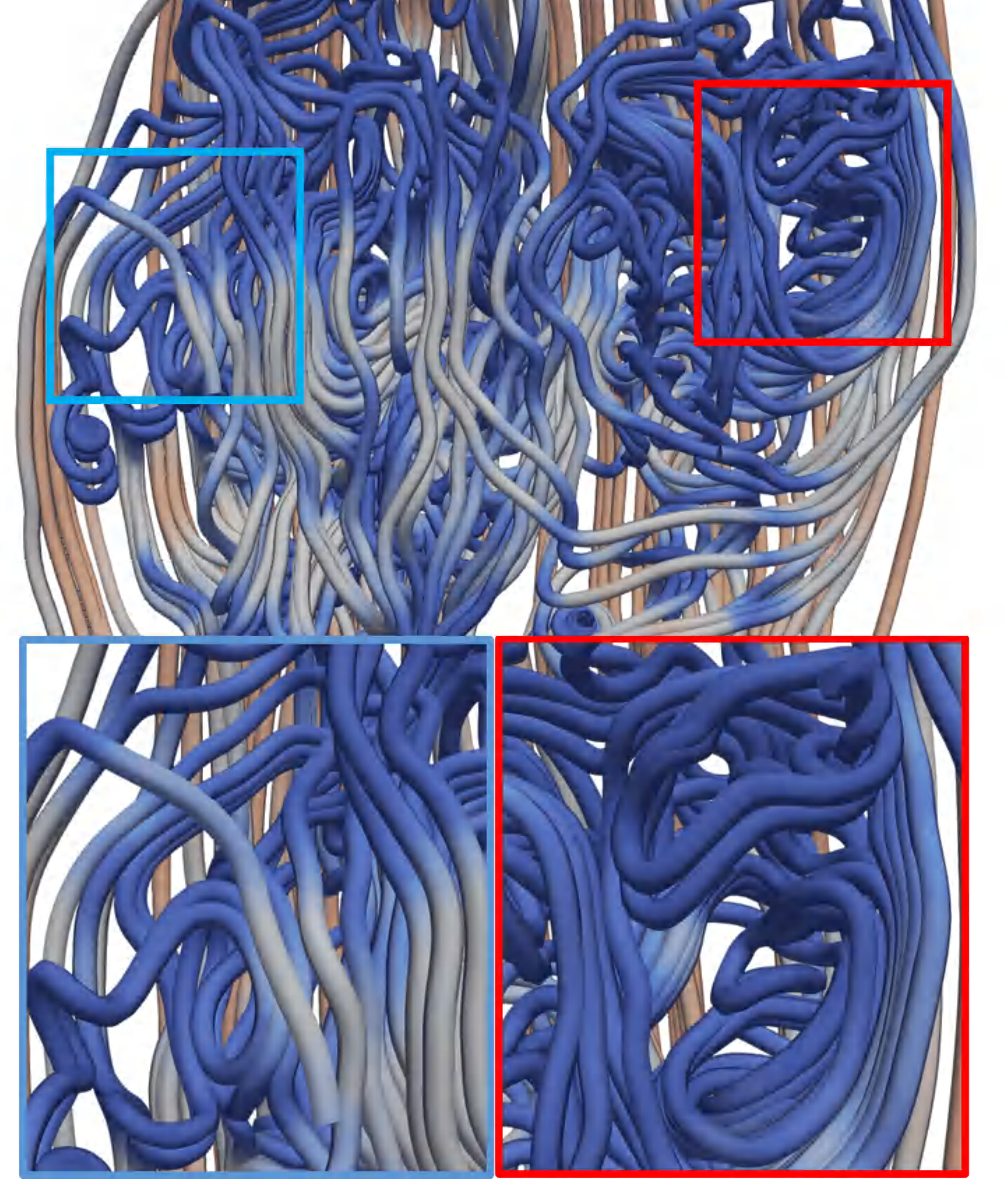} &
		\includegraphics[width=0.14\textwidth,height=0.18\textheight,keepaspectratio]{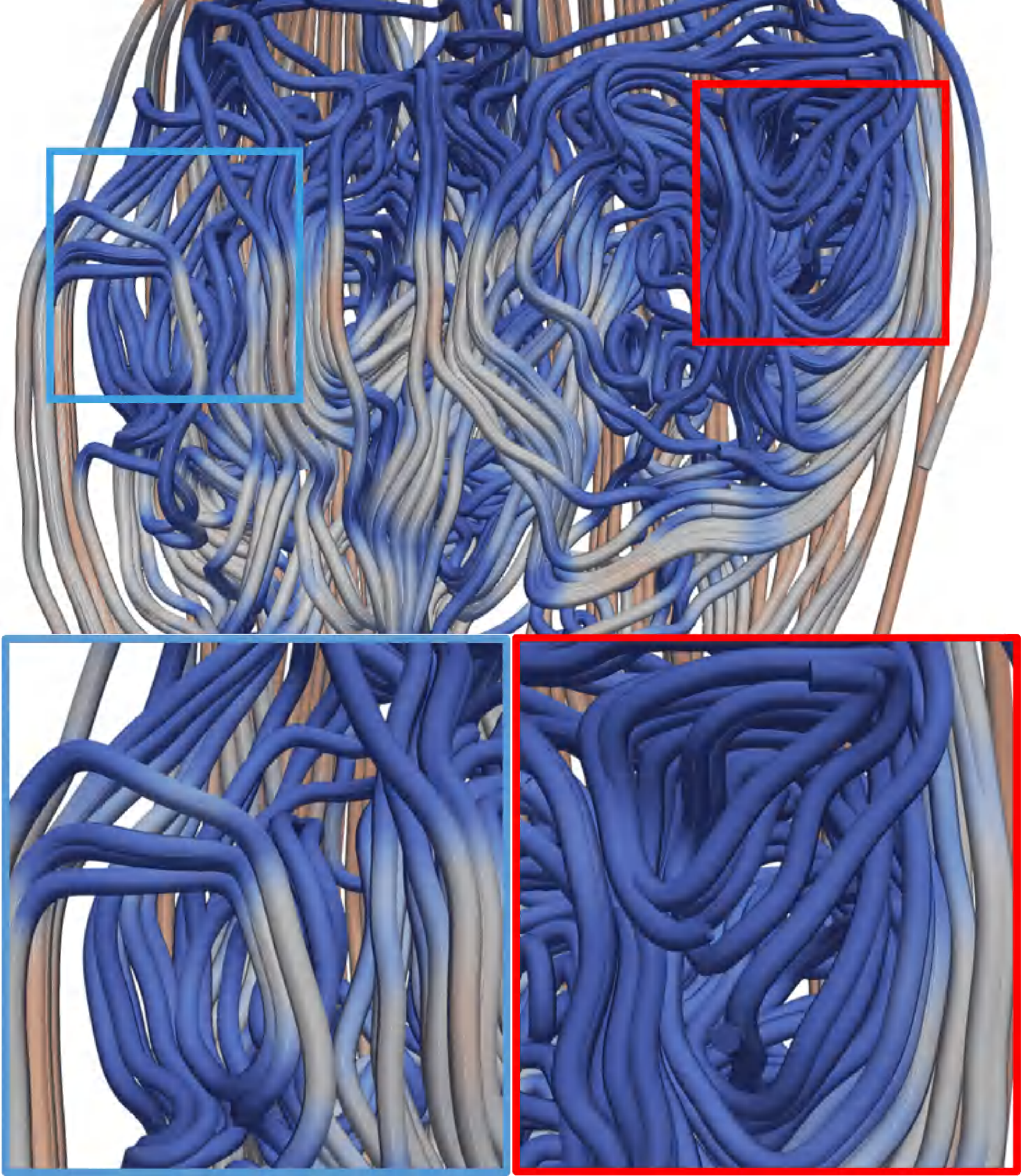} &
		\includegraphics[width=0.14\textwidth,height=0.18\textheight,keepaspectratio]{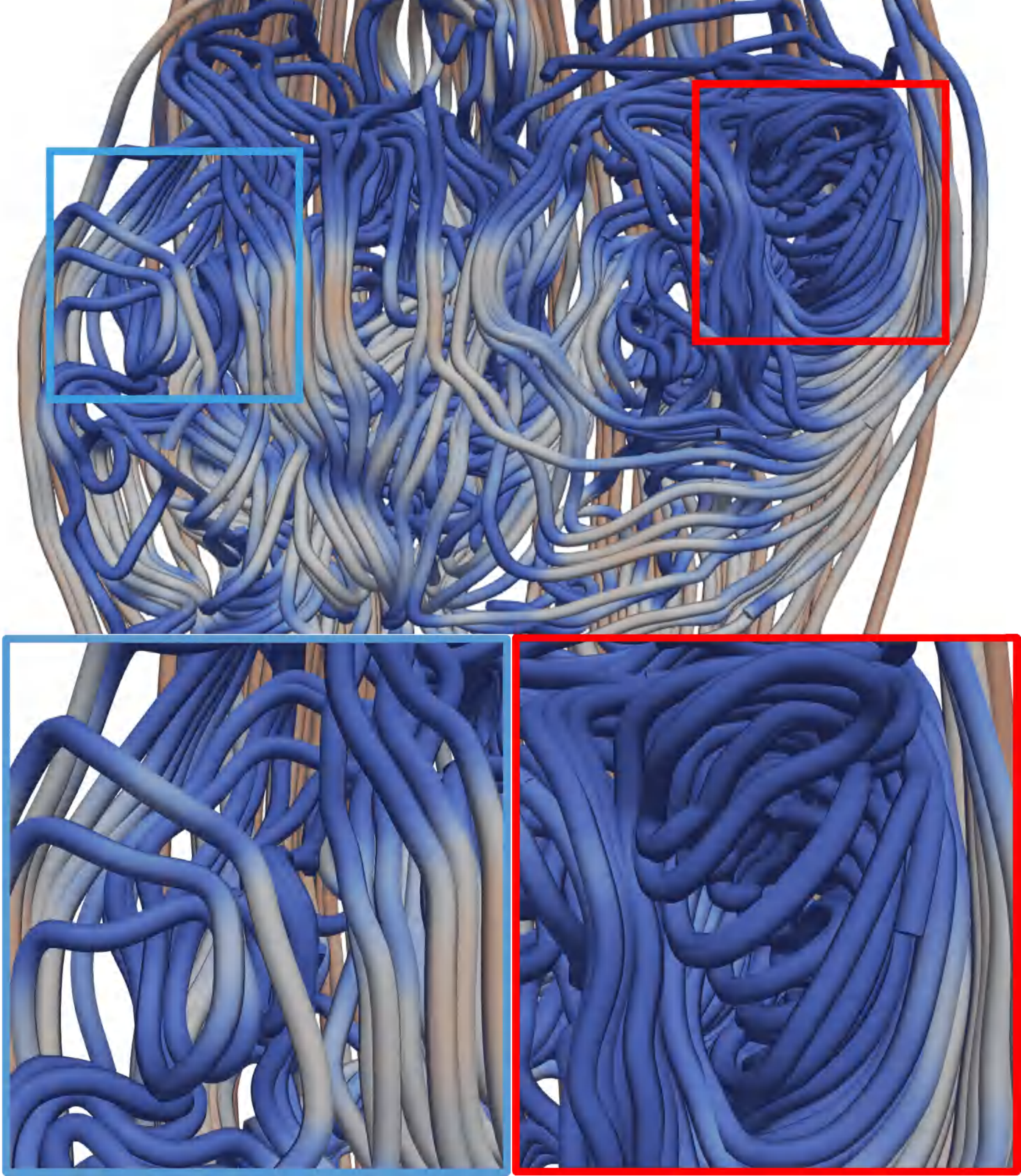} &
		\includegraphics[width=0.14\textwidth,height=0.18\textheight,keepaspectratio]{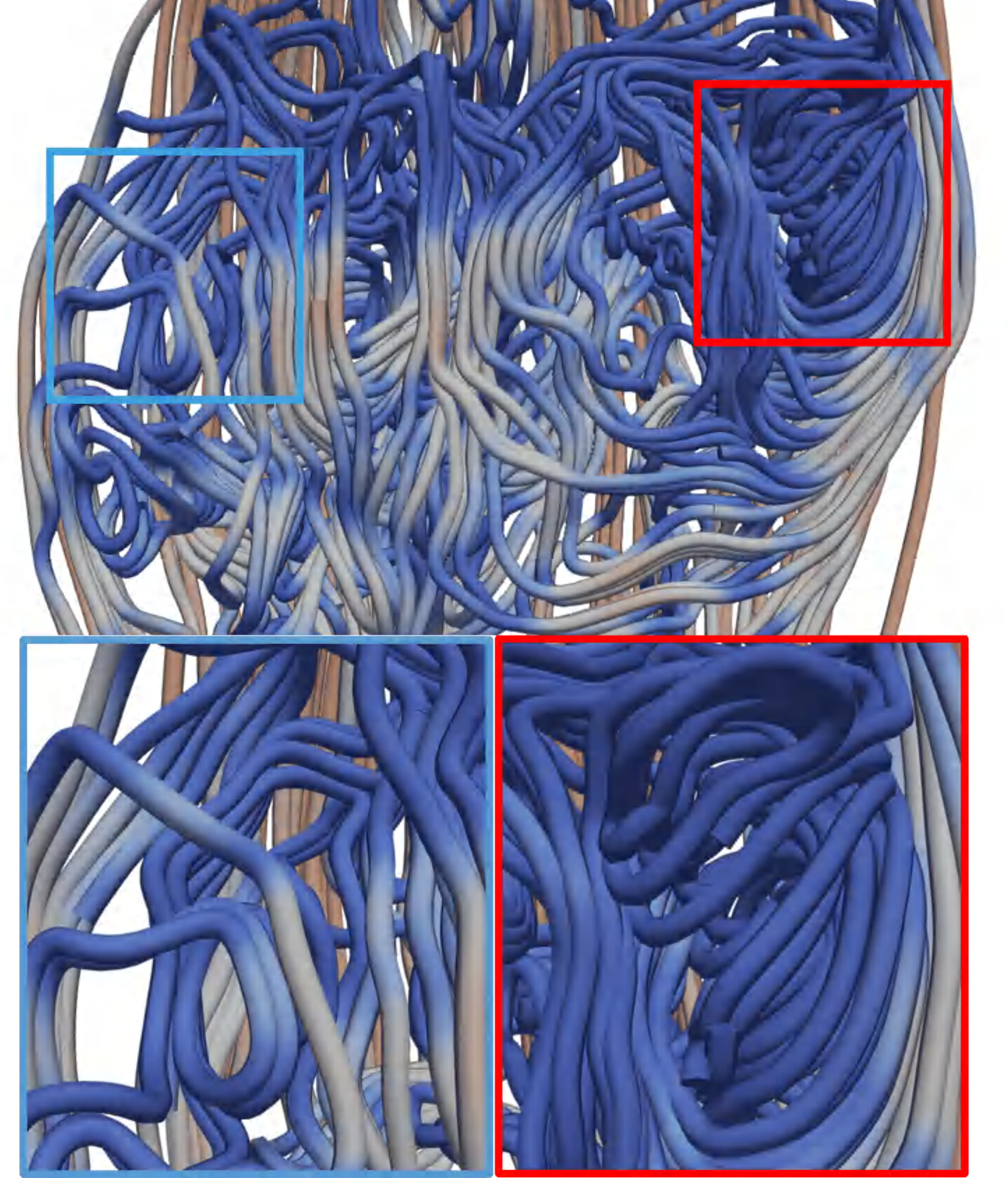} \\
		
		\includegraphics[width=0.14\textwidth,height=0.18\textheight,keepaspectratio]{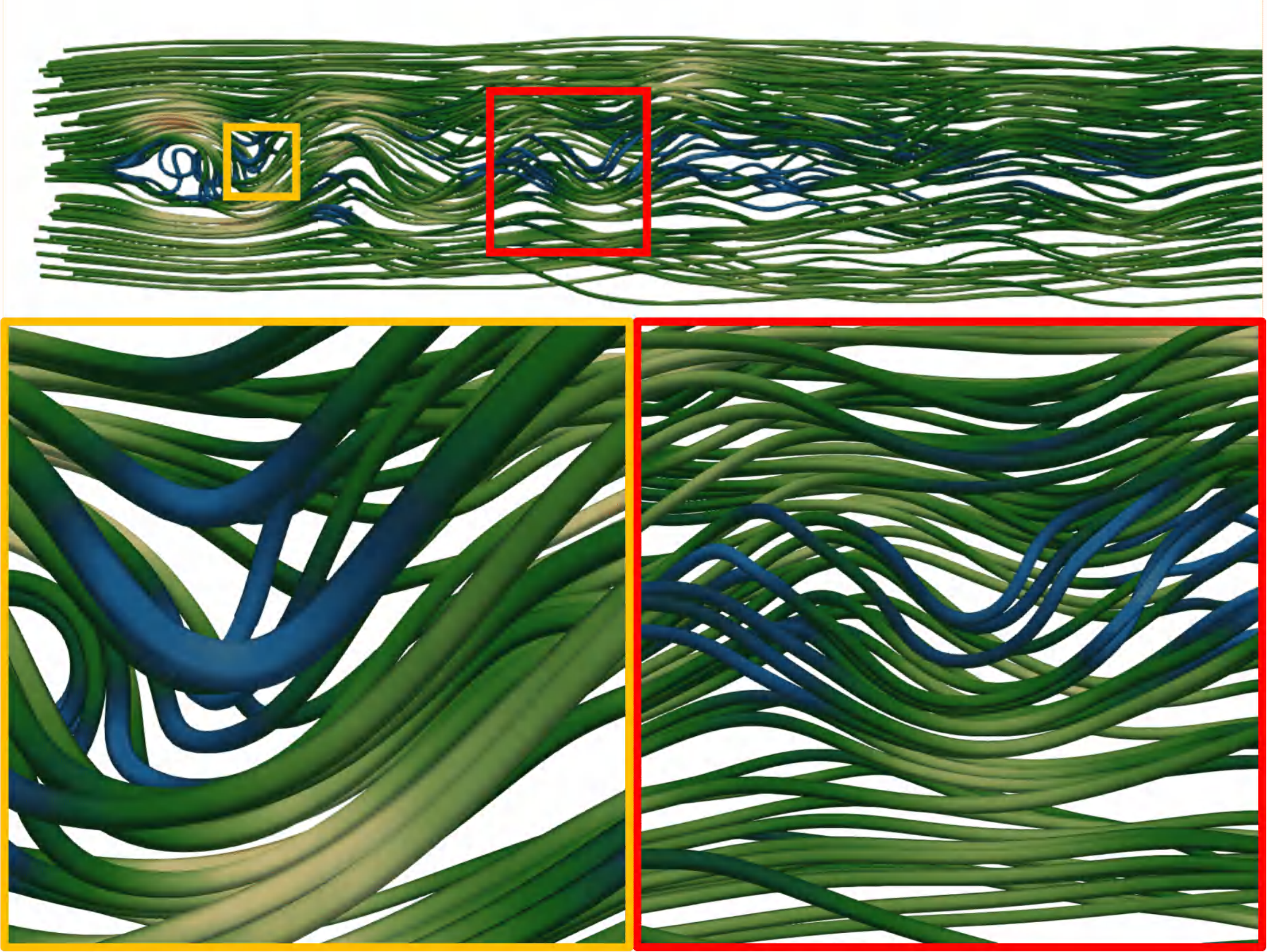} &
		\includegraphics[width=0.14\textwidth,height=0.18\textheight,keepaspectratio]{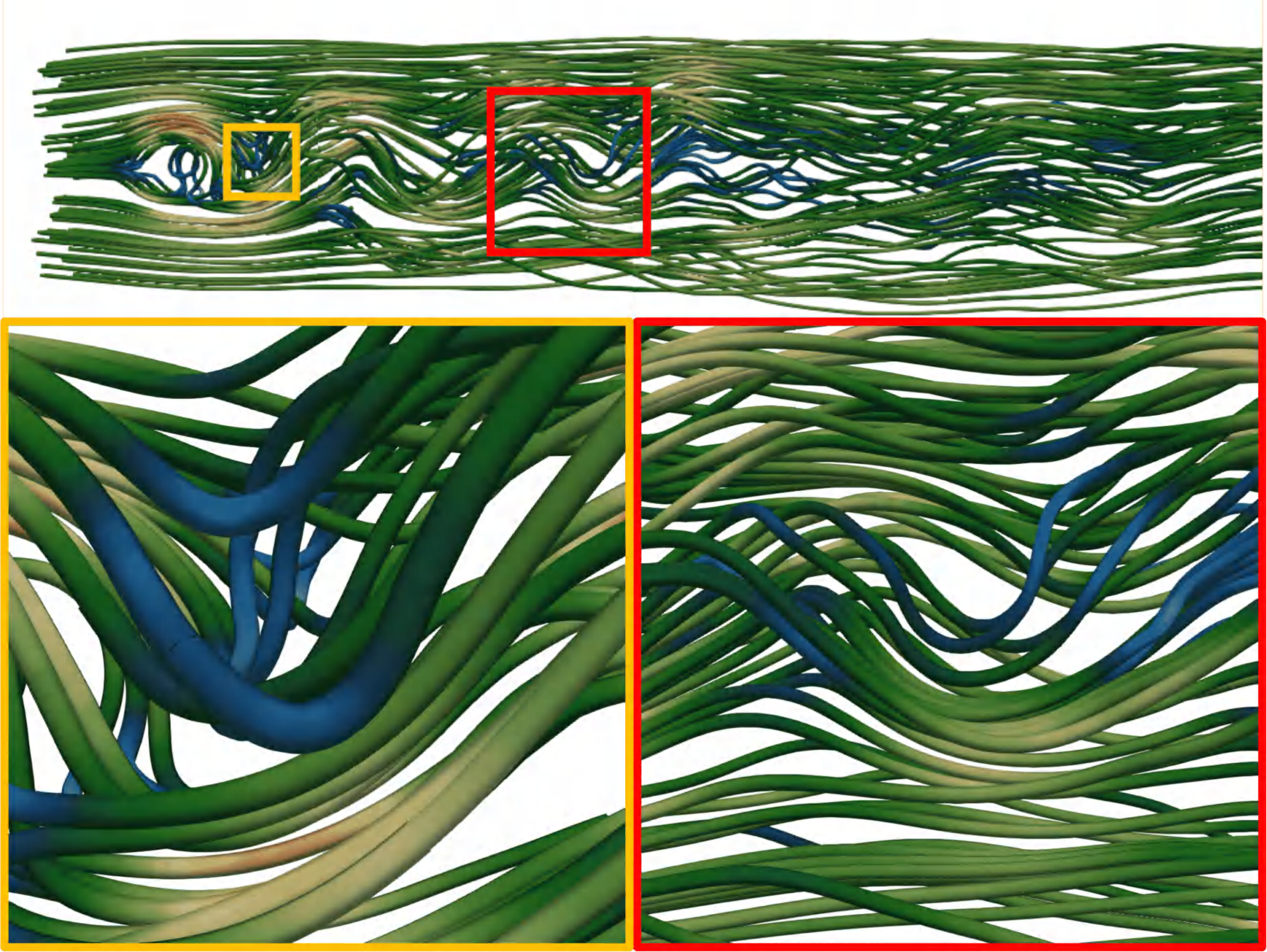} &
		\includegraphics[width=0.14\textwidth,height=0.18\textheight,keepaspectratio]{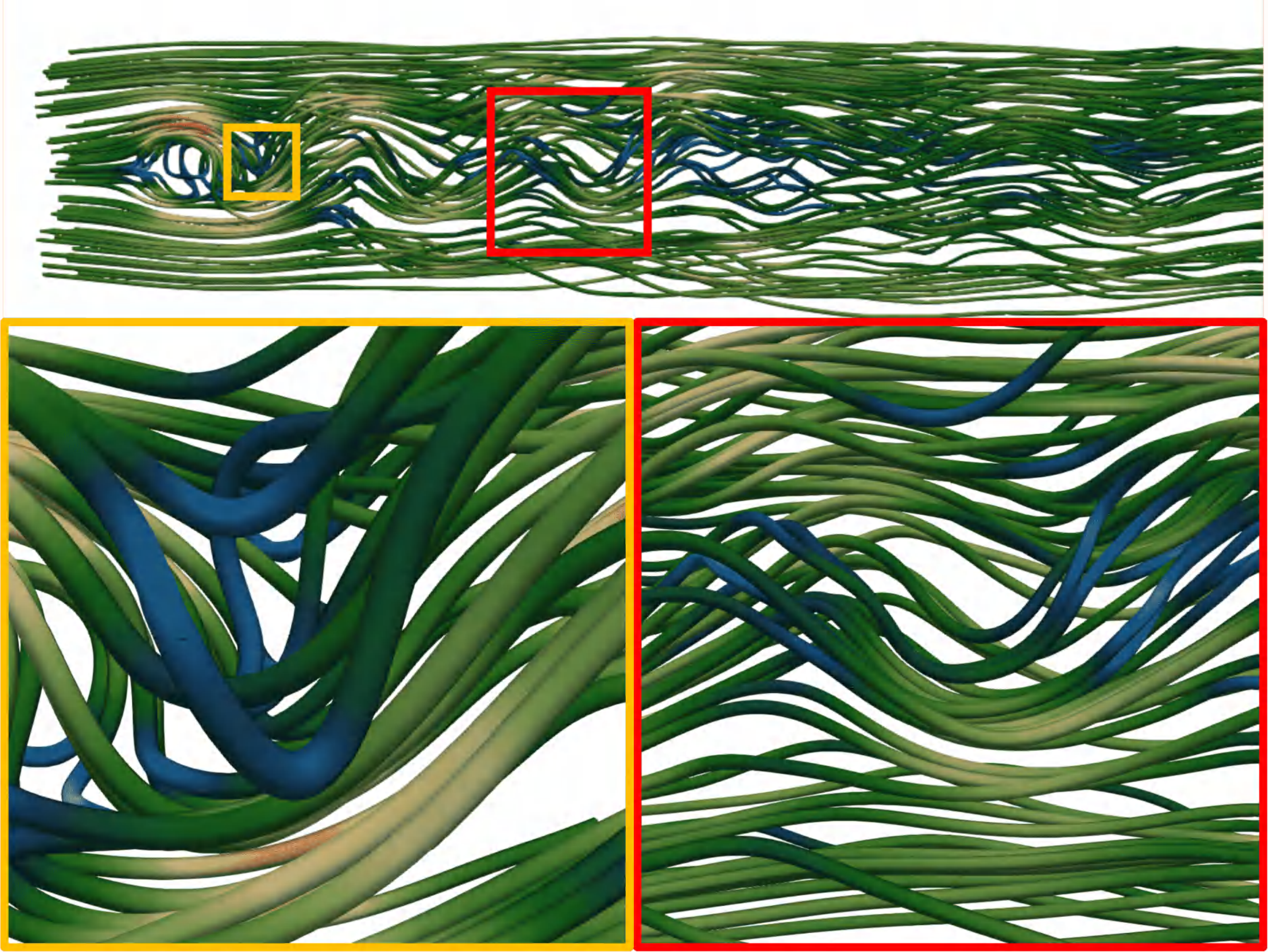} &
		\includegraphics[width=0.14\textwidth,height=0.18\textheight,keepaspectratio]{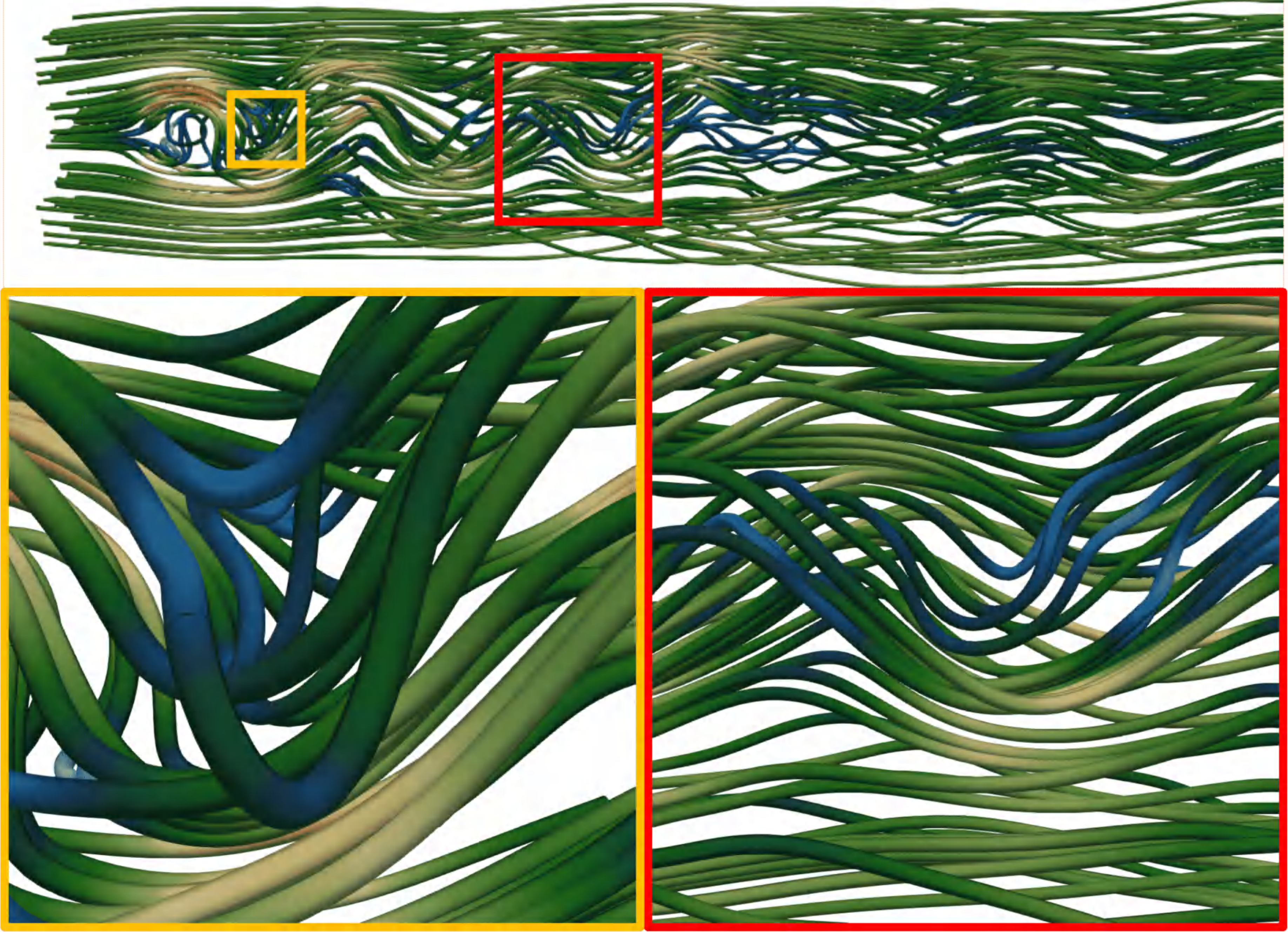} &
		\includegraphics[width=0.14\textwidth,height=0.18\textheight,keepaspectratio]{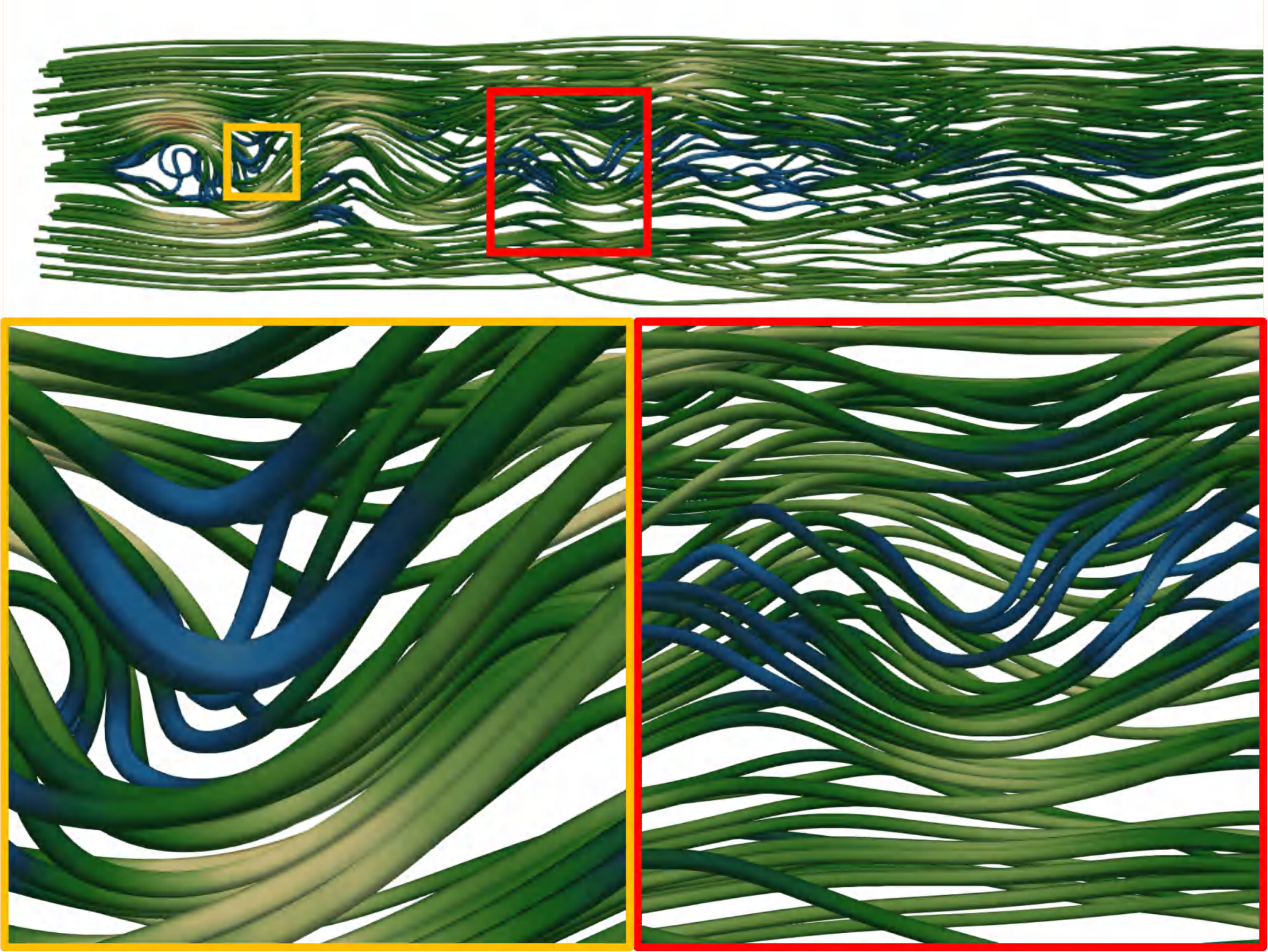} &
		\includegraphics[width=0.14\textwidth,height=0.18\textheight,keepaspectratio]{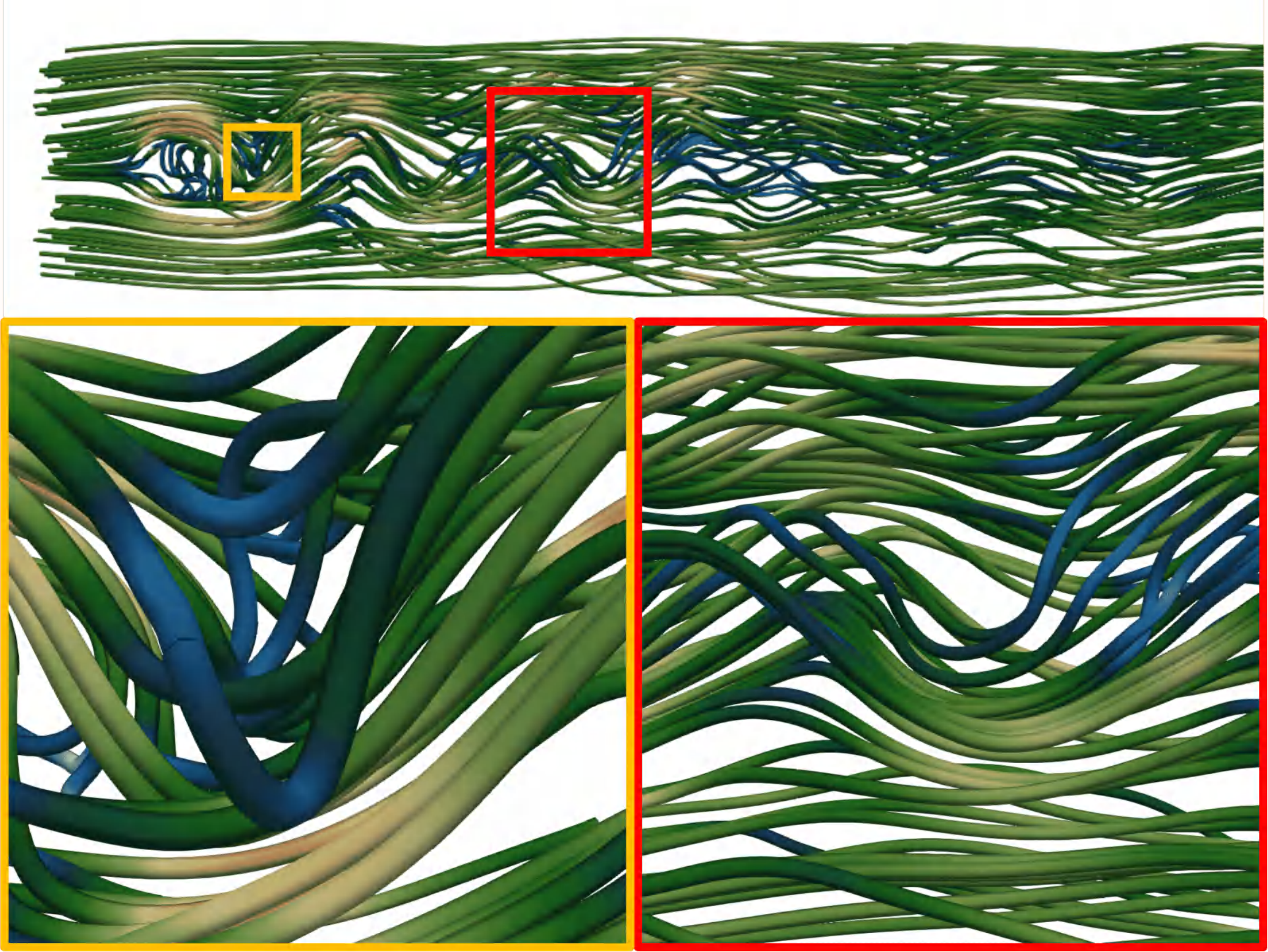} \\
		
		\includegraphics[width=0.14\textwidth,height=0.18\textheight,keepaspectratio]{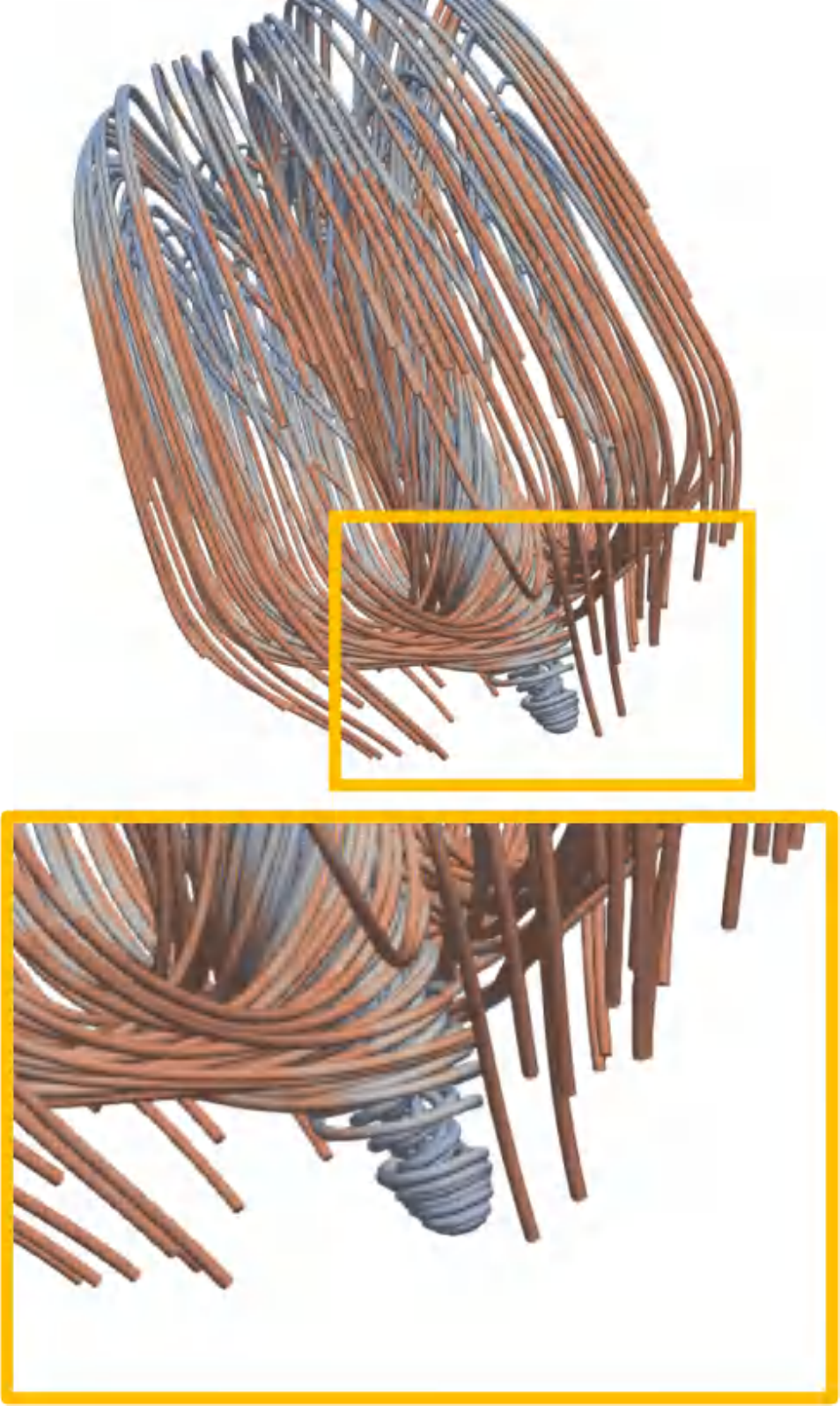} &
		\includegraphics[width=0.14\textwidth,height=0.18\textheight,keepaspectratio]{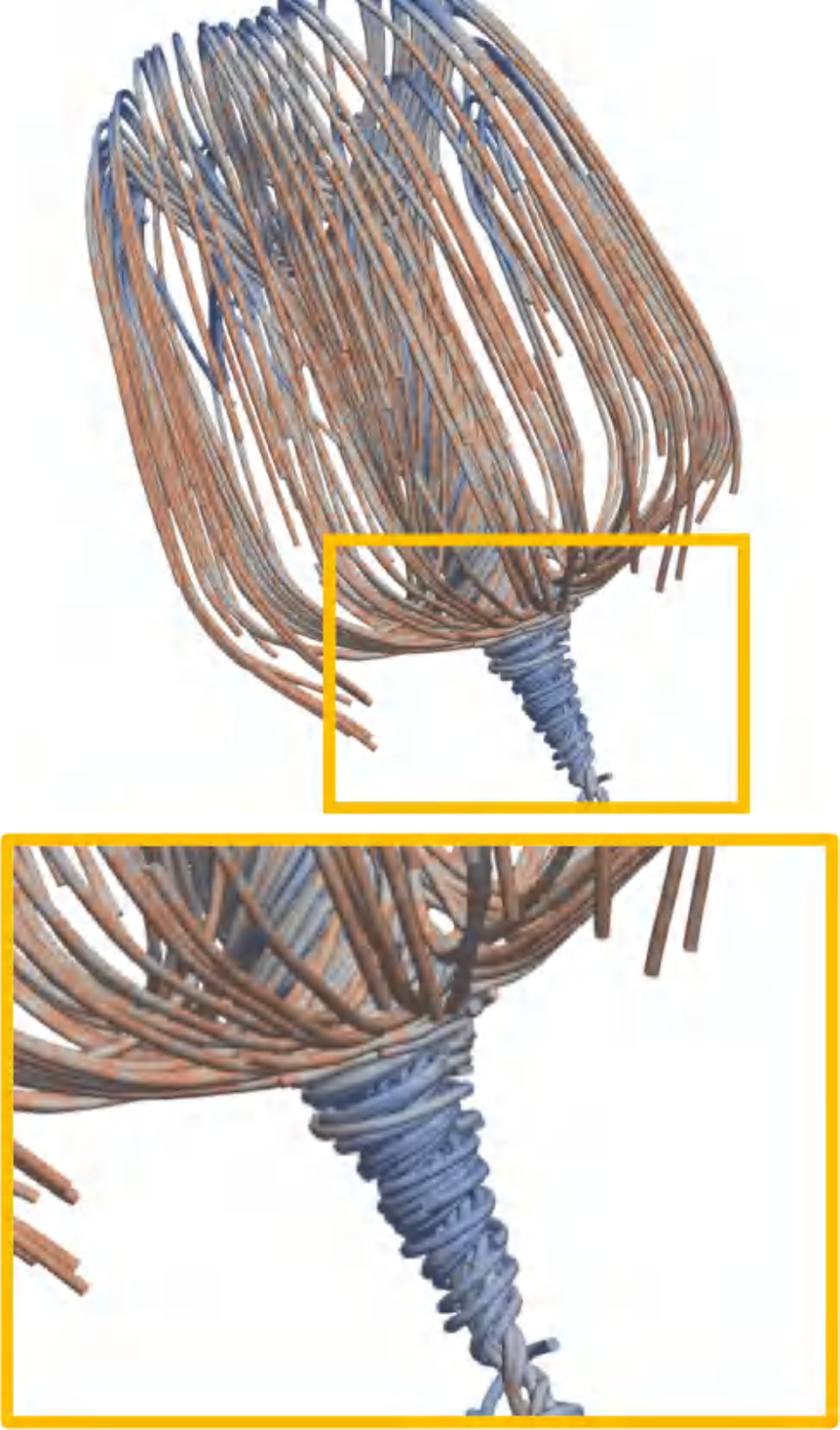} &
		\includegraphics[width=0.14\textwidth,height=0.18\textheight,keepaspectratio]{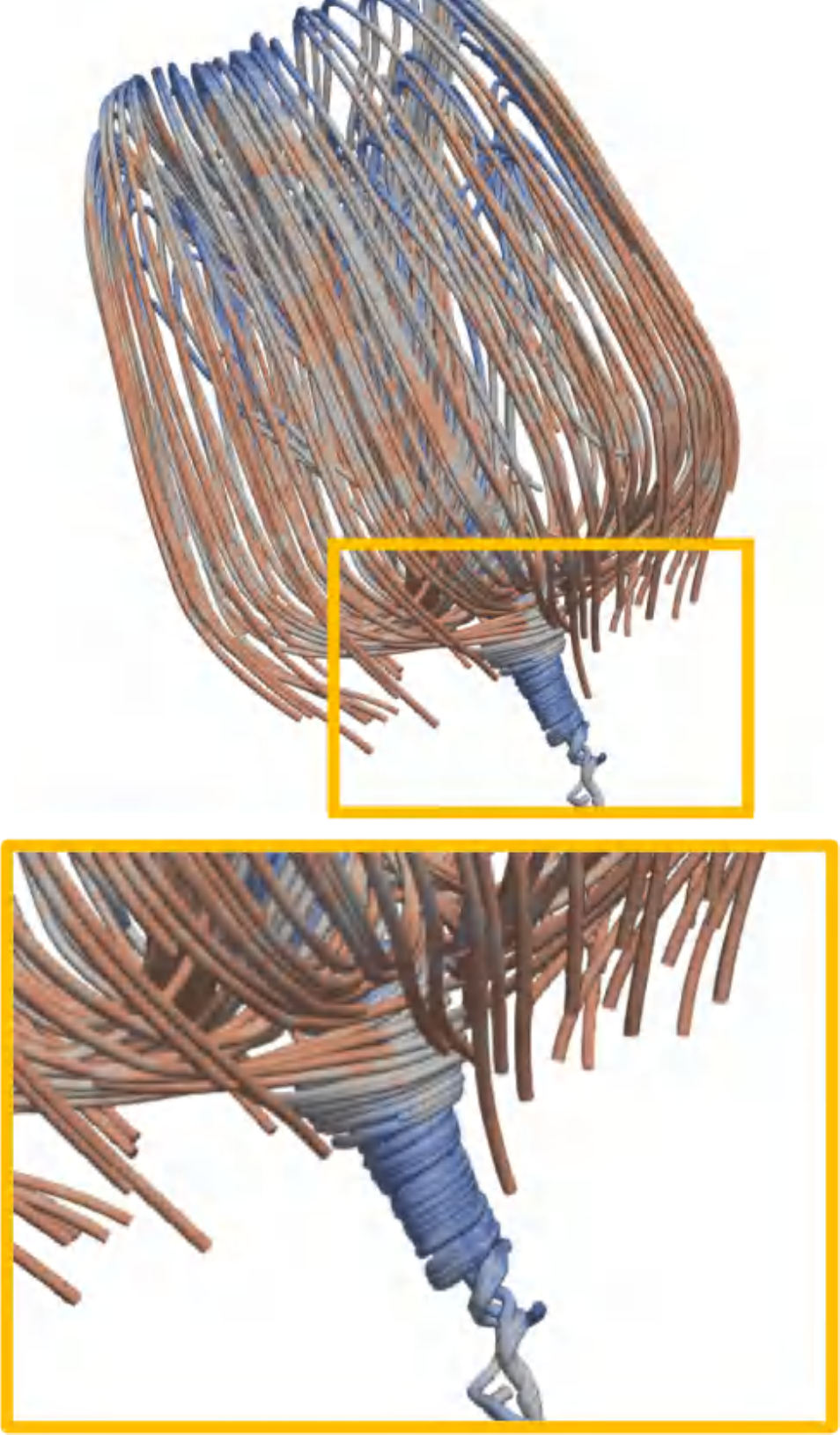} &
		\includegraphics[width=0.14\textwidth,height=0.18\textheight,keepaspectratio]{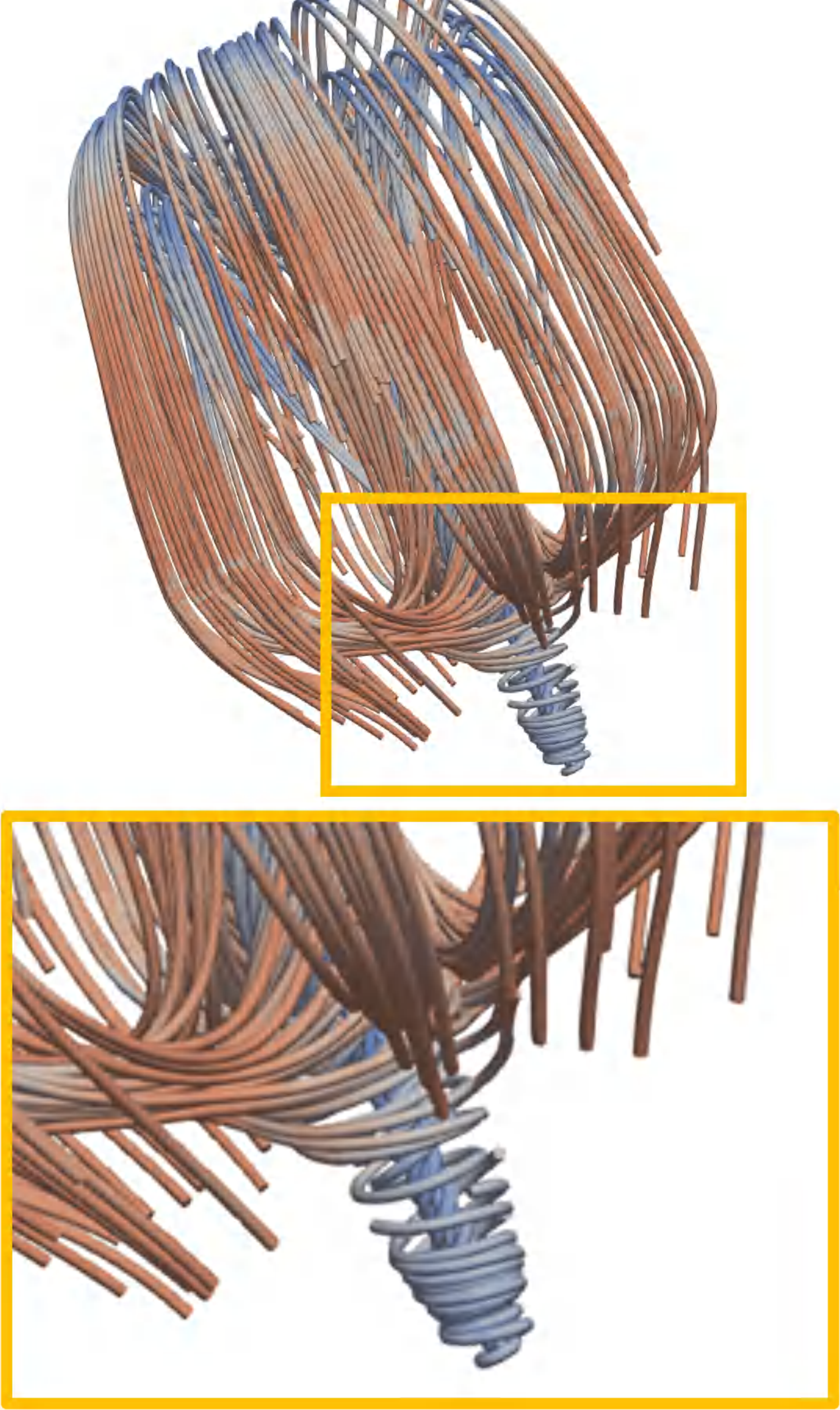} &
		\includegraphics[width=0.14\textwidth,height=0.18\textheight,keepaspectratio]{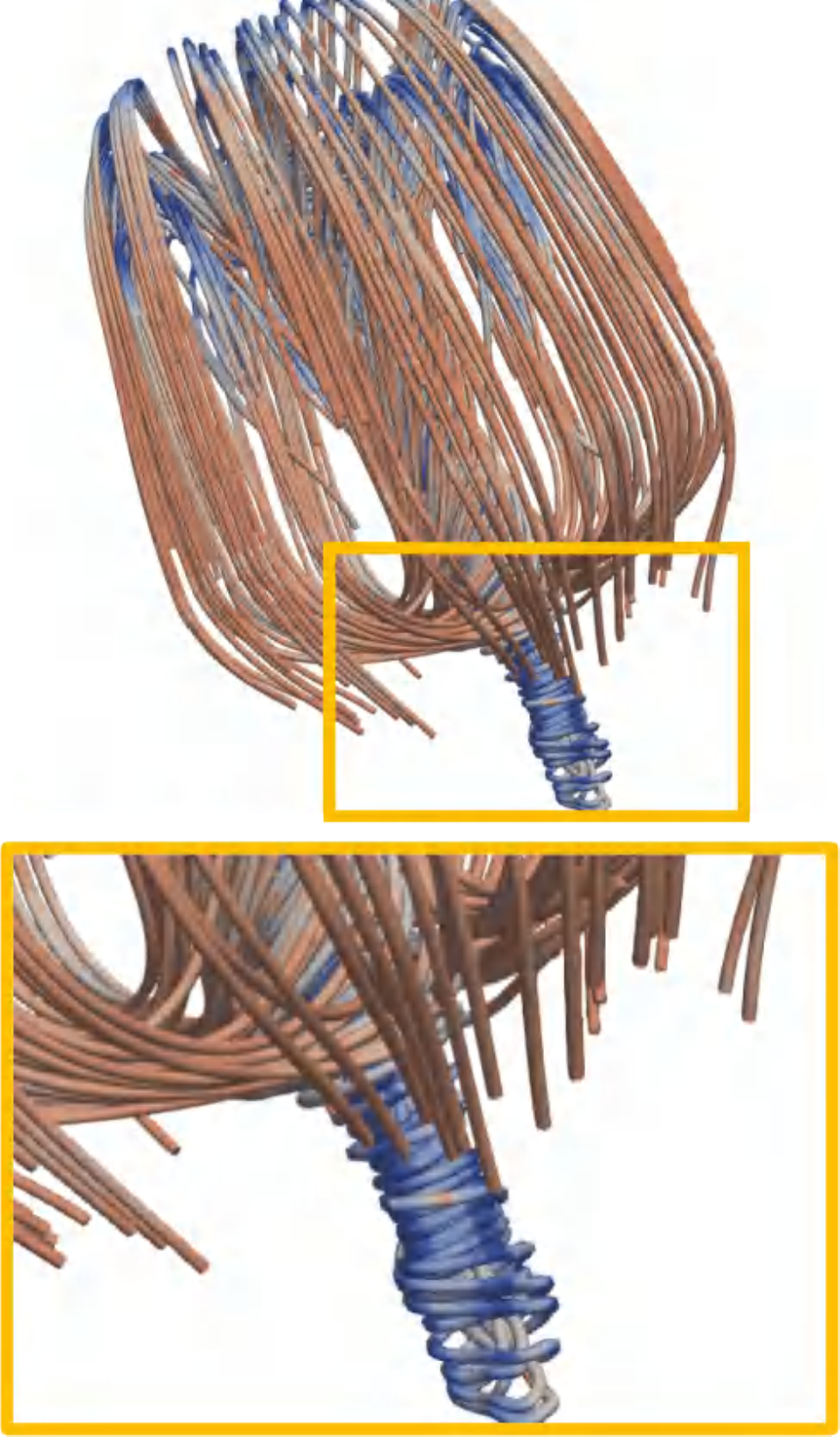} &
		\includegraphics[width=0.14\textwidth,height=0.18\textheight,keepaspectratio]{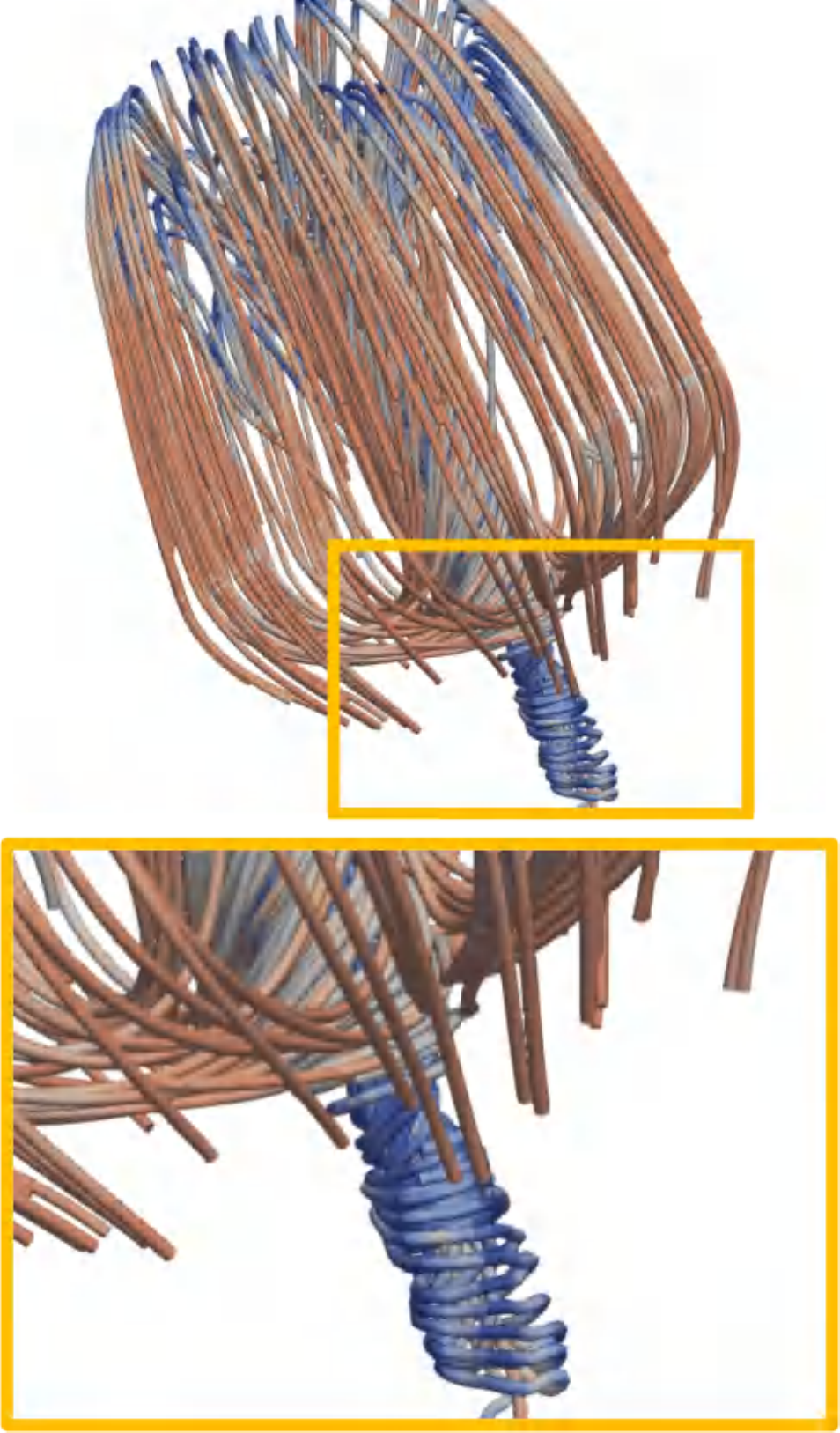} \\
		
		\includegraphics[width=0.14\textwidth,height=0.15\textheight,keepaspectratio]{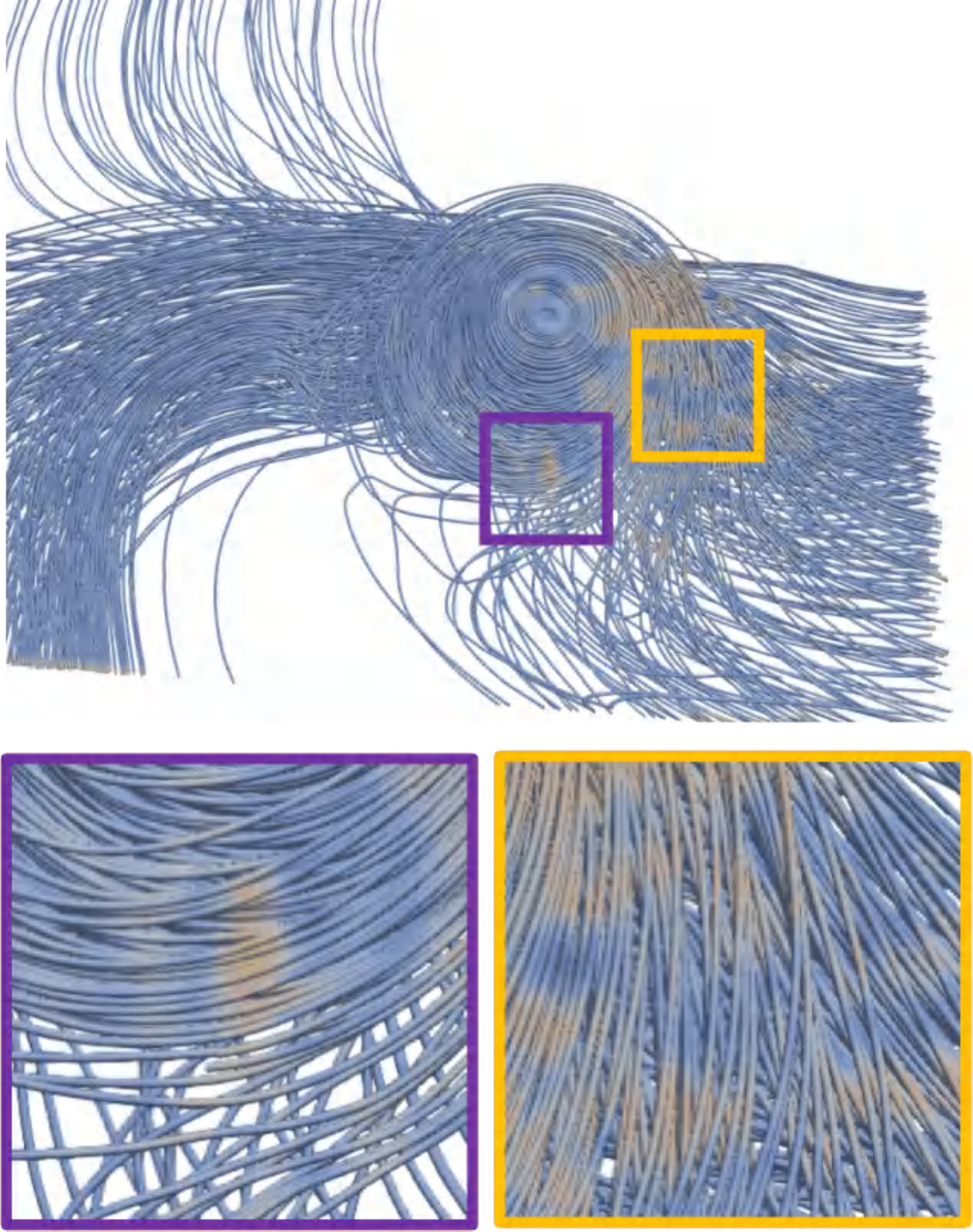} &
		\includegraphics[width=0.14\textwidth,height=0.15\textheight,keepaspectratio]{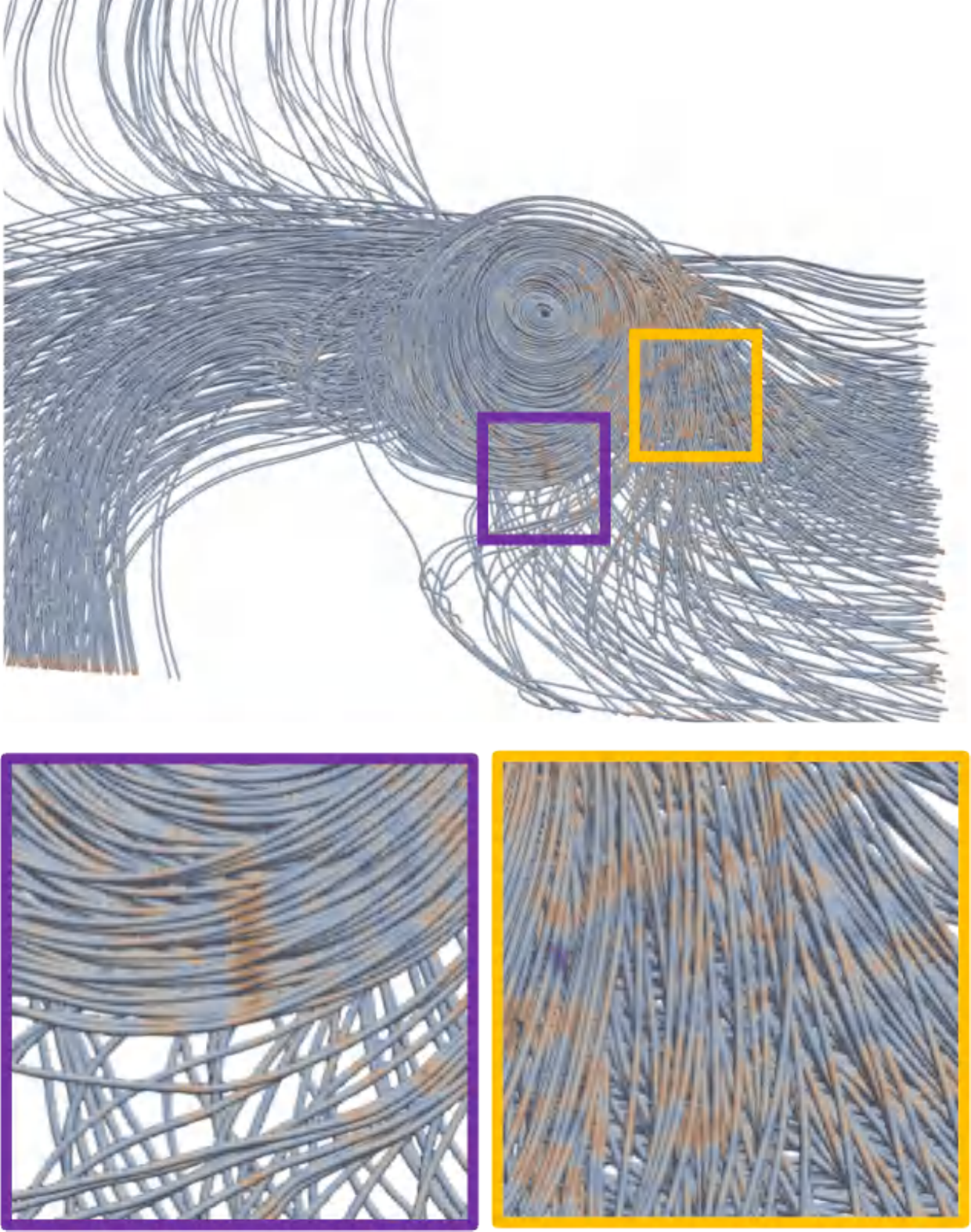} &
		\includegraphics[width=0.14\textwidth,height=0.15\textheight,keepaspectratio]{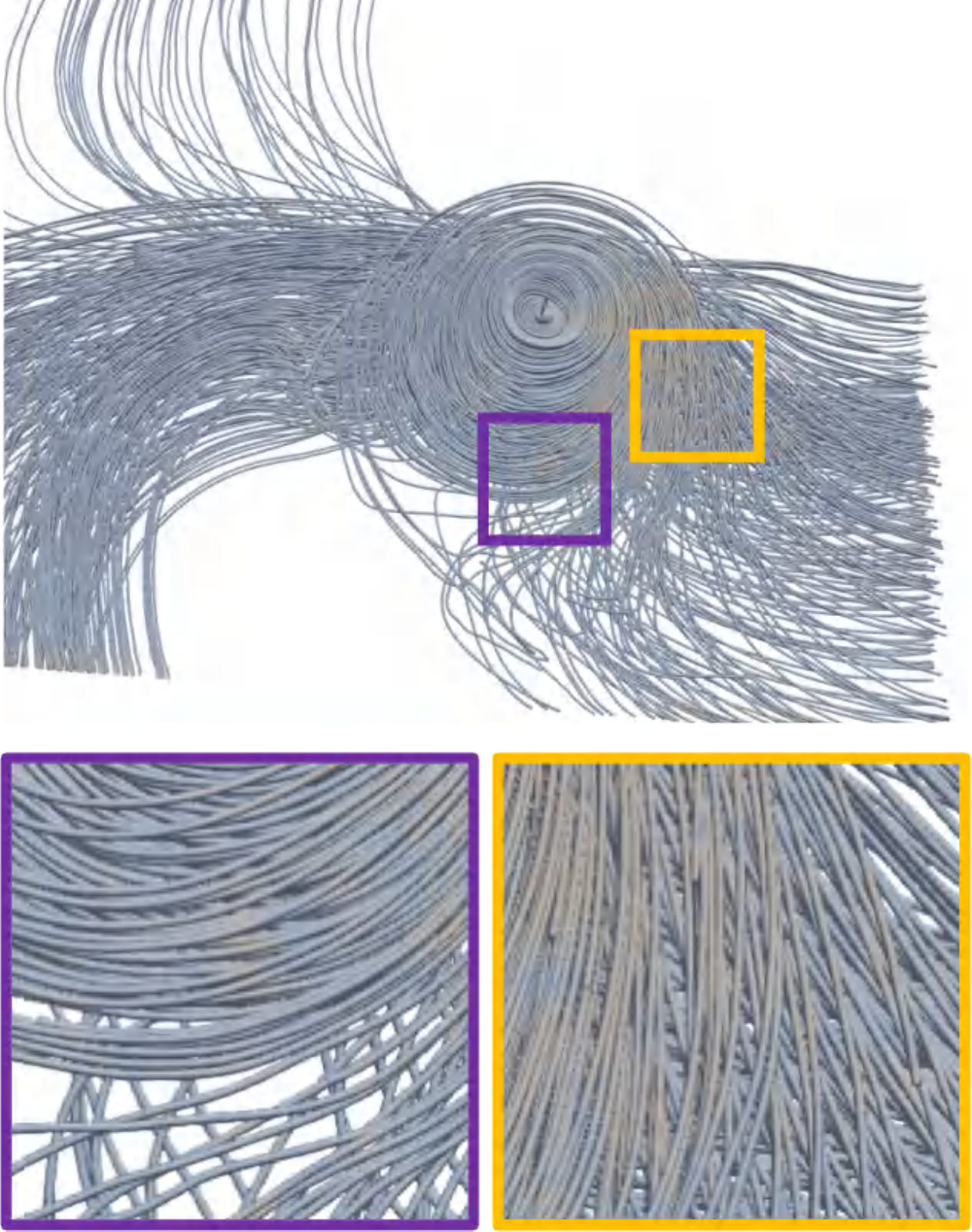} &
		\includegraphics[width=0.14\textwidth,height=0.15\textheight,keepaspectratio]{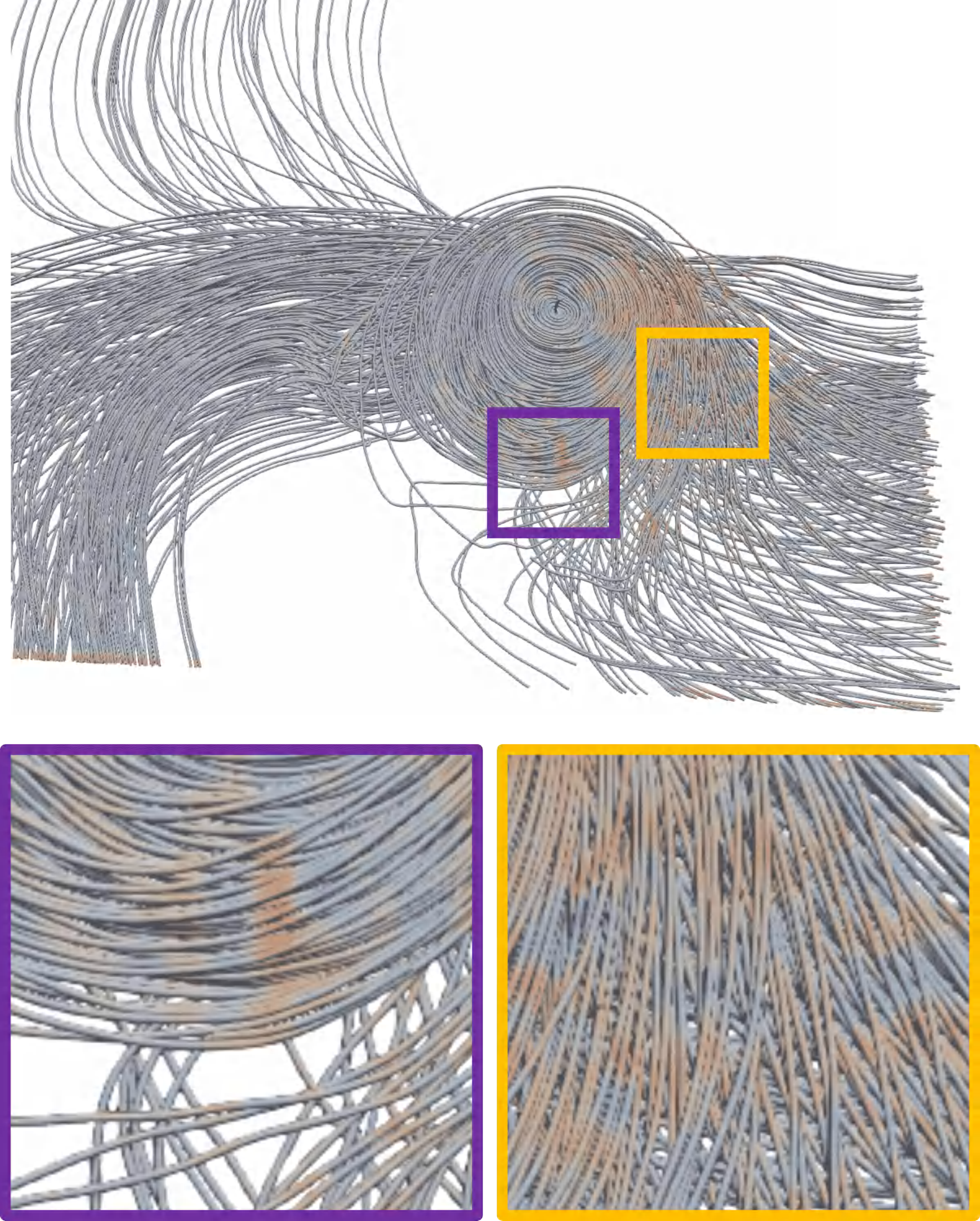} &
		\includegraphics[width=0.14\textwidth,height=0.15\textheight,keepaspectratio]{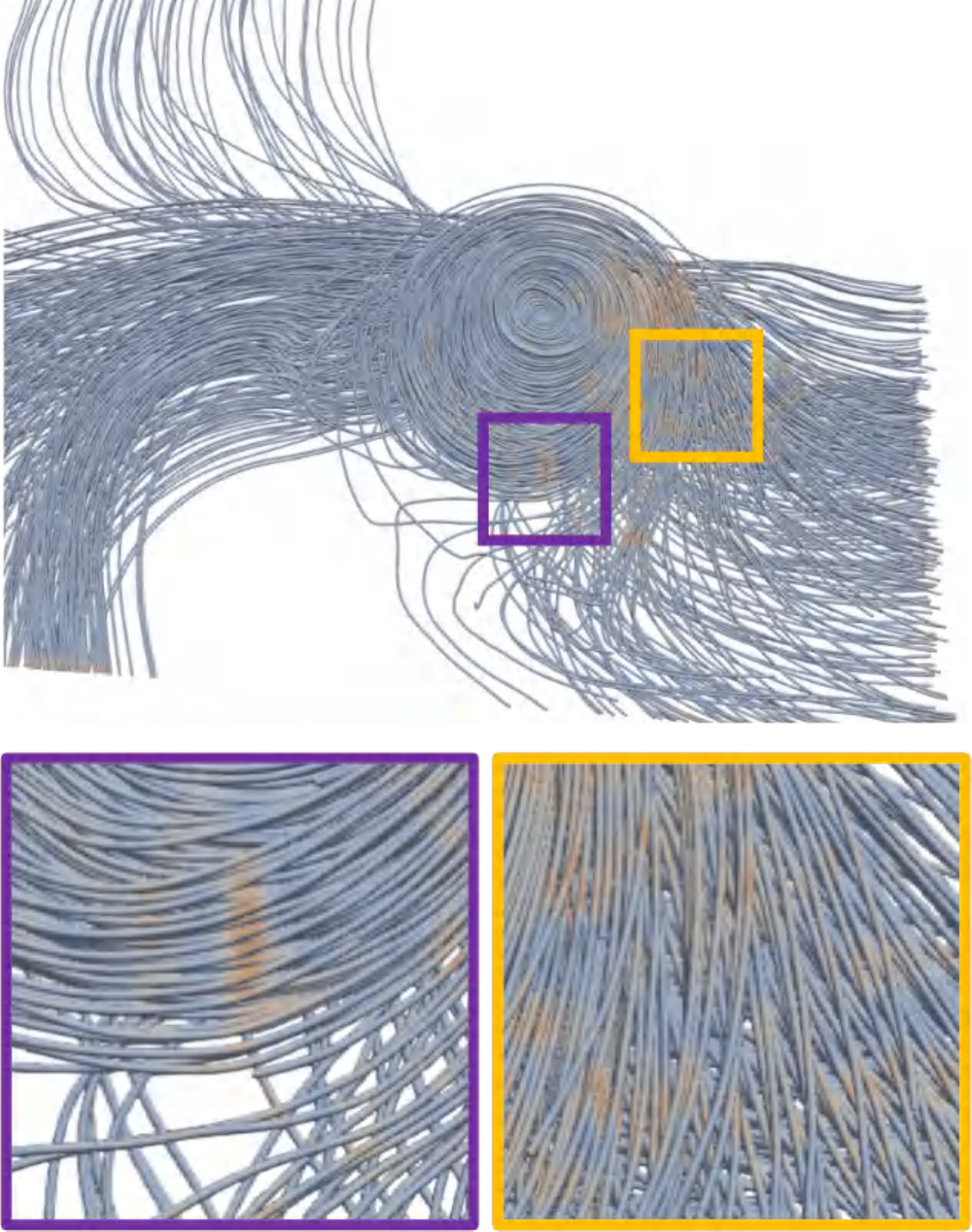} &
		\includegraphics[width=0.14\textwidth,height=0.15\textheight,keepaspectratio]{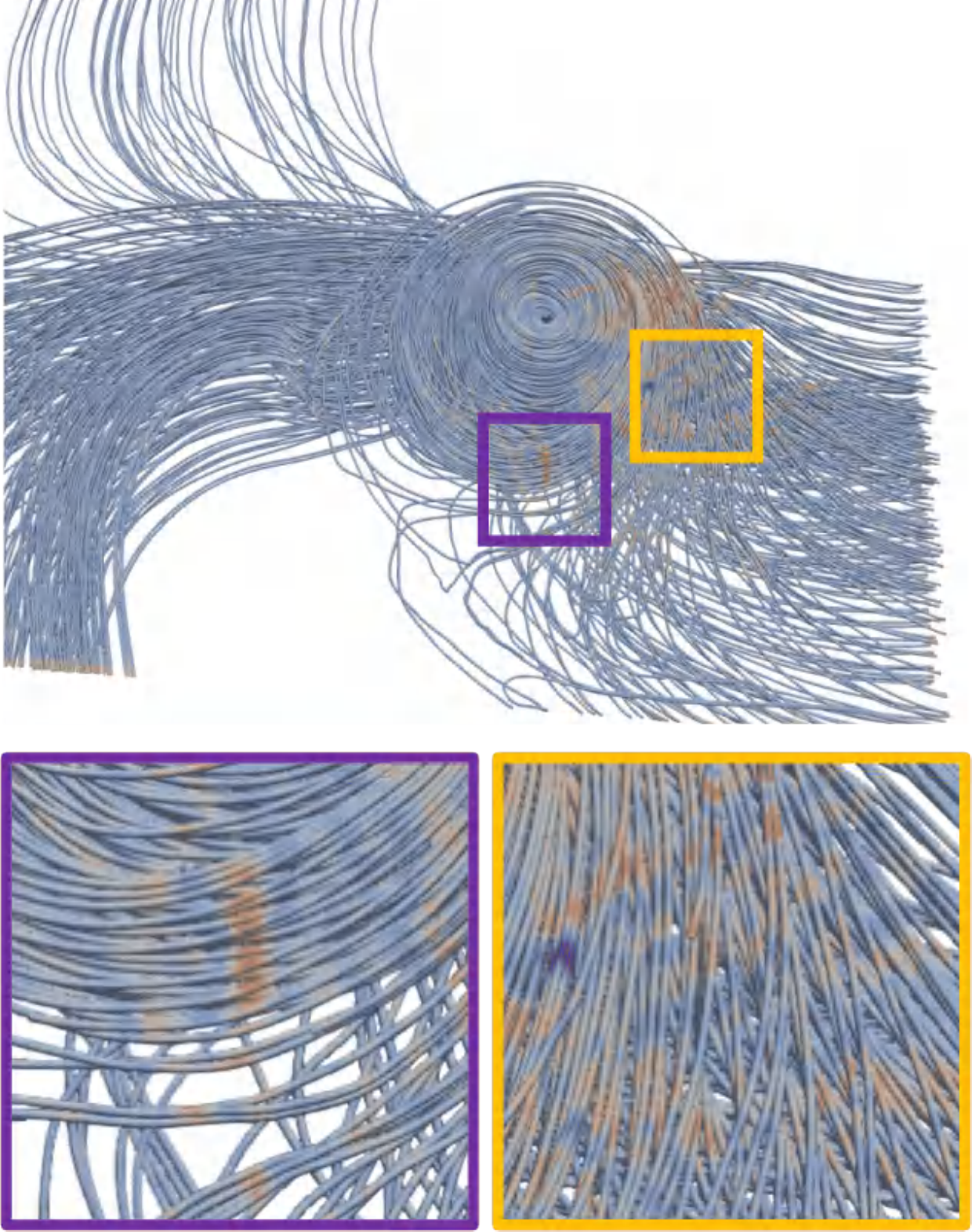} \\
		
		(a) TI & (b) SRGAN & (c) SSR-VFD & (d) PSRFlow & (e) CD-TVD & (f) GT \\
		
	\end{tabular}
	\vspace{-.15in} 
	\caption{Comparison of streamline rendering using TI, SRGAN, SSR-VFD, PSRFlow, and CD-TVD. Top to bottom: Research Vessel Tangaroa, Half Cylinder Ensemble, Shock Interaction Vortex, Hurricane.}
	\label{fig:streamline-rendering}
\end{figure*}

\subsubsection{Evaluation Metrics}
\label{sec:eval_metrics}

We employed three primary metrics to evaluate our super-resolution results. First, we used the Peak Signal-to-Noise Ratio (PSNR) for volumetric reconstructions. Let \(r\) be the maximum fluctuation in the dataset, and \(\text{MSE}\) be the mean squared error between the ground truth data \(F\) and our super-resolved result \(\hat{F}\).

Second, we employed the Learned Perceptual Image Patch Similarity (LPIPS)~\cite{zhang2018unreasonable} metric to evaluate perceptual quality at the image level. Note that LPIPS is computed from rendered images and thus depends on the chosen viewpoint. We rendered images from five random viewpoints to mitigate this issue and reported the average LPIPS score, ensuring a robust and consistent qualitative assessment.

Lastly, for vector fields, we measured the similarity between reconstructed data and ground-truth data by computing the Chamfer Distance (CD)~\cite{barrow1977parametric} between their respective streamlines. Specifically, we generated streamlines from a fixed set of 200 identical seed points across all datasets and measured the spatial positional differences between streamline point sets:

\begin{equation}\label{equ-4-4}
	d_{\mathrm{CD}}(F, \hat{F})=\frac{1}{F} \sum_{x \in F} \min_{y \in \hat{F}}\|x-y\|_2^2
	+\frac{1}{\hat{F}} \sum_{y \in \hat{F}} \min_{x \in F}\|y-x\|_2^2.
\end{equation}

Here, \(x\) and \(y\) are points on the reconstructed and true streamlines, respectively. These complementary metrics jointly quantify volumetric fidelity, perceptual quality, and flow field accuracy.

\subsubsection{Computational Cost Evaluation}

To further evaluate the practical applicability of our CD-TVD method, we compared its model size and training time with several baseline approaches on the Shock Interaction Vortex dataset, as summarized in Table~\ref{tab:model_size_training_time}. In terms of memory footprint, CD-TVD required moderately more memory than PSRFlow but significantly less memory than SSR-TVD, and was comparable to SRGAN. 

\begin{table}[b]
	\centering
	\caption{
		Comparison of model size in MB and training time in hours for different methods on the Shock Interaction Vortex dataset.
	}
	{\scriptsize
		\begin{tabular}{cccc}
			\hline
			\textbf{Method} & \textbf{Model Size} & \textbf{Pre-training Time} & \textbf{Fine-tuning Time} \\
			\hline
			CD-TVD & 19.5& 36 & 5 \\ 
			SSR-VFD & 51.4 & 0 & 4 \\ 
			SRGAN & 20.9  & 0 & 3 \\
			PSRFlow & 16.6 & 0 & 4 \\
			\hline
		\end{tabular}
	}
	\label{tab:model_size_training_time}
\end{table}

In terms of training time, CD-TVD consists of two stages: a one-time pre-training stage lasting approximately 36 hours, followed by a fine-tuning stage of about 5 hours. Although the pre-training stage is significantly longer compared to other methods, it is a one-time effort that can be conducted in advance. During fine-tuning, CD-TVD requires approximately 5 hours, slightly longer than the baselines. This additional time is primarily because diffusion models are trained not only on the original data but also on multiple versions of the data with varying levels of noise corruption. This data augmentation process effectively increases the amount of training data, leading to higher computational costs. However, it significantly improves the generalizability of the model, enabling it to recover high-frequency details under a wide range of degraded conditions.

Overall, CD-TVD strikes a good balance between model size and training efficiency, while offering stronger robustness and generalization capabilities due to the intrinsic properties of the diffusion process.

\subsection{Qualitative and Quantitative Analysis.} 
\subsubsection{Quantitative Analysis}

\begin{figure*}[tb]
	\centering
	\begin{tabular}{cccc}
		\includegraphics[width=0.22\textwidth,height=0.12\textheight]{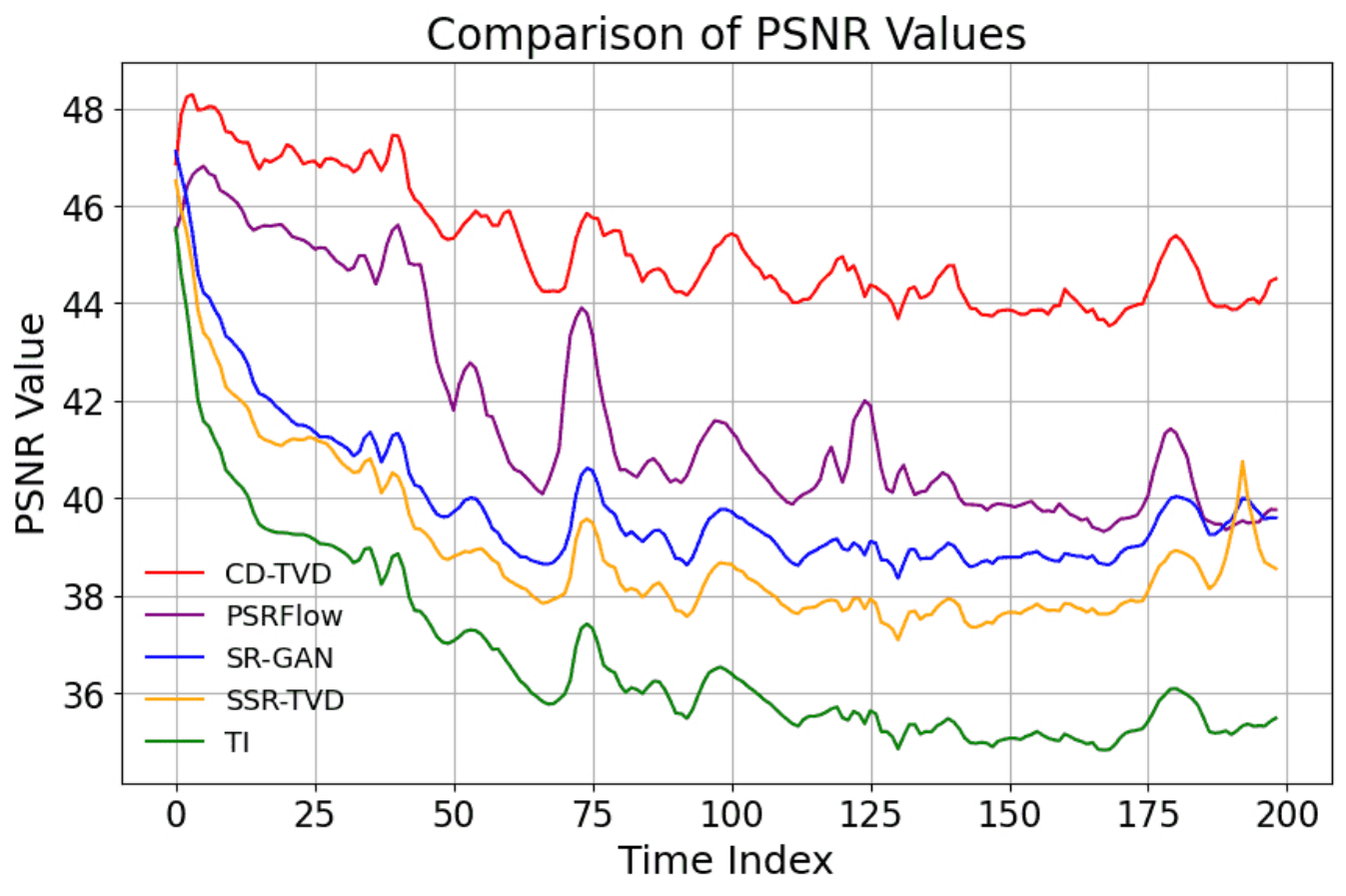} &
		\includegraphics[width=0.22\textwidth,height=0.12\textheight]{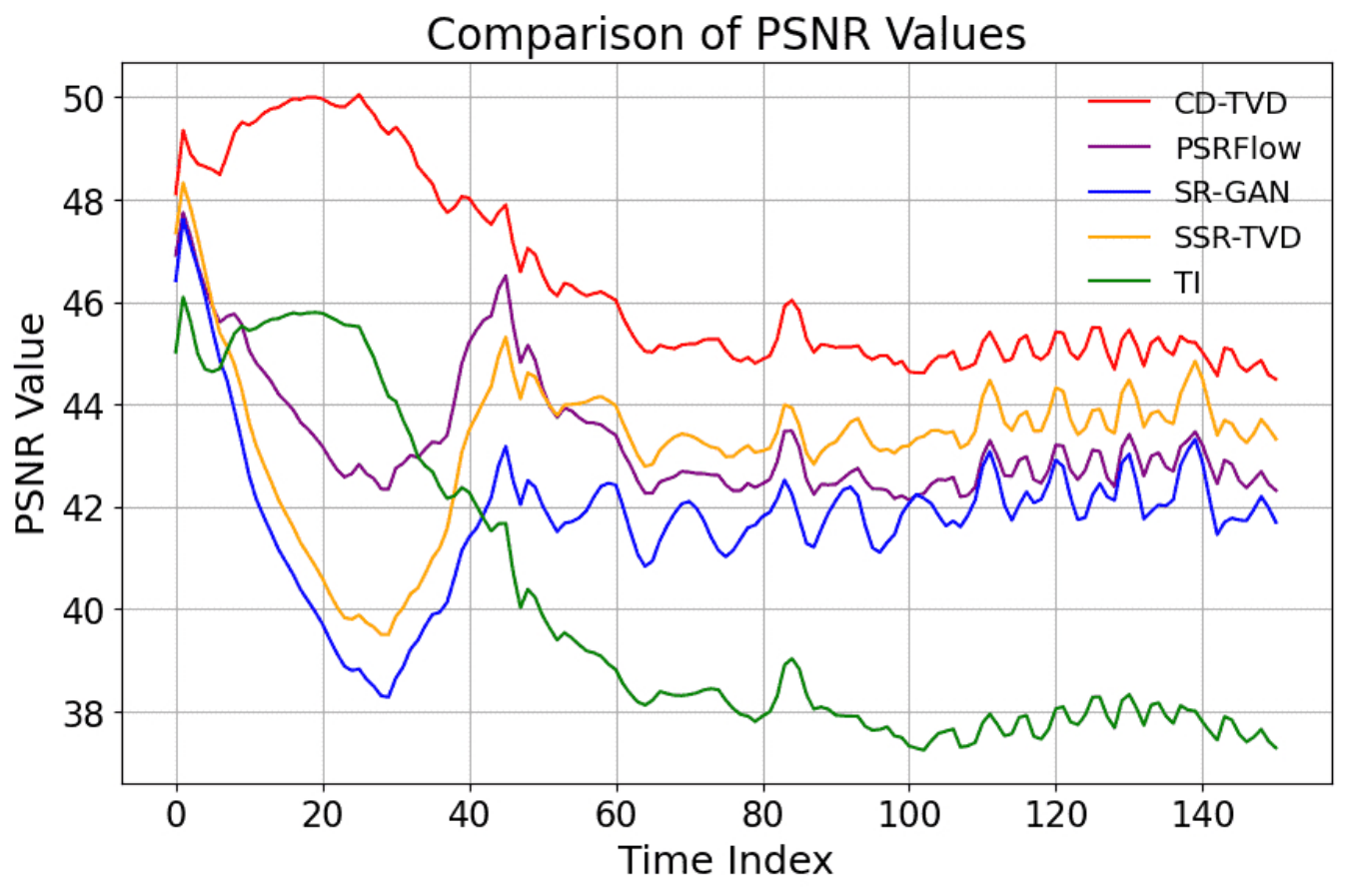} &
		\includegraphics[width=0.22\textwidth,height=0.12\textheight]{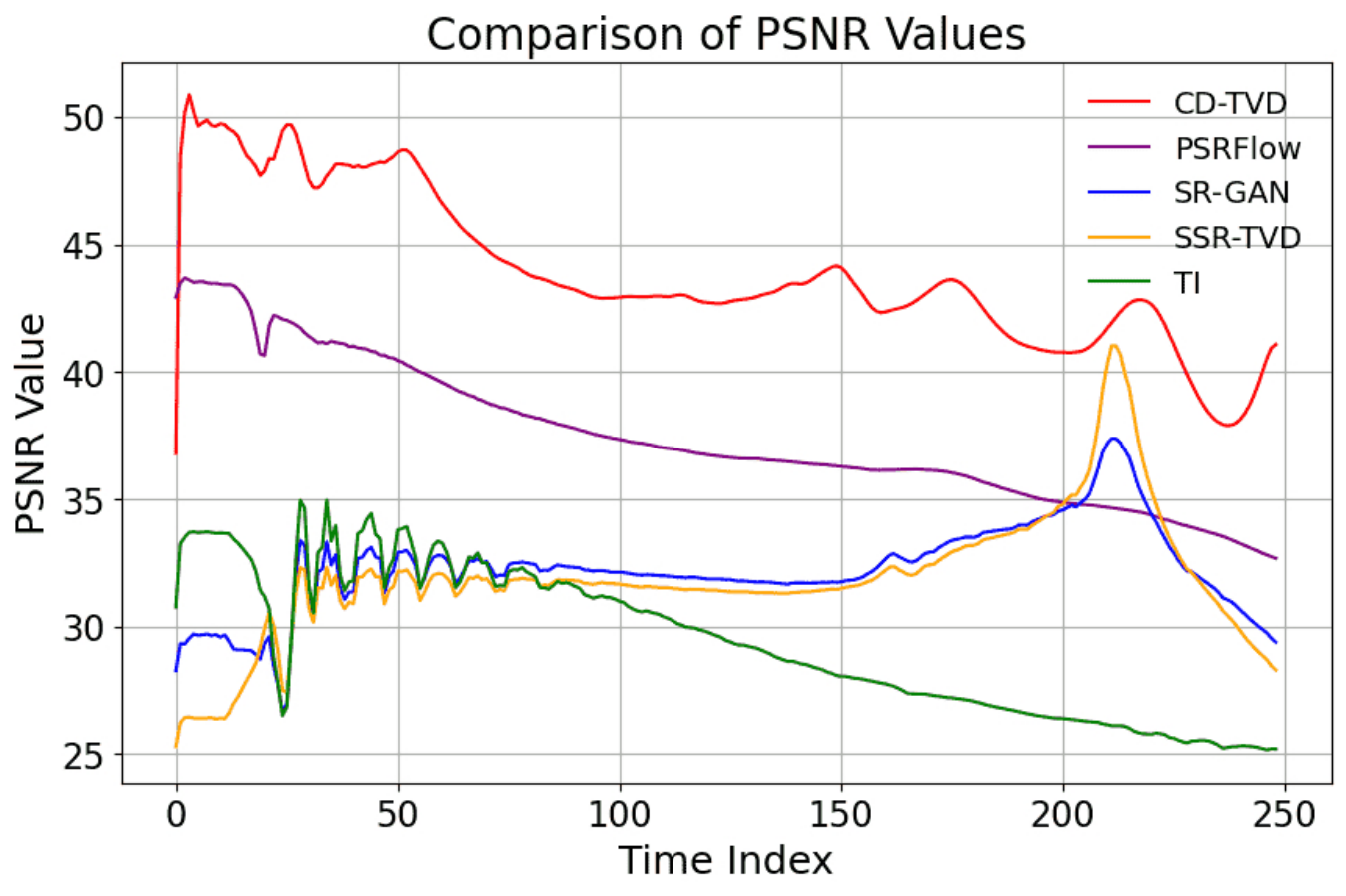} &
		\includegraphics[width=0.22\textwidth,height=0.12\textheight]{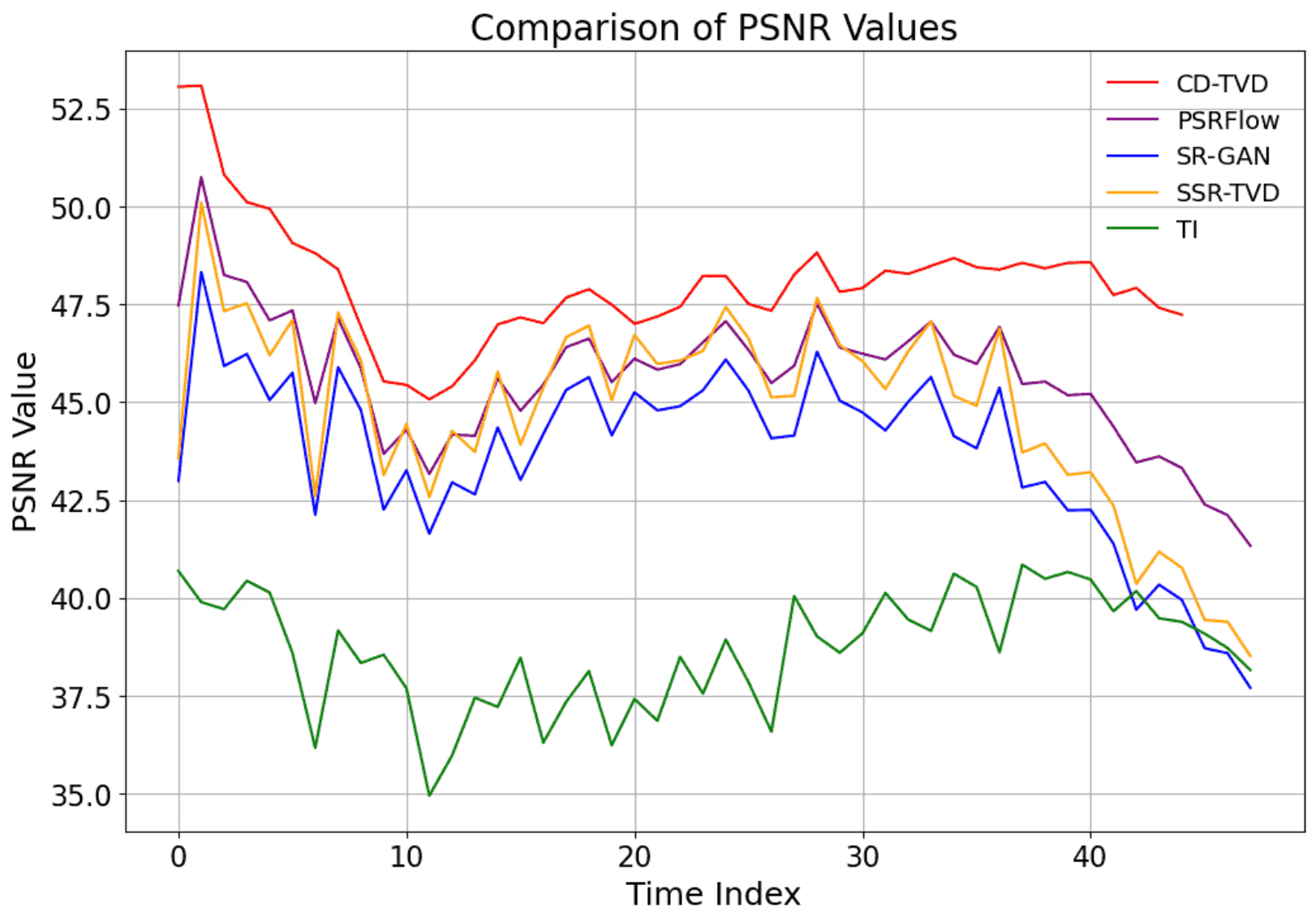} \\
		
		\includegraphics[width=0.22\textwidth,height=0.12\textheight]{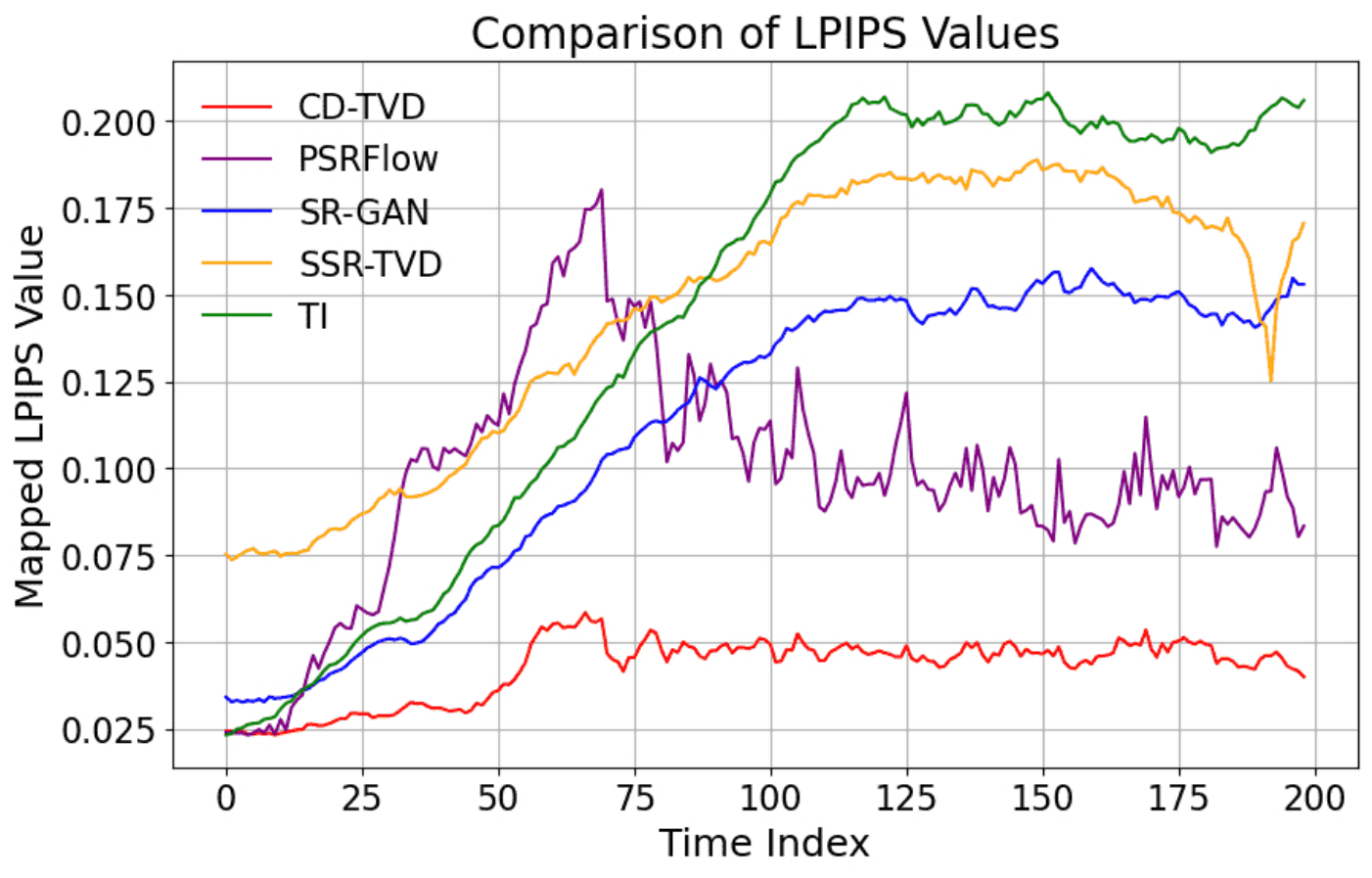} &
		\includegraphics[width=0.22\textwidth,height=0.12\textheight]{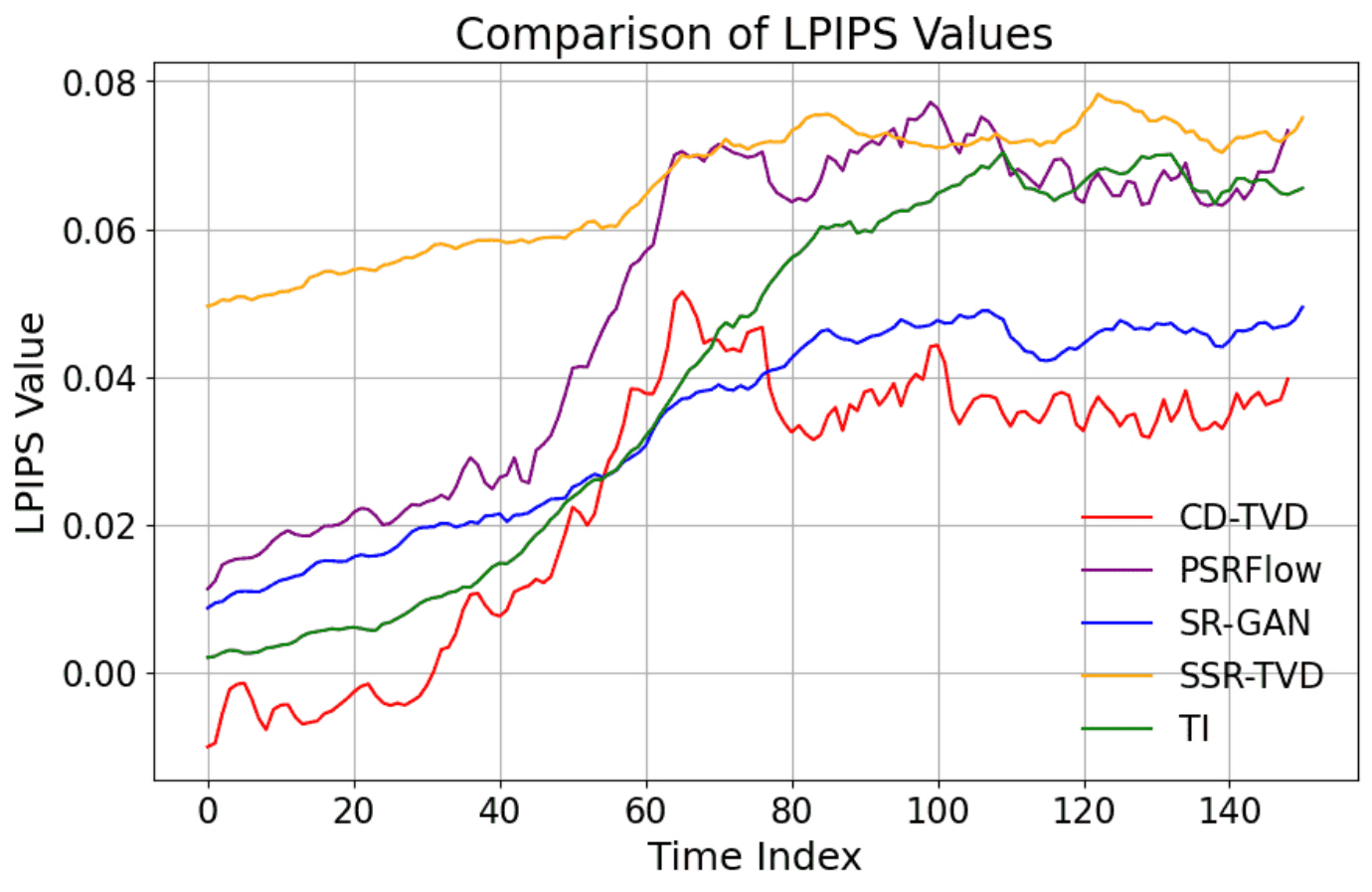} &
		\includegraphics[width=0.22\textwidth,height=0.12\textheight]{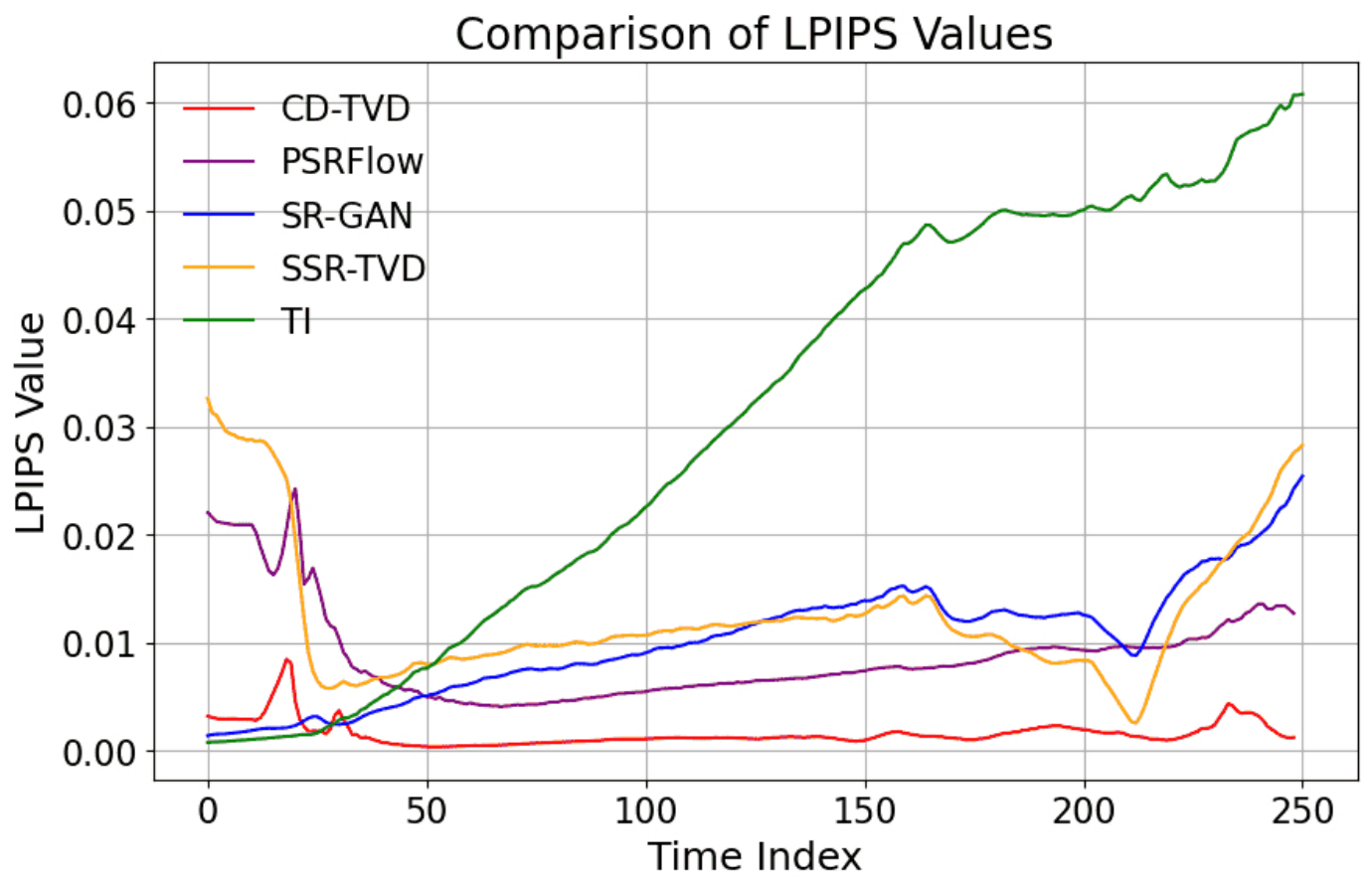} &
		\includegraphics[width=0.22\textwidth,height=0.12\textheight]{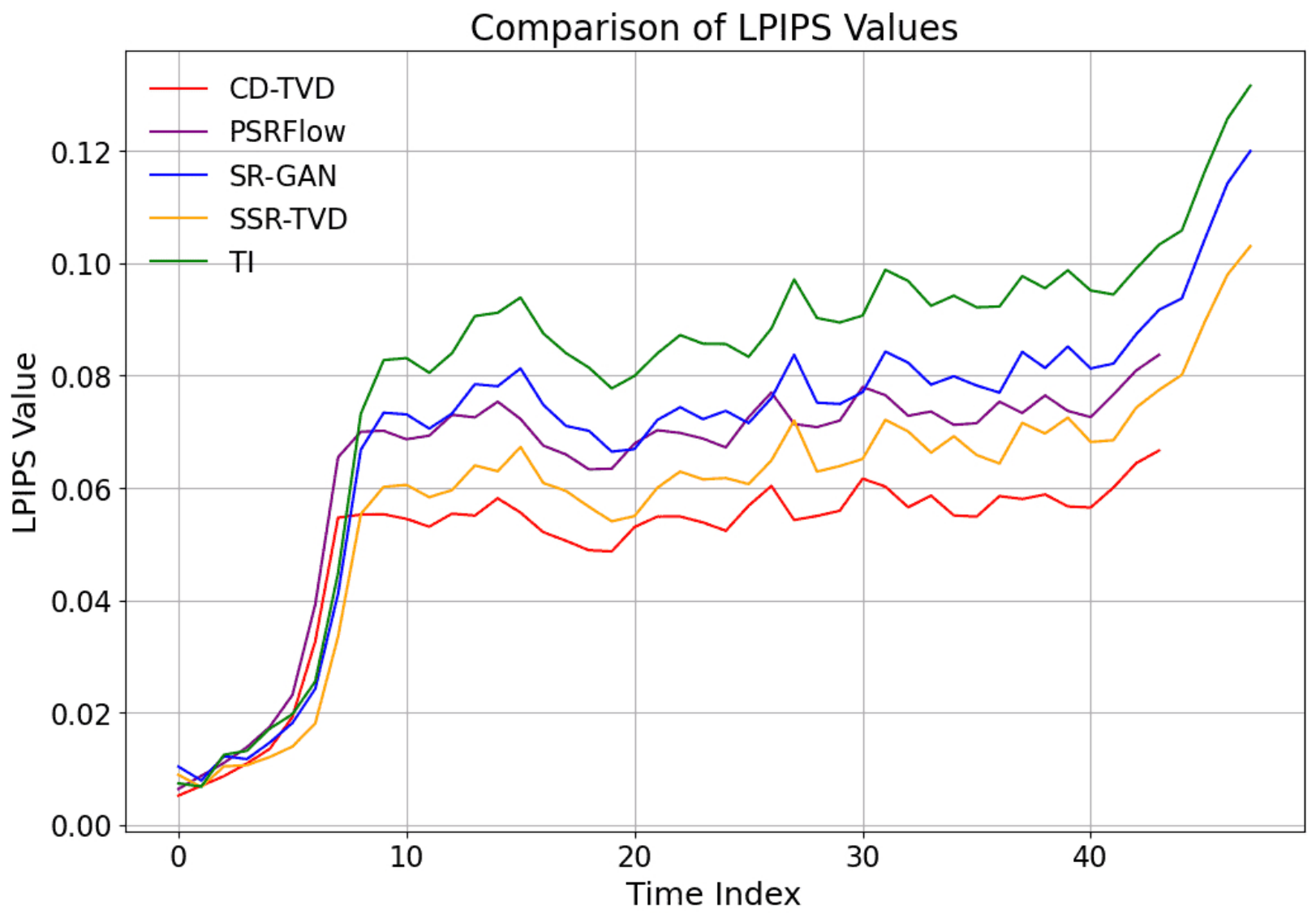} \\
		
		\includegraphics[width=0.22\textwidth,height=0.12\textheight,keepaspectratio]{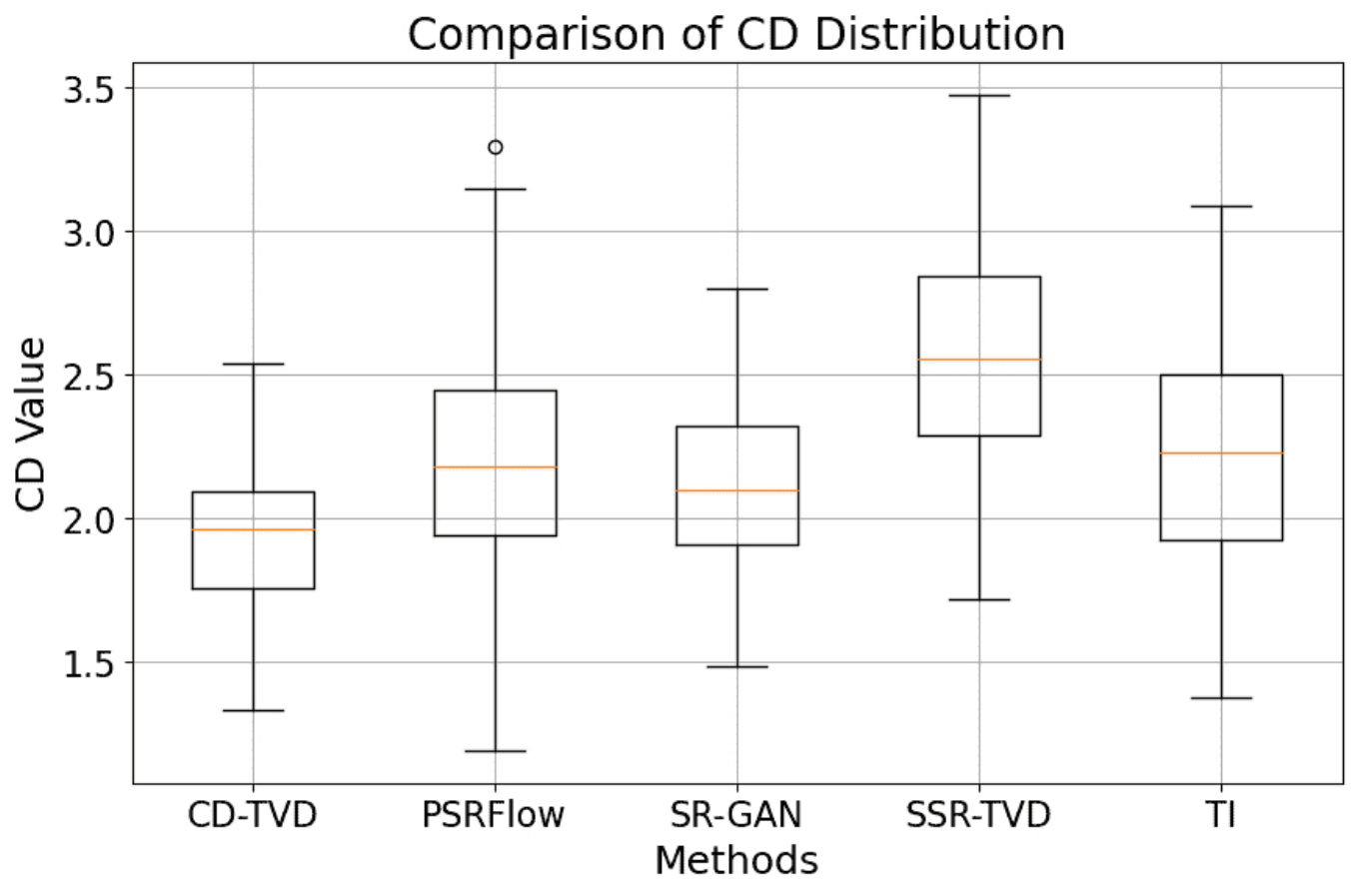} &
		\includegraphics[width=0.22\textwidth,height=0.12\textheight,keepaspectratio]{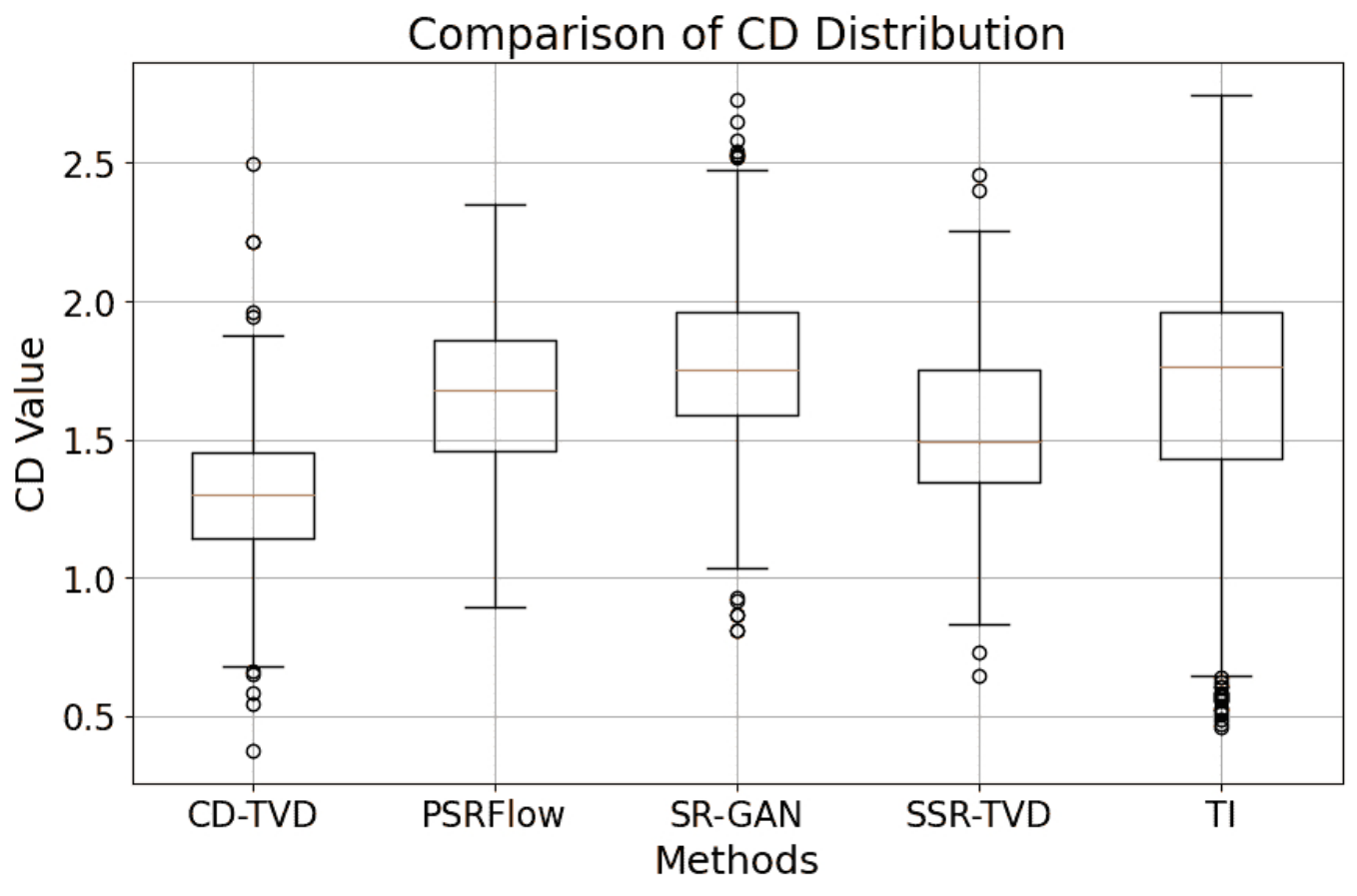} &
		\includegraphics[width=0.22\textwidth,height=0.12\textheight,keepaspectratio]{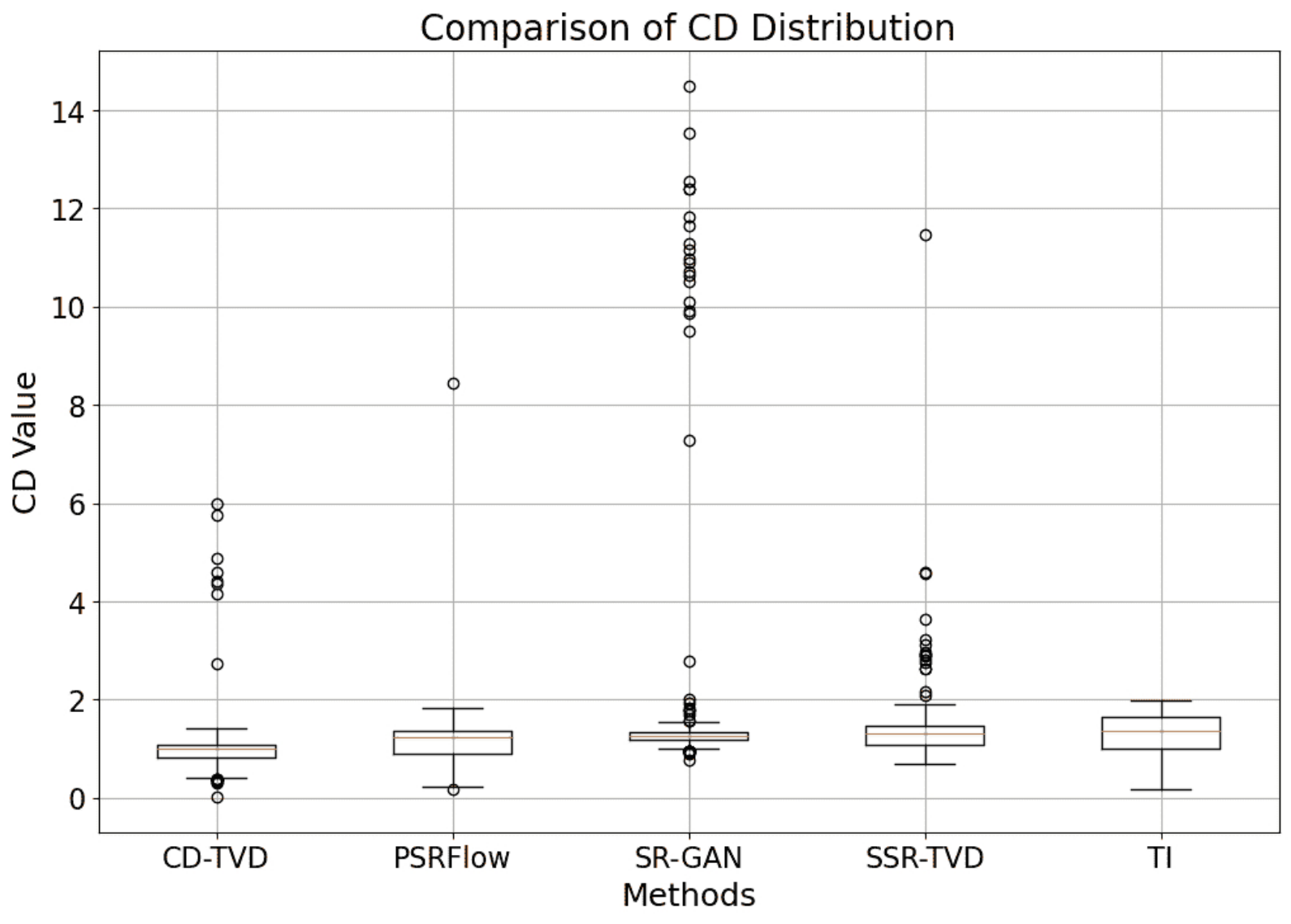} &
		\includegraphics[width=0.22\textwidth,height=0.12\textheight,keepaspectratio]{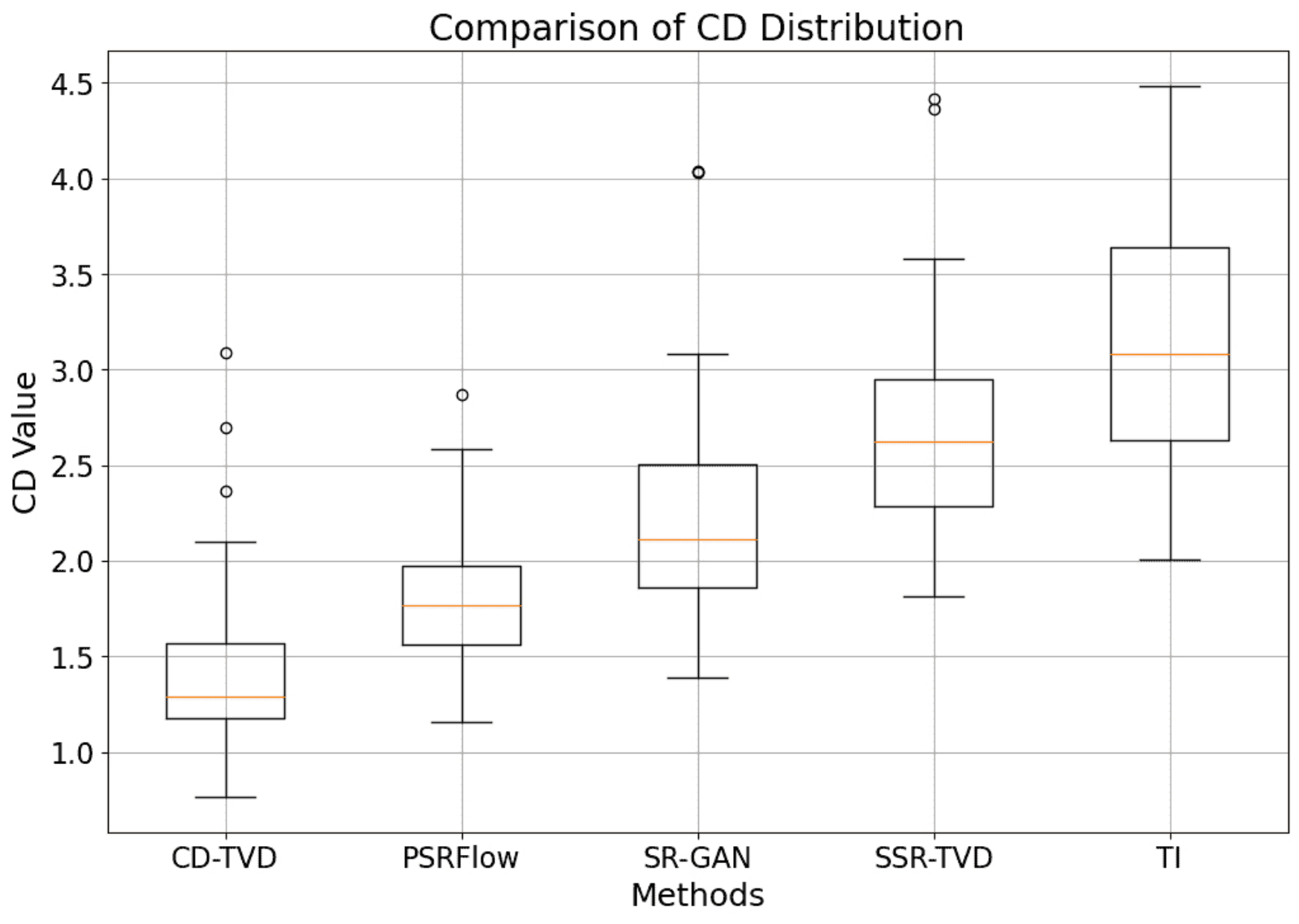} \\
		
		(a) Research Vessel Tangaroa & (b) Half Cylinder Ensemble & (c) Shock Interaction Vortex & (d) Hurricane \\
	\end{tabular}
	\vspace{-.15in} 
	\caption{Comparison of the synthesized vector fields using TI, SRGAN, SSR-VFD, and CD-TVD methods. Rows from top to bottom show PSNR (higher is better), LPIPS (lower is better), and CD (lower is better) results at the image level.}
	\label{fig:combined_vector_field_comparison}
\end{figure*}

In Fig.~\ref{fig:combined_vector_field_comparison}, we compared the PSNR performance of five methods across four datasets. The PSNR curves illustrate the accuracy of the reconstruction in pixel-wise over multiple time steps,  where higher values indicate better reconstruction quality. In all four datasets, CD-TVD consistently achieved the highest and most stable PSNR scores, demonstrating its strong ability to recover fine-scale details from LR inputs. A collective analysis of the results shows that CD-TVD does not exhibit notable weaknesses or significant fluctuations. This robustness largely arises from its pretraining process, which effectively learns the underlying data degradation patterns, enabling the model to maintain high performance across different timesteps.

In contrast, the other methods show marked performance drops or fluctuations in specific ranges, mainly because they struggle to adapt to sparse HR data and thus do not learn the features of the data sufficiently. For instance, in the Half Cylinder Ensemble dataset, the PSNR values of SRGAN and SSR-VFD degrade significantly between timesteps 20 and 50. Moreover, in the Hurricane and Shock Interaction Vortex datasets, methods other than CD-TVD also display pronounced variations in the early timesteps, likely due to the challenges posed by highly dynamic data.

\textcolor{black}{Table~\ref{fig:Average-values}} presents the averaged PSNR, LPIPS, and CD scores over all timesteps for each dataset. The mean PSNR results further confirm that CD-TVD delivers the best overall performance. In contrast, the interpolation-based TI shows the weakest results, primarily because it is unable to capture the complex nonlinear characteristics in the data.

\subsubsection{Volume Rendering Analysis}

\begin{table}[bt]
	\centering
	\caption{Average PSNR, LPIPS, and CD values with a scaling factor of 4. The best ones are highlighted in bold.}
	\begin{tabular}{ccccc}
		\hline
		\textbf{Dataset} & \textbf{Method} & \textbf{PSNR} $\uparrow$ & \textbf{LPIPS} $\downarrow$ & \textbf{CD} $\downarrow$ \\
		\hline
		\multirow{5}{*}{Tangaroa} 
		& TI        & 36.6797 & 0.1439 & 2.1923 \\
		& SRGAN     & 39.8904 & 0.1115 & 2.1292 \\
		& SSR-VFD   & 38.9527 & 0.1445 & 2.5566 \\
		& PSRFlow   & 41.7391 & 0.0970 & 1.9757 \\
		& \textbf{CD-TVD}  & \textbf{45.1457} & \textbf{0.0425} & \textbf{1.9320} \\
		\hline
		\multirow{5}{*}{Half Cylinder} 
		& TI        & 40.0438 & 0.0411 & 1.6275 \\
		& SRGAN     & 41.8104 & 0.0340 & 1.7277 \\
		& SSR-VFD   & 43.3012 & 0.0660 & 1.5289 \\
		& PSRFlow   & 43.3228 & 0.0514 & 1.6310 \\
		& \textbf{CD-TVD}  & \textbf{46.4046} & \textbf{0.0246} & \textbf{1.1318} \\
		\hline
		\multirow{5}{*}{Shock Vortex} 
		& TI        & 29.4103 & 0.0305 & 1.2840 \\
		& SRGAN     & 32.2311 & 0.0102 & 2.0138 \\
		& SSR-VFD   & 31.8848 & 0.0126 & 1.4101 \\
		& PSRFlow   & 37.4574 &0.0089 & 1.1755 \\
		& \textbf{CD-TVD}  & \textbf{43.8938} & \textbf{0.0016} & \textbf{1.0405} \\
		\hline
		\multirow{5}{*}{Hurricane} 
		& TI        & 38.6966 & 0.0802 & 3.1279 \\
		& SRGAN     & 43.6950 & 0.0696 & 2.2285 \\
		& SSR-VFD   & 44.8119 & 0.0584 & 2.6773 \\
		& PSRFlow   & 45.6459 & 0.0632 & 1.7936 \\
		& \textbf{CD-TVD}  & \textbf{48.0591} & \textbf{0.0494} & \textbf{1.4345} \\
		\hline
		
	\end{tabular}
	\label{fig:Average-values}
\end{table}

In \textcolor{black}{Fig.~\ref{fig:combined_vector_field_comparison}}, we observed the LPIPS values for the five methods in various data sets, where lower values indicate better perceptual fidelity. CD-TVD consistently achieves the lowest LPIPS values, demonstrating its superior ability to preserve perceptual details, while interpolation-based methods such as TI exhibit higher scores and thus more noticeable perceptual differences from the ground truth. Although GAN-based methods like SRGAN and SSR-VFD perform better than TI, they still lag behind CD-TVD. \textcolor{black}{Table~\ref{fig:Average-values}} further corroborates this trend, indicating that CD-TVD maintains the lowest mean LPIPS scores in all timesteps among the evaluated methods.

Analyzing the volume rendering results in \textcolor{black}{Fig.~\ref{fig:volume-rendering}} reveals that CD-TVD consistently outperforms competing methods under sparse HR conditions by leveraging prior knowledge from historical simulations. In the \textit{Tangaroa} dataset, for example, CD-TVD achieves the highest restoration fidelity in the marked region, whereas other approaches fail to preserve fine details. Similarly, in the \textit{Half Cylinder} dataset, traditional interpolation-based and GAN-based methods exhibit noticeable shape deformations, while CD-TVD retains a coherent flow structure. In the \textit{Shock Vortex} dataset, only CD-TVD reconstructs the trailing vortex and ring-shaped turbulence features with minimal artifacts, highlighting its ability to integrate learned priors for complex flow patterns. Finally, in the \textit{Hurricane} dataset, CD-TVD captures the high-frequency details near the typhoon eye more effectively than other methods, which struggle with limited HR data. This superior performance is attributed to the model’s contrastive encoding of degradation patterns and its diffusion-based approach, allowing it to recover crucial fine-scale features that other methods, lacking comprehensive prior knowledge, fail to reconstruct accurately.

\subsubsection{Streamline Rendering Analysis}

In Fig.~\ref{fig:combined_vector_field_comparison}, we showed the comparison of the CD across different methods for the four datasets.
The box plots illustrate the distribution of CD values, with lower values signifying better alignment between generated and ground-truth streamlines. CD-TVD consistently achieves the lowest CD values across all datasets, indicating superior alignment with the ground truth. Its narrow interquartile range (IQR) and minimal outliers highlight stable and accurate streamline rendering, attributed to leveraging prior knowledge from historical simulation data to better capture flow dynamics. In contrast, traditional interpolation methods exhibit higher CD values, wider IQR, and numerous outliers, reflecting poor alignment. GAN-based methods like SR-GAN and SSR-TVD display fluctuating CD values and greater variability, especially in datasets (c) and (d), indicating limitations in accurately capturing flow dynamics due to insufficient structural understanding. The superior and stable performance of CD-TVD is further confirmed by the lowest mean CD values presented in Table~\ref{fig:Average-values}.


Rendering results in \textcolor{black}{Fig.~\ref{fig:streamline-rendering}} further support these findings. In the \textit{Research Vessel Tangaroa} dataset, CD-TVD generates clear, accurate streamlines, whereas methods like SRGAN and SSR-VFD produce erratic or disconnected flow lines, especially in high-vorticity regions, due to difficulties learning high-frequency flow dynamics. CD-TVD's effective pretraining allows it to better capture these intricate features even with sparse HR data. Similarly, for the \textit{Half Cylinder Ensemble} dataset, CD-TVD provides smooth, continuous streamlines and accurately represents rapid flow transitions, outperforming other methods. Its robustness derives from pretrained knowledge of degradation patterns.

A few localised artefacts are still visible immediately below the blue bounding box in the first row and the purple bounding box in the fourth row of Fig.~\ref{fig:streamline-rendering}. These discrepancies do not originate from our super-resolved vector field itself. They arise from the accumulation of numerical errors during streamline integration: errors that inevitably grow with the distance traveled from each seed point. To verify that CD-TVD faithfully reconstructs the underlying vector field even in those areas, we include in the supplementary material a voxel-wise comparison of velocity directions. The average angular deviation is below $2^{\circ}$, confirming that the observed artifacts are confined to the visualization step and are not a failure of the reconstruction algorithm.

In conclusion, both quantitative CD analysis and qualitative streamline visualizations highlight CD-TVD's advantages. Its stable, perceptually coherent results across diverse datasets underscore superior generalization, especially in complex dynamic systems. Leveraging prior knowledge, CD-TVD achieves high performance even with limited HR data, making it highly effective for super-resolution tasks involving fluid flow and dynamic systems.

\begin{table}[bt]
	\centering
	\caption{Ablation study on the Tangaroa dataset. PSNR ($\uparrow$), LPIPS ($\downarrow$), and CD ($\downarrow$) are reported.}
	\label{tab:ablation-tangaroa}
	\begin{tabular}{cccc}
		\toprule
		\textbf{Model Variant} & \textbf{PSNR} $\uparrow$ & \textbf{LPIPS} $\downarrow$ & \textbf{CD} $\downarrow$ \\
		\midrule
		Full CD-TVD              & 45.1457 & 0.0425 & 1.3609 \\
		w/o Contrastive Modeling  & 42.5058 & 0.0889 & 2.2052 \\
		w/o Local Attention       & 42.9507 & 0.0901 & 2.1871 \\
		w/o Pre-training          & 41.9359 & 0.0991 & 2.5453 \\
		\bottomrule
	\end{tabular}
\end{table}

\begin{figure}[b]
	\centering
	\includegraphics[width=0.6\linewidth,keepaspectratio]{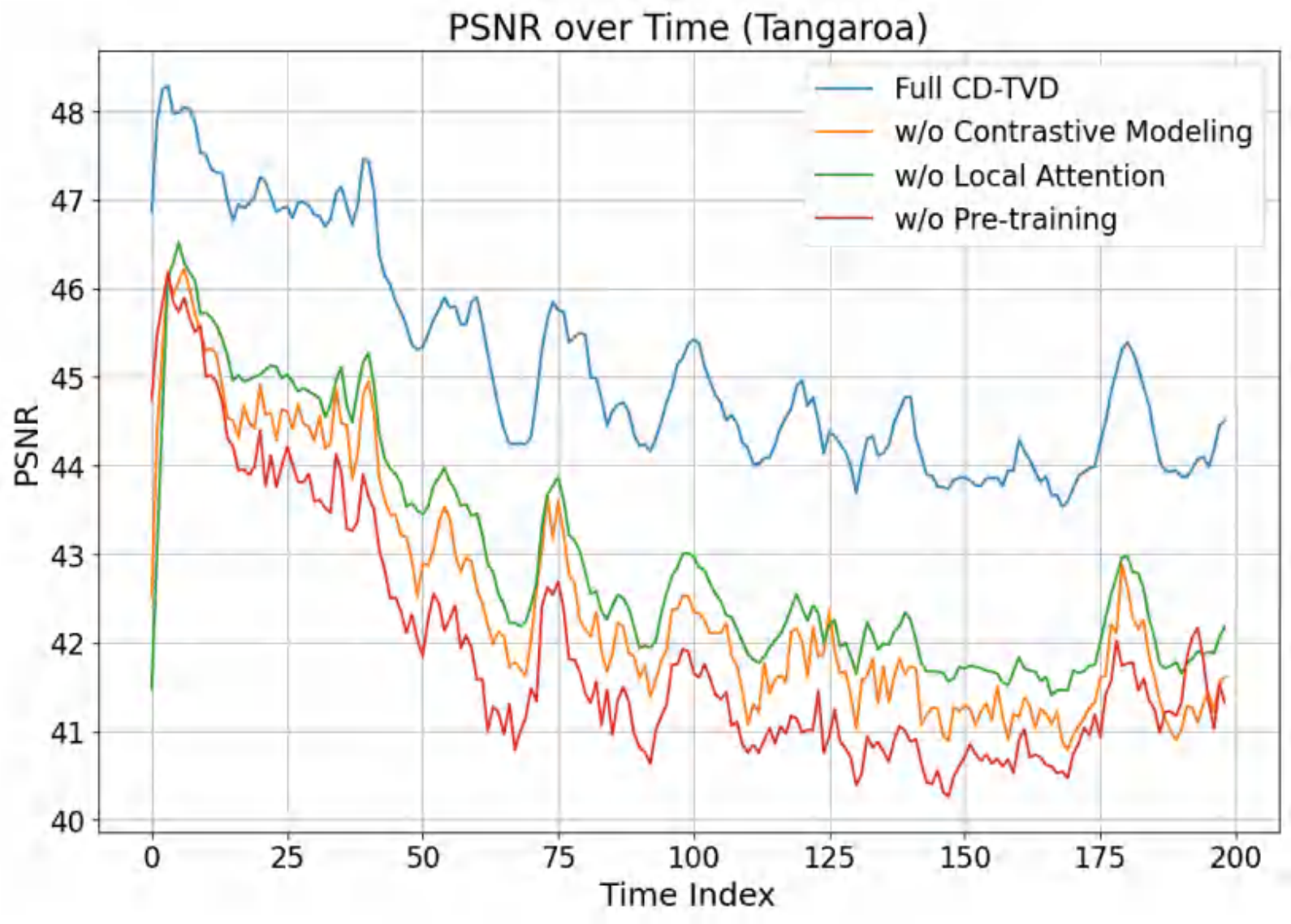}
	\caption{Ablation study results on the Tangaroa dataset evaluated by PSNR. The full CD-TVD model consistently outperforms variants without contrastive modeling, local attention, or pre-training, highlighting each component's contribution to reconstruction performance and stability.}
	\vspace{-0.15in}
	\label{fig:ablation-results}
\end{figure}

\subsection{Ablation Study}

In this section, we present an ablation study on the Tangaroa dataset to investigate the impact of different components in our proposed CD-TVD model. Table~\ref{tab:ablation-tangaroa} shows the mean values of three metrics (PSNR, LPIPS, and CD) across all timesteps under different ablation configurations. Fig.~\ref{fig:ablation-results} compares the variation of PSNR over timesteps.

\textbf{Contrastive Modeling}: Removing contrastive modeling notably decreases reconstruction accuracy and perceptual quality, causing instability and poor results. This confirms its importance in stabilizing the model by effectively learning degradation features.

\textbf{Local Attention}: Eliminating local attention and substituting it with a convolutional network leads to noticeable performance decline. Time-step-wise results (Fig.~\ref{fig:ablation-results}) emphasize local attention’s critical role in capturing fine-scale details, validating its effectiveness in our model.

\textbf{Pre-training}: The absence of pre-training significantly affects reconstructionquality and increases fluctuations, particularly at later timesteps. Pre-training stabilizes the model by supplying essential prior knowledge, enhancing adaptation to complex temporal reconstructions.

In summary, the ablation study highlights the significant contributions of contrastive modeling, local attention, and pre-training to the performance and stability of CD-TVD.

\section{DISCUSSION}
Through contrastive learning and diffusion super-resolution, CD-TVD effectively learns degradation patterns and detailed features from historical simulation data, reducing the model's reliance on HR data. Our experiments demonstrate that in new scenarios, only a single HR timestep is required to achieve super-resolution for other timesteps. However, there are still some limitations in our approach.

\textbf{Performance is sensitive to dataset similarity:}
Our method leverages features learned from historical data and applies them to new scenarios, inevitably making its performance sensitive to dataset similarity. Experiments revealed a substantial improvement in network performance when the pretraining and fine-tuning datasets were closely aligned. Conversely, notable differences between datasets negatively impacted performance. Fine-tuning with a single HR timestep partially mitigates this issue, enhancing the model's adaptability to new datasets.

\textbf{The current framework lacks end-to-end capability:}
Second, our current framework does not enable end-to-end scientific data super-resolution. Instead, it requires a two-stage process involving pretraining and fine-tuning. While this two-stage approach provides flexibility and robustness, an end-to-end approach remains a goal for future work.

\textbf{Spatial-only super-resolution constrains potential applications:}
Finally, our network currently focuses on super-resolution in the spatial domain. Temporal super-resolution is another important direction for future research. Extending our approach to the temporal dimension will be a crucial step in handling dynamic, time-varying datasets and improving the overall performance of the model in real-world scientific applications. This is one of the key areas we plan to explore in future work.

\section{Conclusions and Future Work}

In this work, we propose CD-TVD, a novel super-resolution framework tailored for scientific simulations with scarce HR temporal data. By modelling the degradation between HR and LR data as a contrastive learning task, CD-TVD effectively extracts discriminative degradation features from historical data and generalises across various physical scenarios. 

Experimental results demonstrate that CD-TVD significantly outperforms classical and state-of-the-art methods, including TI, SRGAN, SSR-VFD, and PSRFlow in both quantitative metrics (e.g., PSNR, LPIPS, CD) and visual quality. The model not only achieves fine-grained spatial structure recovery but also maintains physical consistency under constrained computational resources, making it well-suited for large-scale scientific visualization tasks. The capability to reconstruct entire time sequences from a single HR timestep greatly alleviates the dependency on data acquisition, thus enhancing the practicality of SR in real-world scientific workflows.

While the current framework addresses spatial super-resolution, scientific simulations often have sparse temporal sampling. Future work will extend CD-TVD to achieve spatiotemporal super-resolution, enabling coherent reconstruction across time and space, and enhancing the efficiency, interpretability, and scalability of scientific analyses under limited data.



\acknowledgments{%
This work was funded in part by National Natural Science Foundation of China under Grant No. 62172294, 62202446, 62302422, and CORE, a joint research center for ocean research between Laoshan Laboratory and The Hong Kong University of Science and Technology. The authors thank the anonymous reviewers for their insightful comments.
}

\bibliographystyle{abbrv-doi-hyperref}

\bibliography{template.bib}

\begin{thebibliography}{10}

\bibitem{an2021stsrnet}
Y.~An, H.-W. Shen, G.~Shan, G.~Li, and J.~Liu.
\newblock Stsrnet: Deep joint space-time super-resolution for vector field
  visualization.
\newblock {\em IEEE Computer Graphics and Applications}, 41(6):122--132, 2021.
  \href{https://doi.org/10.1109/MCG.2021.3097555}
{doi: {{%
10\hspace{.1pt}\discretionary{.}{%
}{.}\hspace{.4pt}1109\discretionary{/}{%
}{/}MCG\hspace{.1pt}\discretionary{.}{%
}{.}\hspace{.4pt}2021\hspace{.1pt}\discretionary{.}{%
}{.}\hspace{.4pt}3097555}}}


\bibitem{barrow1977parametric}
H.~G. Barrow, J.~M. Tenenbaum, R.~C. Bolles, and H.~C. Wolf.
\newblock Parametric correspondence and chamfer matching: Two new techniques
  for image matching.
\newblock In {\em Proceedings of the International Joint Conference on
  Artificial Intelligence}, p. 659–663, 1977.
  \href{https://dl.acm.org/doi/abs/10.5555/1622943.1622971}
{doi: {{%
doi\discretionary{/}{%
}{/}abs\discretionary{/}{%
}{/}10\hspace{.1pt}\discretionary{.}{%
}{.}\hspace{.4pt}5555\discretionary{/}{%
}{/}1622943\hspace{.1pt}\discretionary{.}{%
}{.}\hspace{.4pt}1622971}}}


\bibitem{cao2024survey}
H.~Cao, C.~Tan, Z.~Gao, Y.~Xu, G.~Chen, P.-A. Heng, and S.~Z. Li.
\newblock A survey on generative diffusion models.
\newblock {\em IEEE Transactions on Knowledge and Data Engineering},
  36(7):2814--2830, 2024. \href{https://doi.org/10.1109/TKDE.2024.3361474}
{doi: {{%
10\hspace{.1pt}\discretionary{.}{%
}{.}\hspace{.4pt}1109\discretionary{/}{%
}{/}TKDE\hspace{.1pt}\discretionary{.}{%
}{.}\hspace{.4pt}2024\hspace{.1pt}\discretionary{.}{%
}{.}\hspace{.4pt}3361474}}}


\bibitem{CHEN201944}
G.~Chen, B.~Dong, Y.~Zhang, W.~Lin, D.~Shen, and P.-T. Yap.
\newblock Xq-sr: Joint x-q space super-resolution with application to infant
  diffusion mri.
\newblock {\em Medical Image Analysis}, 57:44--55, 2019.
  \href{https://doi.org/10.1016/j.media.2019.06.010}
{doi: {{%
10\hspace{.1pt}\discretionary{.}{%
}{.}\hspace{.4pt}1016\discretionary{/}{%
}{/}j\hspace{.1pt}\discretionary{.}{%
}{.}\hspace{.4pt}media\hspace{.1pt}\discretionary{.}{%
}{.}\hspace{.4pt}2019\hspace{.1pt}\discretionary{.}{%
}{.}\hspace{.4pt}06\hspace{.1pt}\discretionary{.}{%
}{.}\hspace{.4pt}010}}}


\bibitem{9941138}
H.~Chung, E.~S. Lee, and J.~C. Ye.
\newblock Mr image denoising and super-resolution using regularized reverse
  diffusion.
\newblock {\em IEEE Transactions on Medical Imaging}, 42(4):922--934, 2023.
  \href{https://doi.org/10.1109/TMI.2022.3220681}
{doi: {{%
10\hspace{.1pt}\discretionary{.}{%
}{.}\hspace{.4pt}1109\discretionary{/}{%
}{/}TMI\hspace{.1pt}\discretionary{.}{%
}{.}\hspace{.4pt}2022\hspace{.1pt}\discretionary{.}{%
}{.}\hspace{.4pt}3220681}}}


\bibitem{croitoru2023diffusion}
F.-A. Croitoru, V.~Hondru, R.~T. Ionescu, and M.~Shah.
\newblock {Diffusion models in vision: A survey}.
\newblock {\em IEEE Transactions on Pattern Analysis and Machine Intelligence},
  45(9):10850--10869, 2023. \href{https://doi.org/10.1109/TPAMI.2023.3261988}
{doi: {{%
10\hspace{.1pt}\discretionary{.}{%
}{.}\hspace{.4pt}1109\discretionary{/}{%
}{/}TPAMI\hspace{.1pt}\discretionary{.}{%
}{.}\hspace{.4pt}2023\hspace{.1pt}\discretionary{.}{%
}{.}\hspace{.4pt}3261988}}}


\bibitem{daniels2021score}
M.~Daniels, T.~Maunu, and P.~Hand.
\newblock Score-based generative neural networks for large-scale optimal
  transport.
\newblock In {\em Proceedings of Advances in Neural Information Processing
  Systems}, pp. 12955--12965, 2021.
  \href{https://dl.acm.org/doi/abs/10.5555/3540261.3541253}
{doi: {{%
doi\discretionary{/}{%
}{/}abs\discretionary{/}{%
}{/}10\hspace{.1pt}\discretionary{.}{%
}{.}\hspace{.4pt}5555\discretionary{/}{%
}{/}3540261\hspace{.1pt}\discretionary{.}{%
}{.}\hspace{.4pt}3541253}}}


\bibitem{deng2019super}
Z.~Deng, C.~He, Y.~Liu, and K.~C. Kim.
\newblock Super-resolution reconstruction of turbulent velocity fields using a
  generative adversarial network-based artificial intelligence framework.
\newblock {\em Physics of Fluids}, 31(12):125111, 2019.
  \href{https://doi.org/10.1063/1.5127031}
{doi: {{%
10\hspace{.1pt}\discretionary{.}{%
}{.}\hspace{.4pt}1063\discretionary{/}{%
}{/}1\hspace{.1pt}\discretionary{.}{%
}{.}\hspace{.4pt}5127031}}}


\bibitem{NEURIPS2021_49ad23d1}
P.~Dhariwal and A.~Nichol.
\newblock Diffusion models beat gans on image synthesis.
\newblock In {\em Proceedings of Advances in Neural Information Processing
  Systems}, pp. 8780--8794, 2021.
  \href{https://doi.org/10.48550/arXiv.2105.05233}
{doi: {{%
10\hspace{.1pt}\discretionary{.}{%
}{.}\hspace{.4pt}48550\discretionary{/}{%
}{/}arXiv\hspace{.1pt}\discretionary{.}{%
}{.}\hspace{.4pt}2105\hspace{.1pt}\discretionary{.}{%
}{.}\hspace{.4pt}05233}}}


\bibitem{gao2024bayesian}
H.~Gao, X.~Han, X.~Fan, L.~Sun, L.-P. Liu, L.~Duan, and J.-X. Wang.
\newblock Bayesian conditional diffusion models for versatile spatiotemporal
  turbulence generation.
\newblock {\em Computer Methods in Applied Mechanics and Engineering},
  427:117023, 2024. \href{https://doi.org/10.1016/j.cma.2024.117023}
{doi: {{%
10\hspace{.1pt}\discretionary{.}{%
}{.}\hspace{.4pt}1016\discretionary{/}{%
}{/}j\hspace{.1pt}\discretionary{.}{%
}{.}\hspace{.4pt}cma\hspace{.1pt}\discretionary{.}{%
}{.}\hspace{.4pt}2024\hspace{.1pt}\discretionary{.}{%
}{.}\hspace{.4pt}117023}}}


\bibitem{Gao_2023_CVPR}
S.~Gao, X.~Liu, B.~Zeng, S.~Xu, Y.~Li, X.~Luo, J.~Liu, X.~Zhen, and B.~Zhang.
\newblock {Implicit diffusion models for continuous super-resolution}.
\newblock In {\em Proceedings of the IEEE Conference on Computer Vision and
  Pattern Recognition}, pp. 10021--10030, 2023.
  \href{https://doi.org/10.48550/arXiv.2303.16491}
{doi: {{%
10\hspace{.1pt}\discretionary{.}{%
}{.}\hspace{.4pt}48550\discretionary{/}{%
}{/}arXiv\hspace{.1pt}\discretionary{.}{%
}{.}\hspace{.4pt}2303\hspace{.1pt}\discretionary{.}{%
}{.}\hspace{.4pt}16491}}}


\bibitem{guo2020ssr}
L.~Guo, S.~Ye, J.~Han, H.~Zheng, H.~Gao, D.~Z. Chen, J.-X. Wang, and C.~Wang.
\newblock Ssr-vfd: Spatial super-resolution for vector field data analysis and
  visualization.
\newblock In {\em Proceedings of IEEE Pacific Visualization Symposium}, pp.
  71--80, 2020. \href{https://doi.org/10.1109/PacificVis48177.2020.8737}
{doi: {{%
10\hspace{.1pt}\discretionary{.}{%
}{.}\hspace{.4pt}1109\discretionary{/}{%
}{/}PacificVis48177\hspace{.1pt}\discretionary{.}{%
}{.}\hspace{.4pt}2020\hspace{.1pt}\discretionary{.}{%
}{.}\hspace{.4pt}8737}}}


\bibitem{han2019tsr}
J.~Han and C.~Wang.
\newblock Tsr-tvd: Temporal super-resolution for time-varying data analysis and
  visualization.
\newblock {\em IEEE Transactions on Visualization and Computer Graphics},
  26(1):205--215, 2019. \href{https://doi.org/10.1109/TVCG.2019.2934255}
{doi: {{%
10\hspace{.1pt}\discretionary{.}{%
}{.}\hspace{.4pt}1109\discretionary{/}{%
}{/}TVCG\hspace{.1pt}\discretionary{.}{%
}{.}\hspace{.4pt}2019\hspace{.1pt}\discretionary{.}{%
}{.}\hspace{.4pt}2934255}}}


\bibitem{han2020ssr}
J.~Han and C.~Wang.
\newblock Ssr-tvd: Spatial super-resolution for time-varying data analysis and
  visualization.
\newblock {\em IEEE Transactions on Visualization and Computer Graphics},
  28(6):2445--2456, 2020. \href{https://doi.org/10.1109/TVCG.2020.3032123}
{doi: {{%
10\hspace{.1pt}\discretionary{.}{%
}{.}\hspace{.4pt}1109\discretionary{/}{%
}{/}TVCG\hspace{.1pt}\discretionary{.}{%
}{.}\hspace{.4pt}2020\hspace{.1pt}\discretionary{.}{%
}{.}\hspace{.4pt}3032123}}}


\bibitem{han2022tsr}
J.~Han and C.~Wang.
\newblock Tsr-vfd: Generating temporal super-resolution for unsteady vector
  field data.
\newblock {\em Computers \& Graphics}, 103(1):168--179, 2022.
  \href{https://doi.org/10.1016/j.cag.2022.02.001}
{doi: {{%
10\hspace{.1pt}\discretionary{.}{%
}{.}\hspace{.4pt}1016\discretionary{/}{%
}{/}j\hspace{.1pt}\discretionary{.}{%
}{.}\hspace{.4pt}cag\hspace{.1pt}\discretionary{.}{%
}{.}\hspace{.4pt}2022\hspace{.1pt}\discretionary{.}{%
}{.}\hspace{.4pt}02\hspace{.1pt}\discretionary{.}{%
}{.}\hspace{.4pt}001}}}


\bibitem{han2021stnet}
J.~Han, H.~Zheng, D.~Z. Chen, and C.~Wang.
\newblock S{TN}et: An end-to-end generative framework for synthesizing
  spatiotemporal super-resolution volumes.
\newblock {\em IEEE Transactions on Visualization and Computer Graphics},
  28(1):270--280, 2021. \href{https://doi.org/10.1109/TVCG.2021.3114815}
{doi: {{%
10\hspace{.1pt}\discretionary{.}{%
}{.}\hspace{.4pt}1109\discretionary{/}{%
}{/}TVCG\hspace{.1pt}\discretionary{.}{%
}{.}\hspace{.4pt}2021\hspace{.1pt}\discretionary{.}{%
}{.}\hspace{.4pt}3114815}}}


\bibitem{ho2022cascaded}
J.~Ho, C.~Saharia, W.~Chan, D.~J. Fleet, M.~Norouzi, and T.~Salimans.
\newblock Cascaded diffusion models for high fidelity image generation.
\newblock {\em Journal of Machine Learning Research}, 23(47):1--33, 2022.
  \href{https://doi.org/10.48550/arXiv.2106.15282}
{doi: {{%
10\hspace{.1pt}\discretionary{.}{%
}{.}\hspace{.4pt}48550\discretionary{/}{%
}{/}arXiv\hspace{.1pt}\discretionary{.}{%
}{.}\hspace{.4pt}2106\hspace{.1pt}\discretionary{.}{%
}{.}\hspace{.4pt}15282}}}


\bibitem{ho2022video}
J.~Ho, T.~Salimans, A.~Gritsenko, W.~Chan, M.~Norouzi, and D.~J. Fleet.
\newblock {Video diffusion models}.
\newblock {\em In Proceedings of Advances in Neural Information Processing
  Systems}, 35(6):8633--8646, 2022.
  \href{https://doi.org/10.48550/arXiv.2204.03458}
{doi: {{%
10\hspace{.1pt}\discretionary{.}{%
}{.}\hspace{.4pt}48550\discretionary{/}{%
}{/}arXiv\hspace{.1pt}\discretionary{.}{%
}{.}\hspace{.4pt}2204\hspace{.1pt}\discretionary{.}{%
}{.}\hspace{.4pt}03458}}}


\bibitem{jiao2024ffeinr}
C.~Jiao, C.~Bi, and L.~Yang.
\newblock {FFEINR}: Flow feature-enhanced implicit neural representation for
  spatiotemporal super-resolution.
\newblock {\em Journal of Visualization}, 27(2):273--289, 2024.
  \href{https://doi.org/10.1007/s12650-024-00959-1}
{doi: {{%
10\hspace{.1pt}\discretionary{.}{%
}{.}\hspace{.4pt}1007\discretionary{/}{%
}{/}s12650\discretionary{%
}{-}{-}024\discretionary{%
}{-}{-}00959\discretionary{%
}{-}{-}1}}}


\bibitem{jiao2023esrgan}
C.~Jiao, C.~Bi, L.~Yang, Z.~Wang, Z.~Xia, and K.~Ono.
\newblock Esrgan-based visualization for large-scale volume data.
\newblock {\em Journal of Visualization}, 26(3):649--665, 2023.
  \href{https://doi.org/10.1007/s12650-022-00891-2}
{doi: {{%
10\hspace{.1pt}\discretionary{.}{%
}{.}\hspace{.4pt}1007\discretionary{/}{%
}{/}s12650\discretionary{%
}{-}{-}022\discretionary{%
}{-}{-}00891\discretionary{%
}{-}{-}2}}}


\bibitem{karatsiolis2022exploiting}
S.~Karatsiolis, C.~Padubidri, and A.~Kamilaris.
\newblock Exploiting digital surface models for inferring super-resolution for
  remotely sensed images.
\newblock {\em IEEE Transactions on Geoscience and Remote Sensing},
  60(1):1--13, 2022. \href{https://doi.org/10.1109/TGRS.2022.3209340}
{doi: {{%
10\hspace{.1pt}\discretionary{.}{%
}{.}\hspace{.4pt}1109\discretionary{/}{%
}{/}TGRS\hspace{.1pt}\discretionary{.}{%
}{.}\hspace{.4pt}2022\hspace{.1pt}\discretionary{.}{%
}{.}\hspace{.4pt}3209340}}}


\bibitem{karniadakis2021physics}
G.~E. Karniadakis, I.~G. Kevrekidis, L.~Lu, P.~Perdikaris, S.~Wang, and
  L.~Yang.
\newblock Physics-informed machine learning.
\newblock {\em Nature Reviews Physics}, 3(6):422--440, 2021.
  \href{https://doi.org/10.1038/s42254-021-00314-5}
{doi: {{%
10\hspace{.1pt}\discretionary{.}{%
}{.}\hspace{.4pt}1038\discretionary{/}{%
}{/}s42254\discretionary{%
}{-}{-}021\discretionary{%
}{-}{-}00314\discretionary{%
}{-}{-}5}}}


\bibitem{kingma2014adam}
D.~P. Kingma.
\newblock Adam: A method for stochastic optimization.
\newblock In {\em Proceedings of International Conference on Learning
  Representations}, 2015. \href{https://doi.org/10.48550/arXiv.1412.6980}
{doi: {{%
10\hspace{.1pt}\discretionary{.}{%
}{.}\hspace{.4pt}48550\discretionary{/}{%
}{/}arXiv\hspace{.1pt}\discretionary{.}{%
}{.}\hspace{.4pt}1412\hspace{.1pt}\discretionary{.}{%
}{.}\hspace{.4pt}6980}}}


\bibitem{8099502}
C.~Ledig, L.~Theis, F.~Huszár, J.~Caballero, A.~Cunningham, A.~Acosta,
  A.~Aitken, A.~Tejani, J.~Totz, Z.~Wang, and W.~Shi.
\newblock Photo-realistic single image super-resolution using a generative
  adversarial network.
\newblock In {\em Proceedings of IEEE Conference on Computer Vision and Pattern
  Recognition}, pp. 105--114, 2017. \href{https://doi.org/10.17863/CAM.51996}
{doi: {{%
10\hspace{.1pt}\discretionary{.}{%
}{.}\hspace{.4pt}17863\discretionary{/}{%
}{/}CAM\hspace{.1pt}\discretionary{.}{%
}{.}\hspace{.4pt}51996}}}


\bibitem{lepcha2023image}
D.~C. Lepcha, B.~Goyal, A.~Dogra, and V.~Goyal.
\newblock Image super-resolution: A comprehensive review, recent trends,
  challenges and applications.
\newblock {\em Information Fusion}, 91(1):230--260, 2023.
  \href{https://doi.org/10.1016/j.inffus.2022.10.007}
{doi: {{%
10\hspace{.1pt}\discretionary{.}{%
}{.}\hspace{.4pt}1016\discretionary{/}{%
}{/}j\hspace{.1pt}\discretionary{.}{%
}{.}\hspace{.4pt}inffus\hspace{.1pt}\discretionary{.}{%
}{.}\hspace{.4pt}2022\hspace{.1pt}\discretionary{.}{%
}{.}\hspace{.4pt}10\hspace{.1pt}\discretionary{.}{%
}{.}\hspace{.4pt}007}}}


\bibitem{LI202247}
H.~Li, Y.~Yang, M.~Chang, S.~Chen, H.~Feng, Z.~Xu, Q.~Li, and Y.~Chen.
\newblock S{RD}iff: Single image super-resolution with diffusion probabilistic
  models.
\newblock {\em Neurocomputing}, 479(1):47--59, 2022.
  \href{https://doi.org/10.1016/j.neucom.2022.01.029}
{doi: {{%
10\hspace{.1pt}\discretionary{.}{%
}{.}\hspace{.4pt}1016\discretionary{/}{%
}{/}j\hspace{.1pt}\discretionary{.}{%
}{.}\hspace{.4pt}neucom\hspace{.1pt}\discretionary{.}{%
}{.}\hspace{.4pt}2022\hspace{.1pt}\discretionary{.}{%
}{.}\hspace{.4pt}01\hspace{.1pt}\discretionary{.}{%
}{.}\hspace{.4pt}029}}}


\bibitem{rs14194834}
J.~Liu, Z.~Yuan, Z.~Pan, Y.~Fu, L.~Liu, and B.~Lu.
\newblock Diffusion model with detail complement for super-resolution of remote
  sensing.
\newblock {\em Remote Sensing}, 14(19):4834, 2022.
  \href{https://doi.org/10.3390/rs14194834}
{doi: {{%
10\hspace{.1pt}\discretionary{.}{%
}{.}\hspace{.4pt}3390\discretionary{/}{%
}{/}rs14194834}}}


\bibitem{liu2024uginr}
K.~Liu, C.~Jiao, X.~Gao, and C.~Bi.
\newblock Uginr: Large-scale unstructured grid reduction via implicit neural
  representation.
\newblock {\em Journal of Visualization}, 27(5):983--996, 2024.
  \href{https://doi.org/10.1007/s12650-024-01003-y}
{doi: {{%
10\hspace{.1pt}\discretionary{.}{%
}{.}\hspace{.4pt}1007\discretionary{/}{%
}{/}s12650\discretionary{%
}{-}{-}024\discretionary{%
}{-}{-}01003\discretionary{%
}{-}{-}y}}}


\bibitem{liu2023pre}
P.~Liu, W.~Yuan, J.~Fu, Z.~Jiang, H.~Hayashi, and G.~Neubig.
\newblock Pre-train, prompt, and predict: A systematic survey of prompting
  methods in natural language processing.
\newblock {\em ACM computing surveys}, 55(9):1--35, 2023.
  \href{https://doi.org/10.1145/3560815}
{doi: {{%
10\hspace{.1pt}\discretionary{.}{%
}{.}\hspace{.4pt}1145\discretionary{/}{%
}{/}3560815}}}


\bibitem{Metzger_2023_CVPR}
N.~Metzger, R.~C. Daudt, and K.~Schindler.
\newblock Guided depth super-resolution by deep anisotropic diffusion.
\newblock In {\em Proceedings of the IEEE Conference on Computer Vision and
  Pattern Recognition}, pp. 18237--18246, 2023.
  \href{https://doi.org/10.48550/arXiv.2211.11592}
{doi: {{%
10\hspace{.1pt}\discretionary{.}{%
}{.}\hspace{.4pt}48550\discretionary{/}{%
}{/}arXiv\hspace{.1pt}\discretionary{.}{%
}{.}\hspace{.4pt}2211\hspace{.1pt}\discretionary{.}{%
}{.}\hspace{.4pt}11592}}}


\bibitem{Nichol2021ImprovedDD}
A.~Q. Nichol and P.~Dhariwal.
\newblock Improved denoising diffusion probabilistic models.
\newblock In {\em Proceedings of International Conference on Machine Learning},
  pp. 8162--8171, 2021. \href{https://doi.org/10.48550/arXiv.2102.09672}
{doi: {{%
10\hspace{.1pt}\discretionary{.}{%
}{.}\hspace{.4pt}48550\discretionary{/}{%
}{/}arXiv\hspace{.1pt}\discretionary{.}{%
}{.}\hspace{.4pt}2102\hspace{.1pt}\discretionary{.}{%
}{.}\hspace{.4pt}09672}}}


\bibitem{ning2016joint}
L.~Ning, K.~Setsompop, O.~Michailovich, N.~Makris, M.~E. Shenton, C.-F. Westin,
  and Y.~Rathi.
\newblock A joint compressed-sensing and super-resolution approach for very
  high-resolution diffusion imaging.
\newblock {\em NeuroImage}, 125(1):386--400, 2016.
  \href{https://doi.org/10.1016/j.neuroimage.2015.10.061}
{doi: {{%
10\hspace{.1pt}\discretionary{.}{%
}{.}\hspace{.4pt}1016\discretionary{/}{%
}{/}j\hspace{.1pt}\discretionary{.}{%
}{.}\hspace{.4pt}neuroimage\hspace{.1pt}\discretionary{.}{%
}{.}\hspace{.4pt}2015\hspace{.1pt}\discretionary{.}{%
}{.}\hspace{.4pt}10\hspace{.1pt}\discretionary{.}{%
}{.}\hspace{.4pt}061}}}


\bibitem{rombach2022high}
R.~Rombach, A.~Blattmann, D.~Lorenz, P.~Esser, and B.~Ommer.
\newblock {High-resolution image synthesis with latent diffusion models}.
\newblock In {\em Proceedings of the IEEE Conference on Computer Vision and
  Pattern Recognition}, pp. 10684--10695, 2022.
  \href{https://doi.org/10.48550/arXiv.2112.10752}
{doi: {{%
10\hspace{.1pt}\discretionary{.}{%
}{.}\hspace{.4pt}48550\discretionary{/}{%
}{/}arXiv\hspace{.1pt}\discretionary{.}{%
}{.}\hspace{.4pt}2112\hspace{.1pt}\discretionary{.}{%
}{.}\hspace{.4pt}10752}}}


\bibitem{saharia2022image}
C.~Saharia, J.~Ho, W.~Chan, T.~Salimans, D.~J. Fleet, and M.~Norouzi.
\newblock Image super-resolution via iterative refinement.
\newblock {\em IEEE Transactions on Pattern Analysis and Machine Intelligence},
  45(4):4713--4726, 2022. \href{https://doi.org/10.1109/TPAMI.2022.3204461}
{doi: {{%
10\hspace{.1pt}\discretionary{.}{%
}{.}\hspace{.4pt}1109\discretionary{/}{%
}{/}TPAMI\hspace{.1pt}\discretionary{.}{%
}{.}\hspace{.4pt}2022\hspace{.1pt}\discretionary{.}{%
}{.}\hspace{.4pt}3204461}}}


\bibitem{schanz2023stochastic}
A.~Schanz, F.~List, and O.~Hahn.
\newblock Stochastic super-resolution of cosmological simulations with
  denoising diffusion models.
\newblock {\em The Open Journal of Astrophysics}, 7(8), 2024.
  \href{https://doi.org/10.33232/001c.125902}
{doi: {{%
10\hspace{.1pt}\discretionary{.}{%
}{.}\hspace{.4pt}33232\discretionary{/}{%
}{/}001c\hspace{.1pt}\discretionary{.}{%
}{.}\hspace{.4pt}125902}}}


\bibitem{shen2023psrflow}
J.~Shen and H.-W. Shen.
\newblock {PSRF}low: Probabilistic super resolution with flow-based models for
  scientific data.
\newblock {\em IEEE Transactions on Visualization and Computer Graphics},
  30(3):986--996, 2023. \href{https://doi.org/10.1109/TVCG.2023.3327171}
{doi: {{%
10\hspace{.1pt}\discretionary{.}{%
}{.}\hspace{.4pt}1109\discretionary{/}{%
}{/}TVCG\hspace{.1pt}\discretionary{.}{%
}{.}\hspace{.4pt}2023\hspace{.1pt}\discretionary{.}{%
}{.}\hspace{.4pt}3327171}}}


\bibitem{shen2024generative}
L.~Shen, L.~Deng, X.~Liu, Y.~Wang, X.~Chen, and J.~Liu.
\newblock A generative adversarial network based on an efficient transformer
  for high-fidelity flow field reconstruction.
\newblock {\em Physics of Fluids}, 36(7), 2024.
  \href{https://doi.org/10.1063/5.0215681}
{doi: {{%
10\hspace{.1pt}\discretionary{.}{%
}{.}\hspace{.4pt}1063\discretionary{/}{%
}{/}5\hspace{.1pt}\discretionary{.}{%
}{.}\hspace{.4pt}0215681}}}


\bibitem{shen2024pcsagan}
L.~Shen, L.~Deng, Y.~Wang, J.~Zhang, and J.~Liu.
\newblock Pcsagan: A physics-constrained generative network based on
  self-attention for high-fidelity flow field reconstruction.
\newblock {\em Journal of Visualization}, 27(4):661--676, 2024.
  \href{https://doi.org/10.1007/s12650-024-00987-x}
{doi: {{%
10\hspace{.1pt}\discretionary{.}{%
}{.}\hspace{.4pt}1007\discretionary{/}{%
}{/}s12650\discretionary{%
}{-}{-}024\discretionary{%
}{-}{-}00987\discretionary{%
}{-}{-}x}}}


\bibitem{song2024forecasting}
J.~Song, Z.~Song, P.~Ren, N.~B. Erichson, M.~W. Mahoney, and X.~S. Li.
\newblock Forecasting high-dimensional spatio-temporal systems from sparse
  measurements.
\newblock {\em Machine Learning: Science and Technology}, 5(4):045067, 2024.
  \href{https://doi.org/10.1088/2632-2153/ad9883}
{doi: {{%
10\hspace{.1pt}\discretionary{.}{%
}{.}\hspace{.4pt}1088\discretionary{/}{%
}{/}2632\discretionary{%
}{-}{-}2153\discretionary{/}{%
}{/}ad9883}}}


\bibitem{song2023sparse}
X.~Song, G.~Wang, W.~Zhong, K.~Guo, Z.~Li, X.~Liu, J.~Dong, and Q.~Liu.
\newblock Sparse-view reconstruction for photoacoustic tomography combining
  diffusion model with model-based iteration.
\newblock {\em Photoacoustics}, 33(1):100558, 2023.
  \href{https://doi.org/10.1016/j.pacs.2023.100558}
{doi: {{%
10\hspace{.1pt}\discretionary{.}{%
}{.}\hspace{.4pt}1016\discretionary{/}{%
}{/}j\hspace{.1pt}\discretionary{.}{%
}{.}\hspace{.4pt}pacs\hspace{.1pt}\discretionary{.}{%
}{.}\hspace{.4pt}2023\hspace{.1pt}\discretionary{.}{%
}{.}\hspace{.4pt}100558}}}


\bibitem{VIS2021118673}
G.~Vis, M.~Nilsson, C.-F. Westin, and F.~Szczepankiewicz.
\newblock Accuracy and precision in super-resolution mri: Enabling spherical
  tensor diffusion encoding at ultra-high b-values and high resolution.
\newblock {\em NeuroImage}, 245(2):118673, 2021.
  \href{https://doi.org/10.1016/j.neuroimage.2021.118673}
{doi: {{%
10\hspace{.1pt}\discretionary{.}{%
}{.}\hspace{.4pt}1016\discretionary{/}{%
}{/}j\hspace{.1pt}\discretionary{.}{%
}{.}\hspace{.4pt}neuroimage\hspace{.1pt}\discretionary{.}{%
}{.}\hspace{.4pt}2021\hspace{.1pt}\discretionary{.}{%
}{.}\hspace{.4pt}118673}}}


\bibitem{wang2022dl4scivis}
C.~Wang and J.~Han.
\newblock Dl4scivis: A state-of-the-art survey on deep learning for scientific
  visualization.
\newblock {\em IEEE Transactions on Visualization and Computer Graphics},
  29(8):3714--3733, 2022. \href{https://doi.org/10.1109/TVCG.2022.3167896}
{doi: {{%
10\hspace{.1pt}\discretionary{.}{%
}{.}\hspace{.4pt}1109\discretionary{/}{%
}{/}TVCG\hspace{.1pt}\discretionary{.}{%
}{.}\hspace{.4pt}2022\hspace{.1pt}\discretionary{.}{%
}{.}\hspace{.4pt}3167896}}}


\bibitem{wang2024pmim}
M.~Wang, C.~Bi, L.~Yang, X.~Qiu, Y.~Li, and C.~Yu.
\newblock Pmim: Generating high-resolution air pollution data via masked image
  modeling.
\newblock {\em Journal of Visualization}, 27(3):383--399, 2024.
  \href{https://doi.org/10.1007/s12650-024-00965-3}
{doi: {{%
10\hspace{.1pt}\discretionary{.}{%
}{.}\hspace{.4pt}1007\discretionary{/}{%
}{/}s12650\discretionary{%
}{-}{-}024\discretionary{%
}{-}{-}00965\discretionary{%
}{-}{-}3}}}


\bibitem{wang2022comprehensive}
P.~Wang, B.~Bayram, and E.~Sertel.
\newblock A comprehensive review on deep learning based remote sensing image
  super-resolution methods.
\newblock {\em Earth-Science Reviews}, 232(1):104110, 2022.
  \href{https://doi.org/10.1016/j.earscirev.2022.104110}
{doi: {{%
10\hspace{.1pt}\discretionary{.}{%
}{.}\hspace{.4pt}1016\discretionary{/}{%
}{/}j\hspace{.1pt}\discretionary{.}{%
}{.}\hspace{.4pt}earscirev\hspace{.1pt}\discretionary{.}{%
}{.}\hspace{.4pt}2022\hspace{.1pt}\discretionary{.}{%
}{.}\hspace{.4pt}104110}}}


\bibitem{wang2024semi}
X.~Wang, Y.~Dong, S.~Zou, L.~Zhang, and X.~Deng.
\newblock A semi-supervised framework for computational fluid dynamics
  prediction.
\newblock {\em Applied Soft Computing}, 154(6):111422, 2024.
  \href{https://doi.org/10.1016/j.asoc.2024.111422}
{doi: {{%
10\hspace{.1pt}\discretionary{.}{%
}{.}\hspace{.4pt}1016\discretionary{/}{%
}{/}j\hspace{.1pt}\discretionary{.}{%
}{.}\hspace{.4pt}asoc\hspace{.1pt}\discretionary{.}{%
}{.}\hspace{.4pt}2024\hspace{.1pt}\discretionary{.}{%
}{.}\hspace{.4pt}111422}}}


\bibitem{wang2020deep}
Z.~Wang, J.~Chen, and S.~C. Hoi.
\newblock Deep learning for image super-resolution: A survey.
\newblock {\em IEEE Transactions on Pattern Analysis and Machine Intelligence},
  43(10):3365--3387, 2020. \href{https://doi.org/10.1109/TPAMI.2020.2982166}
{doi: {{%
10\hspace{.1pt}\discretionary{.}{%
}{.}\hspace{.4pt}1109\discretionary{/}{%
}{/}TPAMI\hspace{.1pt}\discretionary{.}{%
}{.}\hspace{.4pt}2020\hspace{.1pt}\discretionary{.}{%
}{.}\hspace{.4pt}2982166}}}


\bibitem{Wu2023Practical}
G.~Wu, J.~Jiang, and X.~Liu.
\newblock A practical contrastive learning framework for single-image
  super-resolution.
\newblock {\em IEEE Transactions on Neural Networks and Learning Systems},
  35(3):15834--15845, 2024. \href{https://doi.org/10.1109/TNNLS.2023.3290038}
{doi: {{%
10\hspace{.1pt}\discretionary{.}{%
}{.}\hspace{.4pt}1109\discretionary{/}{%
}{/}TNNLS\hspace{.1pt}\discretionary{.}{%
}{.}\hspace{.4pt}2023\hspace{.1pt}\discretionary{.}{%
}{.}\hspace{.4pt}3290038}}}


\bibitem{WU2023104901}
Z.~Wu, X.~Chen, S.~Xie, J.~Shen, and Y.~Zeng.
\newblock Super-resolution of brain mri images based on denoising diffusion
  probabilistic model.
\newblock {\em Biomedical Signal Processing and Control}, 85(1):104901, 2023.
  \href{https://doi.org/10.1016/j.bspc.2023.104901}
{doi: {{%
10\hspace{.1pt}\discretionary{.}{%
}{.}\hspace{.4pt}1016\discretionary{/}{%
}{/}j\hspace{.1pt}\discretionary{.}{%
}{.}\hspace{.4pt}bspc\hspace{.1pt}\discretionary{.}{%
}{.}\hspace{.4pt}2023\hspace{.1pt}\discretionary{.}{%
}{.}\hspace{.4pt}104901}}}


\bibitem{9920542}
S.~W. Wurster, H.~Guo, H.-W. Shen, T.~Peterka, and J.~Xu.
\newblock Deep hierarchical super resolution for scientific data.
\newblock {\em IEEE Transactions on Visualization and Computer Graphics},
  29(12):5483--5495, 2023. \href{https://doi.org/10.1109/TVCG.2022.3214420}
{doi: {{%
10\hspace{.1pt}\discretionary{.}{%
}{.}\hspace{.4pt}1109\discretionary{/}{%
}{/}TVCG\hspace{.1pt}\discretionary{.}{%
}{.}\hspace{.4pt}2022\hspace{.1pt}\discretionary{.}{%
}{.}\hspace{.4pt}3214420}}}


\bibitem{xiao2021tackling}
Z.~Xiao, K.~Kreis, and A.~Vahdat.
\newblock Tackling the generative learning trilemma with denoising diffusion
  {GAN}s.
\newblock In {\em Proceedings of International Conference on Learning
  Representations}, 2022. \href{https://doi.org/10.48550/arXiv.2112.07804}
{doi: {{%
10\hspace{.1pt}\discretionary{.}{%
}{.}\hspace{.4pt}48550\discretionary{/}{%
}{/}arXiv\hspace{.1pt}\discretionary{.}{%
}{.}\hspace{.4pt}2112\hspace{.1pt}\discretionary{.}{%
}{.}\hspace{.4pt}07804}}}


\bibitem{xie2018tempogan}
Y.~Xie, E.~Franz, M.~Chu, and N.~Thuerey.
\newblock {Tempo{GAN}: A temporally coherent, volumetric {GAN} for
  super-resolution fluid flow}.
\newblock {\em ACM Transactions on Graphics}, 37(4):1--15, 2018.
  \href{https://doi.org/10.1145/3197517.3201304}
{doi: {{%
10\hspace{.1pt}\discretionary{.}{%
}{.}\hspace{.4pt}1145\discretionary{/}{%
}{/}3197517\hspace{.1pt}\discretionary{.}{%
}{.}\hspace{.4pt}3201304}}}


\bibitem{Yang2022DiffusionMA}
L.~Yang, Z.~Zhang, S.~Hong, R.~Xu, Y.~Zhao, Y.~Shao, W.~Zhang, M.-H. Yang, and
  B.~Cui.
\newblock Diffusion models: A comprehensive survey of methods and applications.
\newblock {\em ACM Computing Surveys}, 56(4):1--39, 2022.
  \href{https://doi.org/10.1145/3626235}
{doi: {{%
10\hspace{.1pt}\discretionary{.}{%
}{.}\hspace{.4pt}1145\discretionary{/}{%
}{/}3626235}}}


\bibitem{yang2019deep}
W.~Yang, X.~Zhang, Y.~Tian, W.~Wang, J.-H. Xue, and Q.~Liao.
\newblock Deep learning for single image super-resolution: A brief review.
\newblock {\em IEEE Transactions on Multimedia}, 21(12):3106--3121, 2019.
  \href{https://doi.org/10.1109/TMM.2019.2919431}
{doi: {{%
10\hspace{.1pt}\discretionary{.}{%
}{.}\hspace{.4pt}1109\discretionary{/}{%
}{/}TMM\hspace{.1pt}\discretionary{.}{%
}{.}\hspace{.4pt}2019\hspace{.1pt}\discretionary{.}{%
}{.}\hspace{.4pt}2919431}}}


\bibitem{yang2024adaptive}
Y.~Yang, C.~Jiao, X.~Gao, X.~Tian, and C.~Bi.
\newblock Adaptive volumetric data compression based on implicit neural
  representation.
\newblock In {\em Proceedings of the International Symposium on Visual
  Information Communication and Interaction}, pp. 1--8, 2024.
  \href{https://doi.org/10.1145/3678698.3678703}
{doi: {{%
10\hspace{.1pt}\discretionary{.}{%
}{.}\hspace{.4pt}1145\discretionary{/}{%
}{/}3678698\hspace{.1pt}\discretionary{.}{%
}{.}\hspace{.4pt}3678703}}}


\bibitem{NEURIPS2023_2ac2eac5}
Z.~Yue, J.~Wang, and C.~C. Loy.
\newblock Resshift: Efficient diffusion model for image super-resolution by
  residual shifting.
\newblock In {\em Proceedings of Advances in Neural Information Processing
  Systems}, pp. 13294--13307, 2023.
  \href{https://doi.org/10.48550/arXiv.2307.12348}
{doi: {{%
10\hspace{.1pt}\discretionary{.}{%
}{.}\hspace{.4pt}48550\discretionary{/}{%
}{/}arXiv\hspace{.1pt}\discretionary{.}{%
}{.}\hspace{.4pt}2307\hspace{.1pt}\discretionary{.}{%
}{.}\hspace{.4pt}12348}}}


\bibitem{zhang2018unreasonable}
R.~Zhang, P.~Isola, A.~A. Efros, E.~Shechtman, and O.~Wang.
\newblock The unreasonable effectiveness of deep features as a perceptual
  metric.
\newblock In {\em Proceedings of the IEEE Conference on Computer Vision and
  Pattern Recognition}, pp. 586--595, 2018.
  \href{https://doi.org/10.1109/CVPR.2018.00068}
{doi: {{%
10\hspace{.1pt}\discretionary{.}{%
}{.}\hspace{.4pt}1109\discretionary{/}{%
}{/}CVPR\hspace{.1pt}\discretionary{.}{%
}{.}\hspace{.4pt}2018\hspace{.1pt}\discretionary{.}{%
}{.}\hspace{.4pt}00068}}}


\bibitem{zhou2017volume}
Z.~Zhou, Y.~Hou, Q.~Wang, G.~Chen, J.~Lu, Y.~Tao, and H.~Lin.
\newblock Volume upscaling with convolutional neural networks.
\newblock In {\em Proceedings of the Computer Graphics International
  Conference}, pp. 1--6, 2017. \href{https://doi.org/10.1145/3095140.3095178}
{doi: {{%
10\hspace{.1pt}\discretionary{.}{%
}{.}\hspace{.4pt}1145\discretionary{/}{%
}{/}3095140\hspace{.1pt}\discretionary{.}{%
}{.}\hspace{.4pt}3095178}}}


\bibitem{zuo2023high}
Z.~Zuo, T.~Fang, H.~Wu, and Z.~Zhang.
\newblock High-resolution reconstruction algorithm for the three-dimensional
  velocity field produced by atomization of two impinging jets based on deep
  learning.
\newblock {\em Physics of Fluids}, 35(6):063306, 2023.
  \href{https://doi.org/10.1063/5.0152779}
{doi: {{%
10\hspace{.1pt}\discretionary{.}{%
}{.}\hspace{.4pt}1063\discretionary{/}{%
}{/}5\hspace{.1pt}\discretionary{.}{%
}{.}\hspace{.4pt}0152779}}}


\end{thebibliography}

\appendix 

\end{document}